\documentclass[10pt]{article} 
\usepackage[accepted]{tmlr}
\usepackage{amsmath,amsfonts,bm}
\usepackage{hyperref}
\usepackage{url}
\usepackage{tikz}
\usetikzlibrary{positioning}
\usepackage{subcaption}
\usepackage{dirtytalk}
\usepackage{dsfont}
\usepackage[normalem]{ulem}
\usepackage{color}
\usepackage{tabularx}
\usepackage{multirow}
\usepackage{booktabs}
\usepackage{graphicx}
\usepackage[export]{adjustbox}
\usepackage{longtable}
\usepackage{wrapfig}
\usepackage{standalone}
\usepackage{floatrow}

\definecolor{airforceblue}{rgb}{0.36, 0.54, 0.66}
\definecolor{bleudefrance}{rgb}{0.19, 0.55, 0.91}
\definecolor{azure}{rgb}{0.0, 0.5, 1.0}
\definecolor{ao}{rgb}{0.0, 0.0, 1.0}
\definecolor{blue(pigment)}{rgb}{0.2, 0.2, 0.6}
\definecolor{brandeisblue}{rgb}{0.0, 0.44, 1.0}
\definecolor{cobalt}{rgb}{0.0, 0.28, 0.67}
\definecolor{darkcerulean}{rgb}{0.03, 0.27, 0.49}
\definecolor{darkmidnightblue}{rgb}{0.0, 0.2, 0.4}
\definecolor{darkpowderblue}{rgb}{0.0, 0.2, 0.6}
\definecolor{darkslateblue}{rgb}{0.28, 0.24, 0.55}
\definecolor{egyptianblue}{rgb}{0.06, 0.2, 0.65}
\definecolor{internationalkleinblue}{rgb}{0.0, 0.18, 0.65}
\definecolor{mediumblue}{rgb}{0.0, 0.0, 0.8}
  
\DeclareMathOperator{\cossim}{sim}
\newcommand{\red}[1]{{\color{red}{#1}}}

\newcommand{\blu}[1]{{\color{mediumblue}{#1}}}

\title{Know Your Self-supervised Learning: A Survey on\\Image-based Generative and Discriminative Training}

\author{\name Utku Ozbulak${}^{1,2}$ \email utku.ozbulak@ghent.ac.kr \\
      \name Hyun Jung Lee${}^{1,2}$ \email hyunjung.lee@ghent.ac.kr \\
      \name Beril Boga${}^{3}$ \email beril.boga@bshg.com\\
      \name Esla Timothy Anzaku${}^{1,2}$ \email eslatimothy.anzaku@ghent.ac.kr \\
      \name Homin Park${}^{1,2}$ \email homin.park@ghent.ac.kr \\
      \name Arnout Van Messem${}^{4}$ \email arnout.vanmessem@uliege.be \\
      \name Wesley De Neve${}^{1,2}$ \email wesley.deneve@ghent.ac.kr \\
      \name Joris Vankerschaver${}^{1,2}$ \email joris.vankerschaver@ghent.ac.kr \\
      \vspace{0.25em}
      \\
      \addr ${}^{1}$Ghent University, Belgium \\
      \addr ${}^{2}$Ghent University Global Campus, South Korea \\
      \addr ${}^{3}$BSH Hausgeräte GmbH, Germany \\
      \addr ${}^{4}$University of Liège, Belgium
      }

\def\month{05}  
\def\year{2023} 
\def\openreview{\url{https://openreview.net/forum?id=Ma25S4ludQ}} 

\begin{document}
\maketitle

\begin{abstract}
Although supervised learning has been highly successful in improving the state-of-the-art in the domain of image-based computer vision in the past, the margin of improvement has diminished significantly in recent years, indicating that a plateau is in sight. Meanwhile, the use of self-supervised learning (SSL) for the purpose of natural language processing (NLP) has seen tremendous successes during the past couple of years, with this new learning paradigm yielding powerful language models. Inspired by the excellent results obtained in the field of NLP, self-supervised methods that rely on clustering, contrastive learning, distillation, and information-maximization, which all fall under the banner of discriminative SSL, have experienced a swift uptake in the area of computer vision. Shortly afterwards, generative SSL frameworks that are mostly based on masked image modeling, complemented and surpassed the results obtained with discriminative SSL. Consequently, within a span of three years, over $100$ unique general-purpose frameworks for generative and discriminative SSL, with a focus on imaging, were proposed. In this survey, we review a plethora of research efforts conducted on image-oriented SSL, providing a historic view and paying attention to best practices as well as useful software packages. While doing so, we discuss pretext tasks for image-based SSL, as well as techniques that are commonly used in image-based SSL. Lastly, to aid researchers who aim at contributing to image-focused SSL, we outline a number of promising research directions. 
\end{abstract}

\section{Introduction}
\label{sec:introduction}

The remarkable feature extraction capabilities of deep neural networks (DNNs) have enabled their effective utilization in numerous visual tasks. Although the core building blocks that are in common use today were already proposed two decades ago~\citep{lecun1998gradient}, DNNs only became the go-to models after the introduction of AlexNet~\citep{Alexnet}, a DNN architecture that was able to obtain exceptional results for the ImageNet Large Scale Visual Recognition Challenge~\citep{ILSVRC15:rus} that took place in 2012, by leveraging vast amounts of computational resources (at that time) and large amounts of labeled data. Since then, the availability of standardized datasets in the image domain such as MNIST~\citep{lecun1998gradient}, CIFAR~\citep{cifar}, SVHN~\citep{svhn}, COCO~\citep{coco}, and ImageNet enabled standardized experimentation, with these datasets acting as catalysts for major advancements in the area of supervised learning. Starting with AlexNet, the classification accuracy of DNNs on ImageNet improved year after year thanks to better and novel architectural designs (e.g., VGG~\citep{VGG}, ResNet~\citep{resnet}, InceptionNet~\citep{Inceptionv1,inceptionv3}, ViT~\citep{vit}), augmentation techniques, optimizers, and activation functions, as well as methods for smoother training~\citep{cosine_anneal,cutmix,batch_norm,adaptive_momentum,ELU_activation}.

\begin{figure}
\centering
\begin{subfigure}{.49\textwidth}
\includegraphics[width=\textwidth]{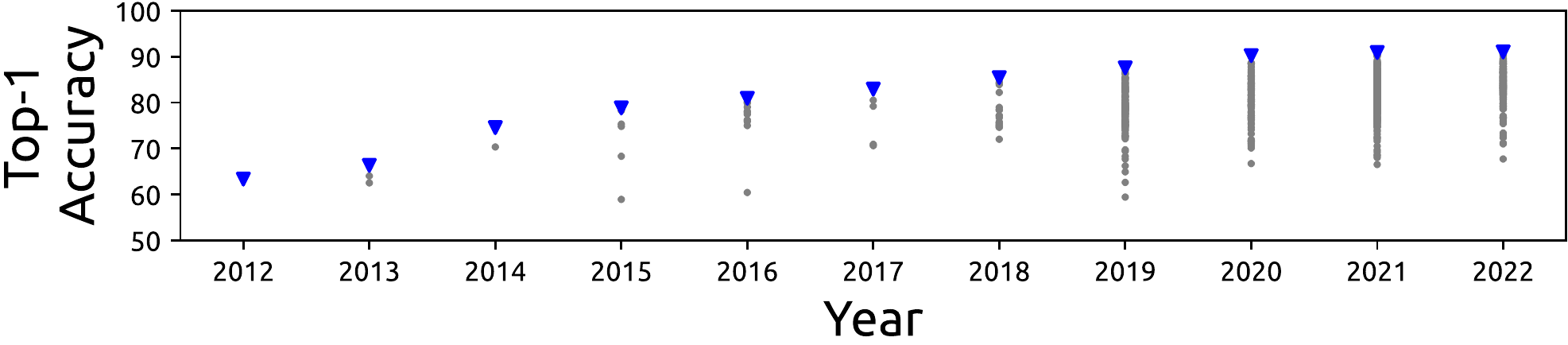}
\caption{ImageNet top-1 accuracy}
\end{subfigure}
\begin{subfigure}{.49\textwidth}
\includegraphics[width=\textwidth]{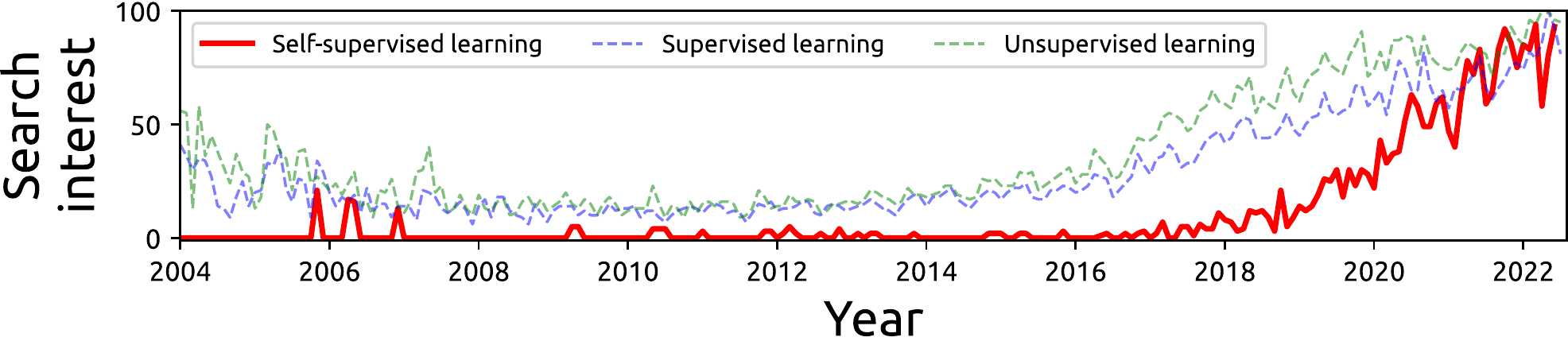}
\caption{Interest over time for different learning paradigms}
\end{subfigure}
\caption{(a) ImageNet top-1 accuracy for DNNs proposed between 2012 - 2022 and (b) interest over time for three popular learning paradigms between 2004 - 2022, as measured with Google Trends.}
\label{fig:ssl_interest}
\end{figure}

Unfortunately, not all datasets come with an abundance of labeled training data. In order to overcome this hurdle and to facilitate the application of DNNs to smaller datasets, transfer learning was introduced and soon became the dominant method to transfer knowledge across image datasets~\citep{transfer_learning_1}. Although transfer learning enables the usage of DNNs for smaller datasets thanks to features extracted from larger datasets, models trained in this way are known to be brittle and sensitive to small changes in the data~\citep{transfer_learning_brittle} due to the use of supervised pre-training. Furthermore, shortcomings of supervised learning also became apparent when improvements obtained with these methods came to a halt in recent years (see Figure~\ref{fig:ssl_interest} for top-1 accuracy on ImageNet), thus calling for research efforts that go beyond the use of supervised learning~\citep{zisserman_ssl}. In order to overcome the limitations of supervised learning, countless studies investigated the line of unsupervised learning, which aims at enabling robust feature extraction through the training of models without label information~\citep{unsupervsed_survey}. Unfortunately, results obtained by these methods on image datasets fell short until recently~\citep{puzzle_ssl,inpainting_ssl}, while the use of self-supervised methods in the field of natural language processing (NLP) achieved state-of-the-art results, compared to supervised learning techniques~\citep{bert,gpt_2}.

As mentioned above, the field of NLP enjoyed the success of self-supervised models over supervised ones earlier than the field of computer vision, with models such as \texttt{BERT}, \texttt{GPT}, and their variants achieving state-of-the art results~\citep{bert,gpt_2,gpt3}. One reason which explains the success of SSL in NLP is the abundance of unlabeled text data, such as books, online websites, and blogs~\citep{wiki_data,reddit_data}, which prompted researchers to investigate SSL over supervised training. Another reason that explains their success, as discussed by~\citet{mocov1}, is the fundamental difference between the signal space of NLP and the signal space of computer vision, given that language data are discrete and structured (i.e., words), whereas image data are high dimensional, continuous, and unstructured. Nevertheless, we can state that the success of SSL in the field of NLP prompted the computer vision community to put more investigative efforts into this learning paradigm.

In order to alleviate issues regarding label requirements, as well as to enable robust feature extraction, self-supervised learning in computer vision emerged as a method for extracting robust features from unlabeled data using the properties of images themselves~\citep{mocov1,simclr}. The idea behind SSL is straightforward: devise an experimental setting in which the task that provides the supervisory signal can be solved without human annotation and then train DNNs to solve it.

Note that the description provided above for SSL also covers a number of additional approaches including autoencoders~\citep{ssl_autoencoder}, generative models, and clustering-based methods that leverage self-labeling~\citep{deepc}, and that these approaches also fall into the category of unsupervised learning (since human annotation is not necessary). Furthermore, most of the training routines described in this manuscript also use the term \say{self-supervised learning} interchangeably with \say{representation learning} when supervision is provided by the data, while representation learning is described by~\citet{representation_learning} as \say{learning representations of the data that make it easier to extract useful information when building classifiers or other predictors}, irrespective of the supervisory nature of the learning methodology. So, how did \say{self-supervision} become such a popular term in recent years?

\textbf{Resurgence of the term \say{self-supervised learning} in computer vision}\,\textendash\,Beyond a number of niche use cases such as image colorization~\citep{colorization_ssl_1}, image inpainting~\citep{inpainting_1}, and puzzle-solvers~\citep{puzzle_embedding} that explicitly use self-supervision, the term \say{self-supervised learning} was previously not employed to describe many techniques. Furthermore, compared to other learning paradigms, the use of SSL was not popular until recently (see Figure~\ref{fig:ssl_interest}). In fact, research efforts that are now considered to be pioneers in SSL and that are used for SSL benchmarking, such as \texttt{Deep Cluster}~\citep{deepc}, \texttt{InstDist}~\citep{instdist}, \texttt{CPC}~\citep{cpc_infonce_1}, and \texttt{Local Aggregation}~
\citep{local_agg}, were published as unsupervised training methods, distancing themselves from SSL. 

The resurgence of interest in self-supervision and the re-branding of corresponding methodologies can be attributed to the popularization of the term by both authoritative researchers and tech giants in the field between $2018$ and $2020$~\citep{zisserman_ssl,gelato_ssl,microsoft_ssl,facebook_ssl,google_ssl,fastai_ssl}. The reason for this re-branding is straightforward: most of the tasks discussed above that fell under the banner of unsupervised learning were deemed misleading, since the training was not completely unsupervised. Instead, the supervision was provided by the data itself, without explicit human labeling~\citep{zisserman_ssl,lecun_ssl_facebook}. As a result of this re-branding, while most papers published before $2020$ use unsupervised learning to describe their work, those that are published after $2020$ use the description self-supervised learning, hence the conflict between the use of the two terms.

An interesting moment in this timeline, and the one that furthered the popularity of the term SSL, is the revision by Yann LeCun of his now-famous cake analogy from NeurIPS-16, during a talk he gave at ISSCC-19 and later at AAAI-20~\citep{lecun_ssl}: \say{If intelligence is a cake, the bulk of the cake is \sout{unsupervised} \textit{self-supervised} learning, the icing on the cake is supervised learning, and the cherry on the cake is reinforcement learning}~\citep{lecun_cake}. 

In summary, we can say that self-supervised learning refers to a recently popularized learning paradigm, encompassing predictive tasks where the supervisory signal is provided by the data, without relying on the explicit use of human labels.


\textbf{Generative and discriminative SSL}\,\textendash\,In general, self-supervision approaches can be grouped into two categories: generative and discriminative~\citep{ssl_context_pred}. In generative self-supervision, the task is to build appropriate distributions over a collection of data while operating in the pixel space. A common criticism of generative self-supervision is that it is computationally expensive, does not work well with high-resolution images, and that it may be superfluous for representation learning~\citep{simclr,byol}. Typical models relying on this kind of self-supervision are autoencoders (AEs) and generative adversarial networks (GANs)~\citep{vae,autoencoders,gan}. It should be noted that although both AEs and GANs are categorized as \say{generative} models, they achieve self-supervision in different and distinct ways.

In contrast to generative SSL, in discriminative self-supervision, the task is to learn good representations of the data in order to perform a specified pretext task (which we will explain shortly) that does not require a human annotation effort~\citep{ssl_context_pred}. Discriminative self-supervision is similar to supervised learning in the sense that the objective function is often a scoring function that evaluates the discriminative power of learned representations. Most of the SSL frameworks we will cover in this manuscript refer to the works of~\cite{ssl_early} and~\citet{siamese} as the earliest research efforts that use discriminative self-supervision in the form it is used nowadays, with the above research efforts investigating representation alignments across different inputs.

\textbf{Purpose of this survey}\,\textendash\,Thanks to the excellent results obtained by SSL in computer vision, numerous SSL frameworks were proposed within the span of a couple of years. Although most of these frameworks are often specialized in nature, addressing a select number of tasks (such as depth estimation, face recognition, remote sensing, and pose estimation), we could trace their origin to roughly $100$ general-purpose SSL frameworks that are applicable to images. Even though several in-depth surveys are available on the topic of image-based contrastive SSL~\citep{albelwi2022_survey,khan2022contrastive_survey}, due to the fast-paced nature of research in SSL, they do not cover recent non-contrastive SSL methods that transformed the field. As such, a major goal of this survey is to cover the aforementioned image-oriented frameworks for generative and discriminative SSL which benefited from a tremendous research and development efforts in recent years, hereby presenting a concise and aggregate work to readers who take an interest in this field. 

In Section~\ref{sec:pretext_task}, we describe popular pretext tasks for self-supervision, subsequently detailing a number of relevant technical concepts that are commonly used in Section~\ref{sec:important_concepts}. Diving deeper into SSL as it is used nowadays for image-related tasks, in Section~\ref{sec:disc_ssl}, we cover recently proposed SSL frameworks for image-based training in a chronological order and discuss methods of evaluation in Section~\ref{sec:evaluation}. In Section~\ref{sec:code}, we cover relevant libraries, repositories, and publicly available implementations that aim at assisting researchers. Finally, in Section~\ref{sec:conclusions}, we review a number of shortcomings of SSL, identify open problems, and conclude our survey.

\begin{figure}[t!]
\centering
\begin{subfigure}{.32\textwidth}
\begin{tikzpicture}[thick,scale=0.68, every node/.style={scale=0.68}]
\centering
\def\x{0} 
\def\y{0}
\def\h{2}
\def\w{0.7}
\draw[draw=black] (\x, \y) -- (\x+\w, \y) -- (\x+\w,  \y+\h) -- (\x,  \y+\h) -- (\x,  \y);
\node[align=center, rotate=90] at (\x+\w/2, \y+\h/2)  {\footnotesize DNN};
\node[align=center] (net_left) at (\x+\w*0.1, \y+\h/2)  {};
\node[align=center] (net_right) at (\x+\w*0.9, \y+\h/2)  {};
\node[inner sep=0pt] (im2) at (\x+\w+2, \y+\h/2) {\includegraphics[width=2.4cm, frame]{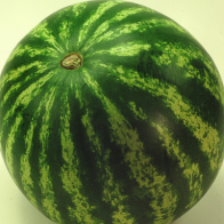}};
\node[inner sep=0pt] (im1) at (\x-2, \y+\h/2)    {\includegraphics[width=2.4cm, frame]{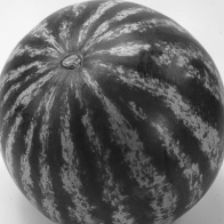}};
\path [->] (im1.east) edge node[below left] {} (net_left);
\path [->] (net_right) edge node[below left] {} (im2.west);
\end{tikzpicture}
\caption{Colorization}
\label{fig:pre_colorization}
\end{subfigure}
\begin{subfigure}{.32\textwidth}
\begin{tikzpicture}[thick,scale=0.68, every node/.style={scale=0.68}]
\centering
\def\x{0} 
\def\y{0}
\def\h{2}
\def\w{0.7}
\draw[draw=black] (\x, \y) -- (\x+\w, \y) -- (\x+\w,  \y+\h) -- (\x,  \y+\h) -- (\x,  \y);
\node[align=center, rotate=90] at (\x+\w/2, \y+\h/2)  {\footnotesize DNN};
\node[align=center] (net_left) at (\x+\w*0.1, \y+\h/2)  {};
\node[align=center] (net_right) at (\x+\w*0.9, \y+\h/2)  {};
\node[inner sep=0pt] (im2) at (\x+\w+2, \y+\h/2) {\includegraphics[width=2.4cm, frame]{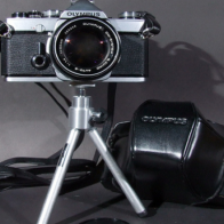}};
\node[inner sep=0pt] (im1) at (\x-2, \y+\h/2)    {\includegraphics[width=2.4cm, frame]{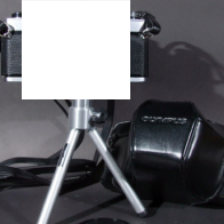}};
\path [->] (im1.east) edge node[below left] {} (net_left);
\path [->] (net_right) edge node[below left] {} (im2.west);
\end{tikzpicture}
\caption{Inpainting}
\label{fig:inpainting}
\end{subfigure}
\begin{subfigure}{.32\textwidth}
\begin{tikzpicture}[thick,scale=0.68, every node/.style={scale=0.68}]
\centering
\def\x{0} 
\def\y{0}
\def\h{2}
\def\w{0.7}
\draw[draw=black] (\x, \y) -- (\x+\w, \y) -- (\x+\w,  \y+\h) -- (\x,  \y+\h) -- (\x,  \y);
\node[align=center, rotate=90] at (\x+\w/2, \y+\h/2)  {\footnotesize DNN};
\node[align=center] (net_left) at (\x+\w*0.1, \y+\h/2)  {};
\node[align=center] (net_right) at (\x+\w*0.9, \y+\h/2)  {};
\node[inner sep=0pt] (im2) at (\x+\w+2, \y+\h/2) {\includegraphics[width=2.4cm, frame]{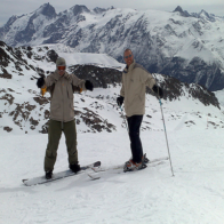}};
\node[inner sep=0pt] (im1) at (\x-2, \y+\h/2)    {\includegraphics[width=2.4cm, frame]{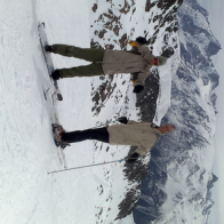}};
\path [->] (im1.east) edge node[below left] {} (net_left);
\path [->] (net_right) edge node[below left] {} (im2.west);
\end{tikzpicture}
\caption{Geometric transformations}
\label{fig:pre_geometric}
\end{subfigure}
\\
\vspace{2em} 
\begin{subfigure}{.32\textwidth}
\begin{tikzpicture}[thick,scale=0.68, every node/.style={scale=0.68}]
\centering
\def\x{0} 
\def\y{0}
\def\h{2}
\def\w{0.7}
\draw[draw=black] (\x, \y) -- (\x+\w, \y) -- (\x+\w,  \y+\h) -- (\x,  \y+\h) -- (\x,  \y);
\node[align=center, rotate=90] at (\x+\w/2, \y+\h/2)  {\footnotesize DNN};
\node[align=center] (net_left) at (\x+\w*0.1, \y+\h/2)  {};
\node[align=center] (net_right) at (\x+\w*0.9, \y+\h/2)  {};
\node[inner sep=0pt] (im1) at (\x-1.2, \y+\h/2+0.8) {\includegraphics[width=0.8cm, frame]{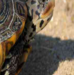}};
\node[inner sep=0pt] (im1right) at (\x-1.2, \y+\h/2) {\includegraphics[width=0.8cm, frame]{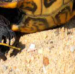}};
\node[inner sep=0pt] (im1) at (\x-1.2, \y+\h/2-0.8) {\includegraphics[width=0.8cm, frame]{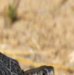}};

\node[inner sep=0pt] (im1) at (\x-2, \y+\h/2+0.8) {\includegraphics[width=0.8cm, frame]{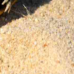}};
\node[inner sep=0pt] (im1) at (\x-2, \y+\h/2) {\includegraphics[width=0.8cm, frame]{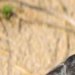}};
\node[inner sep=0pt] (im1) at (\x-2, \y+\h/2-0.8) {\includegraphics[width=0.8cm, frame]{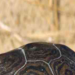}};

\node[inner sep=0pt] (im1) at (\x-2.8, \y+\h/2+0.8) {\includegraphics[width=0.8cm, frame]{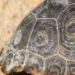}};
\node[inner sep=0pt] (im1) at (\x-2.8, \y+\h/2) {\includegraphics[width=0.8cm, frame]{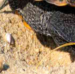}};
\node[inner sep=0pt] (im1) at (\x-2.8, \y+\h/2-0.8) {\includegraphics[width=0.8cm, frame]{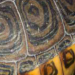}};

\node[inner sep=0pt] (im2) at (\x+\w+2, \y+\h/2) {\includegraphics[width=2.4cm, frame]{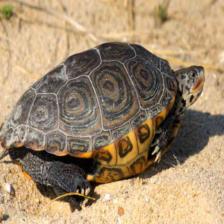}};
\path [->] (im1right.east) edge node[below left] {} (net_left);
\path [->] (net_right) edge node[below left] {} (im2.west);
\end{tikzpicture}
\caption{Puzzle solvers}
\label{fig:jigsaw}
\end{subfigure}
\begin{subfigure}{.32\textwidth}
\begin{tikzpicture}[thick,scale=0.68, every node/.style={scale=0.68}]
\centering
\def\x{0} 
\def\y{0}
\def\h{2}
\def\w{0.7}
\node[align=center] (balancing) at (\x-3.1, \y+\h/2)  {};  
\draw[draw=black] (\x, \y) -- (\x+\w, \y) -- (\x+\w,  \y+\h) -- (\x,  \y+\h) -- (\x,  \y);
\node[align=center, rotate=90] at (\x+\w/2, \y+\h/2)  {\footnotesize DNN};
\node[align=center] (net_left) at (\x+\w*0.1, \y+\h/2)  {};
\node[align=center] (net_right) at (\x+\w*0.9, \y+\h/2)  {};
\node[inner sep=0pt] (im1) at (\x-1.4, \y+\h/2+0.6) {\includegraphics[width=1.1cm, frame]{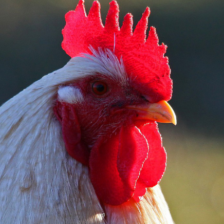}};
\node[inner sep=0pt] (im1) at (\x-1.4, \y+\h/2-0.6) {\includegraphics[width=1.1cm, frame]{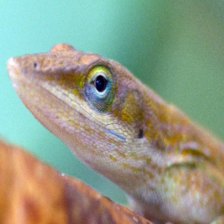}};
\node[inner sep=0pt] (im1) at (\x-2.6, \y+\h/2) {\includegraphics[width=1.1cm, frame]{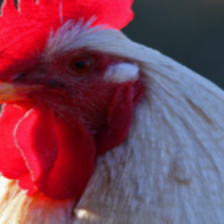}};

\node[inner sep=0pt] (im21) at (\x+\w+1.2, \y+\h/2+0.8) {\includegraphics[width=0.75cm, frame]{ims/1.png}};
\node[inner sep=0pt] (im23) at (\x+\w+1.2, \y+\h/2) {\includegraphics[width=0.75cm, frame]{ims/4.png}};
\node[inner sep=0pt] (im22) at (\x+\w+1.2, \y+\h/2-0.8) {\includegraphics[width=0.75cm, frame]{ims/7.png}};

\path [<->, red] (im21.east) edge[bend left=20] node {\phantom{-}} (im22.east);
\path [<->, red] (im23.east) edge[bend left=20] node {\phantom{-}} (im22.east);
\path [<->, blue] (im21.east) edge[bend left=80] node {\phantom{-}} (im23.east);

\node[align=center] at (\x+\w+2.8, \y+\h/2+0.4)  {\footnotesize \blu{Pull together}};
\node[align=center] at (\x+\w+2.5, \y+\h/2-0.4)  {\footnotesize \red{Contrast}};

\path [->] (im1right.east) edge node[below left] {} (net_left);
\path [->] (net_right) edge node[below left] {} (im2.west);
\end{tikzpicture}
\caption{Instance discrimination}
\label{fig:instance_disc}
\end{subfigure}
\begin{subfigure}{.32\textwidth}
\begin{tikzpicture}[thick,scale=0.68, every node/.style={scale=0.68}]
\centering
\def\x{0} 
\def\y{0}
\def\h{2}
\def\w{0.7}
\node[align=center] (balancing) at (\x-3.1, \y+\h/2)  {};  
\draw[draw=black] (\x, \y) -- (\x+\w, \y) -- (\x+\w,  \y+\h) -- (\x,  \y+\h) -- (\x,  \y);
\node[align=center, rotate=90] at (\x+\w/2, \y+\h/2)  {\footnotesize DNN};
\node[align=center] (net_left) at (\x+\w*0.1, \y+\h/2)  {};
\node[align=center] (net_right) at (\x+\w*0.9, \y+\h/2)  {};
\node[inner sep=0pt] (im1) at (\x-1.4, \y+\h/2+0.635) {\includegraphics[width=1.1cm, frame]{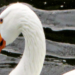}};
\node[inner sep=0pt] (im1) at (\x-1.4, \y+\h/2-0.635) {\includegraphics[width=1.1cm, frame]{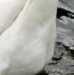}};

\node[inner sep=0pt] (im2) at (\x+\w+2, \y+\h/2) {\includegraphics[width=2.4cm, frame]{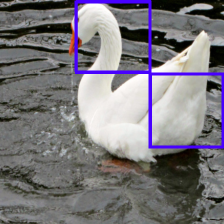}};

\path [->] (im1right.east) edge node[below left] {} (net_left);
\path [->] (net_right) edge node[below left] {} (im2.west);
\end{tikzpicture}
\caption{Context prediction}
\label{fig:context_pred}
\end{subfigure}
\\
\vspace{2em} 
\begin{subfigure}{.32\textwidth}
\begin{tikzpicture}[thick,scale=0.68, every node/.style={scale=0.68}]
\centering
\def\x{0} 
\def\y{0}
\def\h{2}
\def\w{0.7}
\draw[draw=black] (\x, \y) -- (\x+\w, \y) -- (\x+\w,  \y+\h) -- (\x,  \y+\h) -- (\x,  \y);
\node[align=center, rotate=90] at (\x+\w/2, \y+\h/2)  {\footnotesize DNN};
\node[align=center] (net_left) at (\x+\w*0.1, \y+\h/2)  {};
\node[align=center] (net_right) at (\x+\w*0.9, \y+\h/2)  {};
\node[inner sep=0pt] (im2) at (\x+\w+2, \y+\h/2) {\includegraphics[width=2.4cm, frame]{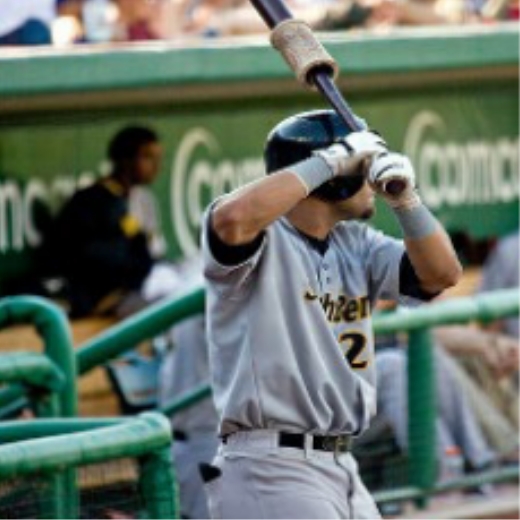}};
\node[inner sep=0pt] (im1) at (\x-2, \y+\h/2)    {\includegraphics[width=2.4cm, frame]{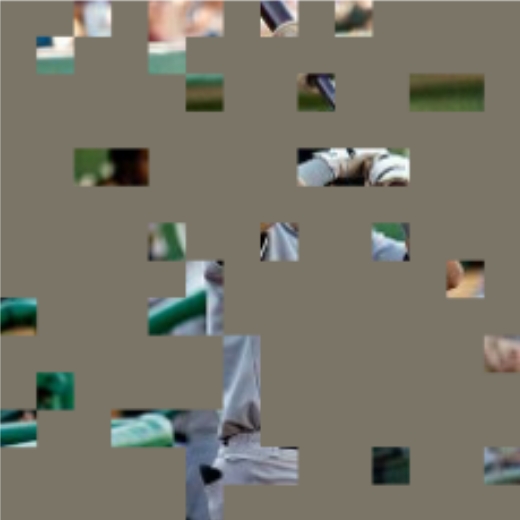}};
\path [->] (im1.east) edge node[below left] {} (net_left);
\path [->] (net_right) edge node[below left] {} (im2.west);
\end{tikzpicture}
\caption{Masked image modeling}
\label{fig:mim}
\end{subfigure}
\begin{subfigure}{.32\textwidth}
\begin{tikzpicture}[thick,scale=0.68, every node/.style={scale=0.68}]
\centering
\def\x{0} 
\def\y{0}
\def\h{2}
\def\w{0.7}
\draw[draw=black] (\x, \y) -- (\x+\w, \y) -- (\x+\w,  \y+\h) -- (\x,  \y+\h) -- (\x,  \y);
\node[align=center, rotate=90] at (\x+\w/2, \y+\h/2)  {\footnotesize DNN};
\node[align=center] (net_left) at (\x+\w*0.1, \y+\h/2)  {};
\node[align=center] (net_right) at (\x+\w*0.9, \y+\h/2)  {};
\node[inner sep=0pt] (im2) at (\x+\w+2, \y+\h/2) {\includegraphics[width=2.4cm, frame]{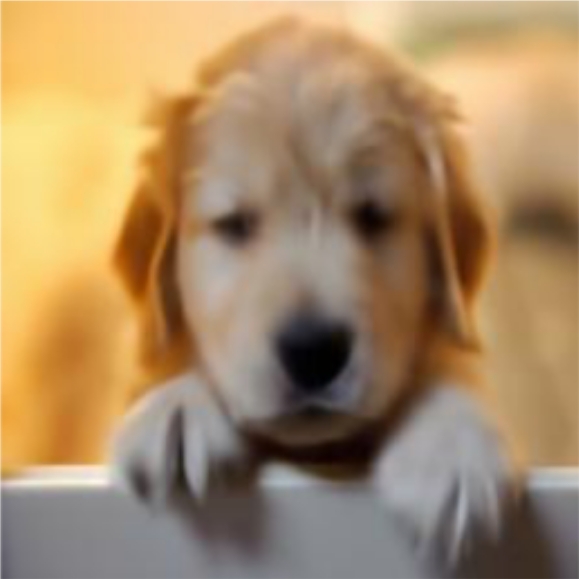}};
\node[inner sep=0pt] (im1) at (\x-2, \y+\h/2)    {\includegraphics[width=2.4cm, frame]{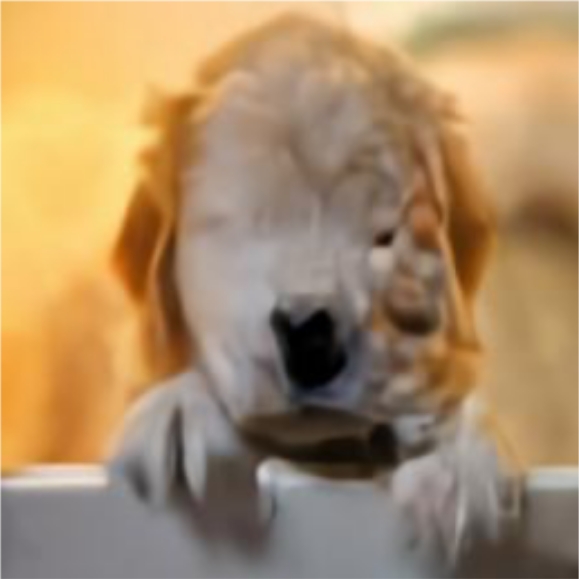}};
\path [->] (im1.east) edge node[below left] {} (net_left);
\path [->] (net_right) edge node[below left] {} (im2.west);
\end{tikzpicture}
\caption{Corrupted image modeling}
\label{fig:cim}
\end{subfigure}
\begin{subfigure}{.32\textwidth}
\begin{tikzpicture}[thick,scale=0.68, every node/.style={scale=0.68}]
\centering
\def\x{0} 
\def\y{0}
\def\h{2}
\def\w{0.7}
\draw[draw=black] (\x, \y) -- (\x+\w, \y) -- (\x+\w,  \y+\h) -- (\x,  \y+\h) -- (\x,  \y);
\node[align=center, rotate=90] at (\x+\w/2, \y+\h/2)  {\footnotesize DNN};
\node[align=center] (net_left) at (\x+\w*0.1, \y+\h/2)  {};
\node[align=center] (net_right) at (\x+\w*0.9, \y+\h/2)  {};
\node[inner sep=0pt] (im2) at (\x+\w+2, \y+\h/2) {\includegraphics[width=2.4cm, frame]{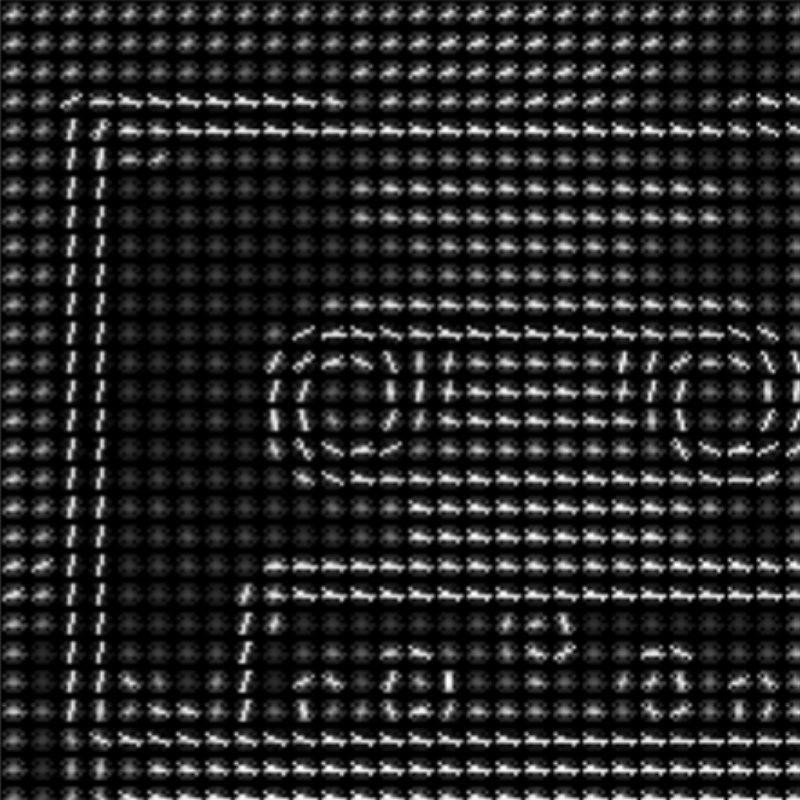}};
\node[inner sep=0pt] (im1) at (\x-2, \y+\h/2)    {\includegraphics[width=2.4cm, frame]{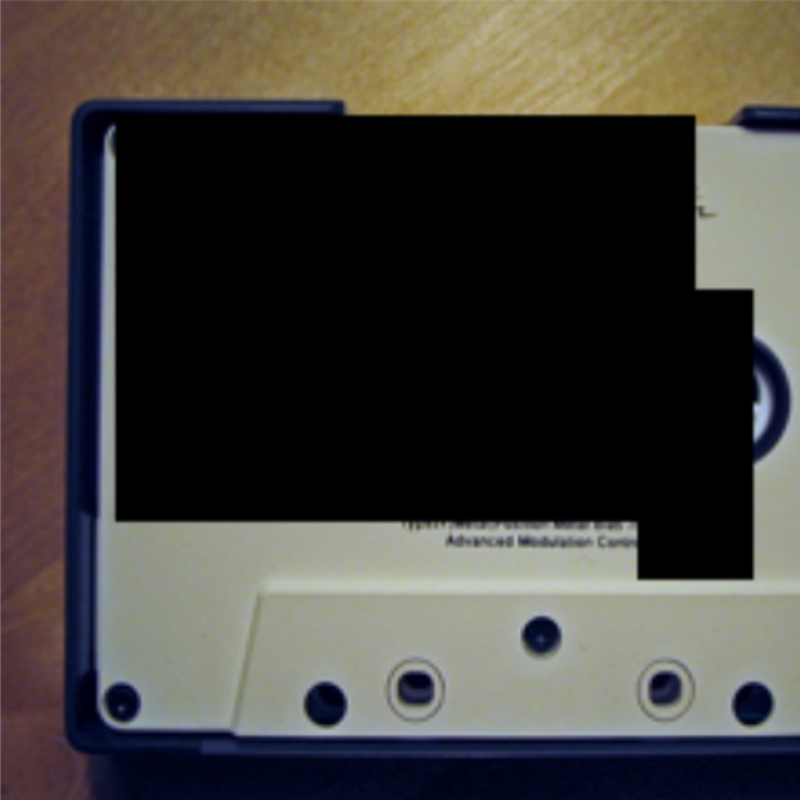}};
\path [->] (im1.east) edge node[below left] {} (net_left);
\path [->] (net_right) edge node[below left] {} (im2.west);
\end{tikzpicture}
\caption{Masked feature prediction}
\label{fig:mfp}
\end{subfigure}

\caption{Illustrations of various image-based pretext tasks for self-supervised learning.}
\label{fig:pretext_task}
\vspace{-1em}
\end{figure}

\section{Pretext tasks for self-supervised learning}
\label{sec:pretext_task}

The image domain allows a number of unique pretext tasks that enable self-supervision. Below we describe the most popular ones and illustrate them in Figure~\ref{fig:pretext_task}.

\textbf{Image colorization}\,\textendash\,Automated colorization of grayscale images is a line of research that was investigated even before the widespread usage of DNNs~\citep{image_colorization_no_dnn_1,image_colorization_no_dnn_2}. However, the availability of large-scale colored datasets such as ImageNet, combined with the versatility of DNNs, further strengthened the interest in high-quality image colorization, especially for the purpose of coloring historical pictures. In parallel to research efforts that aimed at increasing the quality of colorization, such as~\citep{improve_coloring_1,improve_coloring_2}, the idea of using image colorization as a pretext task for representation learning was also investigated~\citep{colorization_ssl_1,colorization_ssl_2,colorization_ssl_3}. Although this task alone was revealed to be too simple to force DNNs to learn complex representations~\citep{swav}, colorization is still used in tandem with other tasks to boost the effectiveness of SSL models.

\textbf{Inpainting}\,\textendash\,The task of predicting a missing part of an image is referred to as image inpainting~\citep{bertalmio2000image}. With the widespread usage of DNNs, inpainting problems also found numerous solutions~\citep{inpainting_1,inpainting_2}. One such solution, and the one that allows for the use of SSL, is proposed by~\citet{inpainting_ssl}, leveraging context encoders that aim at inpainting large parts of images that are missing, forcing models to learn the image context.

\textbf{Geometric transformations}\,\textendash\,Inspired by research efforts that bring together geometric transformations and neural networks~\citep{ssl_geometry_1,ssl_geometry_2}, and taking advantage of image-based datasets that almost always contain upright images,~\citet{rotation_ssl} proposed the idea of predicting image rotations as a method of self-supervision. Following the success of this method, other types of geometric transformations were proposed by~\citet{ssl_geometry_3,ssl_geometry_4,ss_gan}.

\textbf{Puzzle solvers}\,\textendash\,A unique image-based task that can be formulated in a SSL setting is solving a jigsaw puzzle~\citep{puzzle_ssl}, where the goal is to correctly predict the relative location of nine puzzle pieces. This unusual pretext task, as well as a number of derivations, is employed in support of a variety of tasks, including domain generalization~\citep{puzzle_generalization}, generation of image embeddings~\citep{puzzle_embedding}, image retrieval~\citep{puzzle_retrieval}, and auxiliary learning~\citep{puzzle_gan}.

\textbf{Instance discrimination}\,\textendash\,Given differently augmented views (i.e., instances) originating from one image, instance discrimination refers to the idea of recognizing these views as originating from the same image, while discriminating any other image with a different origin~\citep{instdist}. Different from the previously described pretext tasks which achieve representation learning as a by-product of the optimization objective, instance discrimination optimizes for representation learning by directly matching the representations of similar images while contrasting the representations of dissimilar ones. In this context, images that are contrasted to similar ones are called negative samples (e.g., the gecko image in Figure~\ref{fig:instance_disc}). The main idea behind representation matching between similar images and contrasting different images is to help DNNs learn representations that are invariant to commonly used image transformations, since most of these transformations do not alter the visual semantics~\citep{pirl}. The origins of this approach can be traced back to the research efforts presented in~\citet{early_contrastive_1}, \citet{ssl_discriminative_2}, and \citet{ssl_discriminative_3}.

\textbf{Masked image modeling}\,\textendash\,The adaptation of masked language modeling in NLP to computer vision as a new pretext task for self-supervised training was a groundbreaking discovery in generative SSL~\citep{igpt,beit}. This technique is referred to as masked image modeling (MIM)~\citep{beit}. The idea behind MIM is simple: divide an image into a collection of equal-sized patches, mask some of the patches, and task the model with generating their corresponding pattern. As we will discuss in later parts of this paper, while the usage of MIM has popularized generative SSL, this pretext task can be thought of as a variant of image inpainting. The primary difference between MIM and the inpainting method proposed in the work of ~\citet{inpainting_ssl} is that MIM uses non-overlapping patches of equal size. After the rise in popularity of MIM (which is often used in conjunction with vision transformers (ViT)~\citep{vit}), a number of its variants emerged, with corrupted image modeling~\citep{cim} (see Figure~\ref{fig:cim}) and masked feature prediction~\citep{masked_feature_modeling} (see Figure~\ref{fig:mfp}) the two most prominent ones.

\textbf{Others}\,\textendash\,Apart from the mainstream pretext tasks described above, there are a number of unique tasks that do not fit into one of the above categories such as: the split-brain approach which tries to predict a subset of image channels from other channels~\citep{ssl_split_brain}, a feature consistency method involving synthetic images~\citep{ssl_syntethic}, context prediction~\citep{ssl_context_pred}, adversarial feature learning~\citep{bigan,bigbigan}, exemplar networks~\citep{ssl_discriminative_1}, and object counting~\citep{ssl_counting}.

\textbf{Effectiveness of pretext tasks}\,\textendash\,Given the abundance of pretext tasks for self-supervision, which of these tasks enable DNNs to learn the most useful representations? Although there is no clear answer to this question, ever since the works of~\citet{ssl_discriminative_1}, \citet{instdist}, and \citet{cpc_infonce_1}, instance discrimination was established as the dominant pretext task for image-based discriminative SSL, thanks to the superb results achieved using this type of self-supervision~\citep{mocov1,byol,simclr}. On the other hand, MIM has been recognized as a tremendously powerful pretext task that enables generative SSL to reach and even surpass the results obtained with instance discrimination.~\citep{vit,beit,mae}.







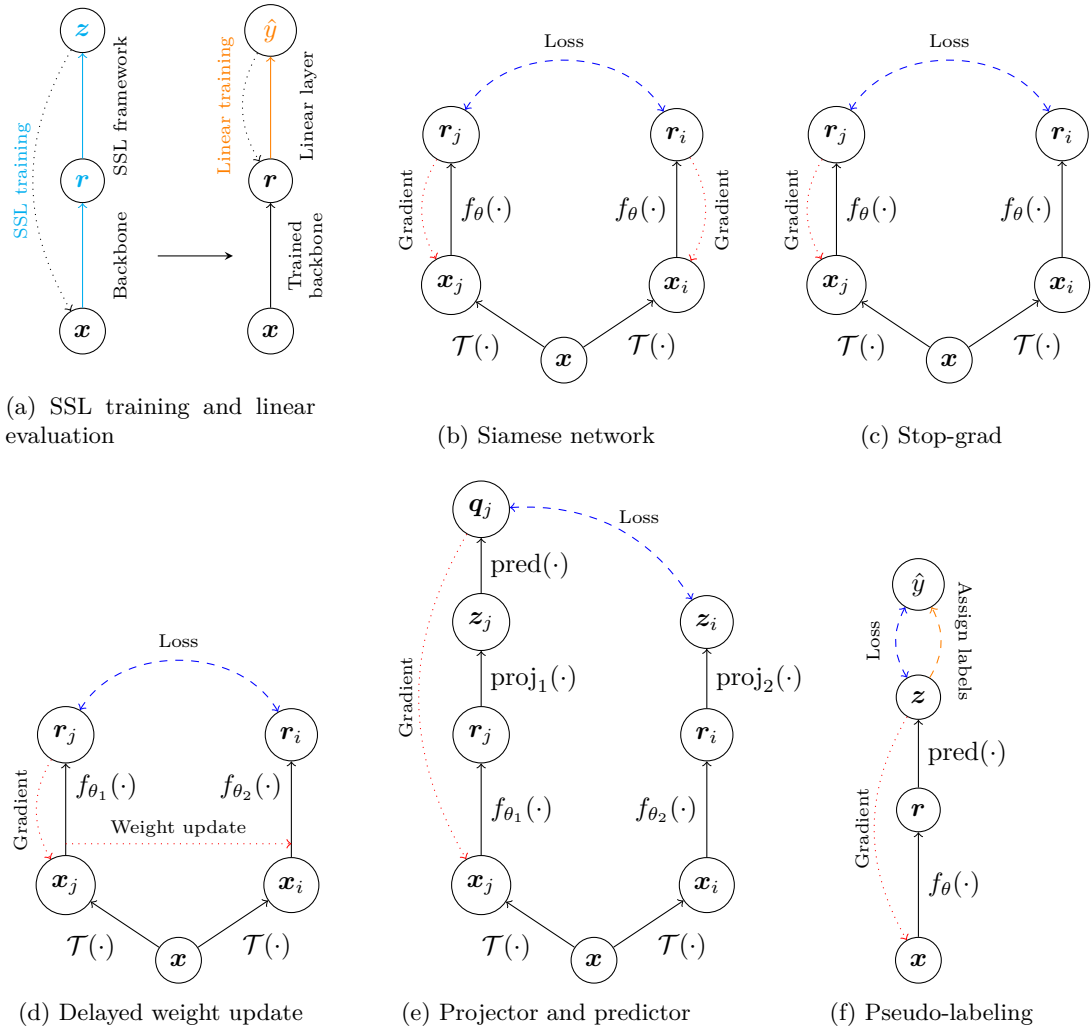
\begin{figure}[t!]
\centering
\begin{subfigure}{.25\textwidth}
\centering
\begin{tikzpicture}
\node [] at (0, -0.5) (alignment_node) {};
\node [] at (0, 4.5) (alignment_node) {};
\def\x{0}

\node[shape=circle,draw=black] (B1) at (\x,0) {$\bm{x}$};
\node[shape=circle,draw=black] (C1) at (\x,2) {\color{cyan}{$\bm{r}$}};
\path [->, cyan] (B1) edge node[right] {} (C1);
\node [rotate=90] at (\x + 0.5, 1) (backbone) {\scriptsize $\text{Backbone}$};
\node[shape=circle,draw=black] (D1) at (\x, 4) {\color{cyan}{$\bm{z}$}};
\path [->, cyan] (C1) edge node[right] {} (D1);
\node [rotate=90] at (\x + 0.5, 3) (ssl_framework) {\scriptsize $\text{SSL framework}$};
\path [->, black] (D1) edge[dotted, bend right] node {\phantom{-}} (B1);
\node [rotate=90, cyan] at (\x - 0.8, 2) (Gradient1) {\scriptsize SSL training};

\def\x{2.5}

\draw [-stealth] (1,1) -- (2,1);
\node[shape=circle,draw=black] (T1) at (\x,0) {$\bm{x}$};
\node[shape=circle,draw=black] (Y1) at (\x,2) {$\bm{r}$};
\path [->] (T1) edge node[right] {} (Y1);
\node [rotate=90] at (\x + 0.32, 1) (backbone) {\scriptsize $\text{Trained}$};
\node [rotate=90] at (\x + 0.6, 1) (backbone) {\scriptsize $\text{backbone}$};
\node[shape=circle,draw=black] (U1) at (\x, 4) {\color{orange}{$\hat{y}$}};
\path [->, orange] (Y1) edge node[right] {} (U1);
\node [rotate=90] at (\x + 0.5, 3) (ssl_framework) {\scriptsize $\text{Linear layer}$};
\path [->, black] (U1) edge[dotted, bend right] node {\phantom{-}} (Y1);
\node [rotate=90, orange] at (\x - 0.6, 3) (Gradient1) {\scriptsize Linear training};
\end{tikzpicture}
\caption{SSL training and linear evaluation}
\label{fig:ssl_arch}
\end{subfigure}
\hspace{2.5em}
\begin{subfigure}{.25\textwidth}
\centering
\begin{tikzpicture}
\node [] at (0, -0.5) (alignment_node) {};
\node [] at (0, 4.5) (alignment_node) {};
\def\x{0}
\node[shape=circle,draw=black] (A) at (\x, 0) {$\bm{x}$};

\node[shape=circle,draw=black] (B2) at (\x + 1.5,1) {$\bm{x}_i$};
\node[shape=circle,draw=black] (B1) at (\x - 1.5,1) {$\bm{x}_j$};

\path [->] (A) edge node[below right] {$\mathcal{T}(\cdot)$} (B2);
\path [->] (A) edge node[below left] {$\mathcal{T}(\cdot)$} (B1);

\node[shape=circle,draw=black] (C2) at (\x + 1.5,3) {$\bm{r}_i$};
\node[shape=circle,draw=black] (C1) at (\x - 1.5,3) {$\bm{r}_j$};

\path [->] (B2) edge node[left] {$f_\theta(\cdot)$} (C2);
\path [->] (B1) edge node[right] {$f_\theta(\cdot)$} (C1);

\path [<->, blue] (C1) edge[dashed, bend left=60] node {\phantom{-}} (C2);
\node at (\x, 4.25) (Loss) {\scriptsize Loss};

\path [->, red] (C2) edge[dotted, bend left] node {\phantom{-}} (B2);
\path [->, red] (C1) edge[dotted, bend right] node {\phantom{-}} (B1);
\node [rotate=90] at (\x + 2.1, 2) (Gradient1) {\scriptsize Gradient};
\node [rotate=90] at (\x - 2.1, 2) (Gradient1) {\scriptsize Gradient};
\end{tikzpicture}
\caption{Siamese network}
\label{fig:siamese}
\end{subfigure}
\hspace{2.5em}
\begin{subfigure}{.25\textwidth}
\centering
\begin{tikzpicture}
\node [] at (0, -0.5) (alignment_node) {};
\node [] at (0, 4.5) (alignment_node) {};
\def\x{0}
\node[shape=circle,draw=black] (A) at (\x, 0) {$\bm{x}$};

\node[shape=circle,draw=black] (B2) at (\x + 1.5,1) {$\bm{x}_i$};
\node[shape=circle,draw=black] (B1) at (\x - 1.5,1) {$\bm{x}_j$};

\path [->] (A) edge node[below right] {$\mathcal{T}(\cdot)$} (B2);
\path [->] (A) edge node[below left] {$\mathcal{T}(\cdot)$} (B1);

\node[shape=circle,draw=black] (C2) at (\x + 1.5,3) {$\bm{r}_i$};
\node[shape=circle,draw=black] (C1) at (\x - 1.5,3) {$\bm{r}_j$};

\path [->] (B2) edge node[left] {$f_\theta(\cdot)$} (C2);
\path [->] (B1) edge node[right] {$f_\theta(\cdot)$} (C1);

\path [<->, blue] (C1) edge[dashed, bend left=60] node {\phantom{-}} (C2);
\node at (\x, 4.25) (Loss) {\scriptsize Loss};

\path [->, red] (C1) edge[dotted, bend right] node {\phantom{-}} (B1);
\node [rotate=90] at (\x - 2.1, 2) (Gradient1) {\scriptsize Gradient};
\end{tikzpicture}
\caption{Stop-grad}
\label{fig:stop-grad}
\end{subfigure}
\\ 
\vspace{1em}
\begin{subfigure}{.25\textwidth}
\centering
\begin{tikzpicture}
\node [] at (0, 0) (alignment_node) {};
\node [] at (0, 6) (alignment_node) {};
\def\x{0}
\node[shape=circle,draw=black] (A) at (\x, 0) {$\bm{x}$};

\node[shape=circle,draw=black] (B2) at (\x + 1.5,1) {$\bm{x}_i$};
\node[shape=circle,draw=black] (B1) at (\x - 1.5,1) {$\bm{x}_j$};

\path [->] (A) edge node[below right] {$\mathcal{T}(\cdot)$} (B2);
\path [->] (A) edge node[below left] {$\mathcal{T}(\cdot)$} (B1);

\node[shape=circle,draw=black] (C2) at (\x + 1.5,3) {$\bm{r}_i$};
\node[shape=circle,draw=black] (C1) at (\x - 1.5,3) {$\bm{r}_j$};

\path [->] (B2) edge node[above left] {$f_{\theta_2}(\cdot)$} (C2);
\path [->] (B1) edge node[above right] {$f_{\theta_1}(\cdot)$} (C1);

\path [<->, blue] (C1) edge[dashed, bend left=60] node {\phantom{-}} (C2);
\node at (\x, 4.25) (Loss) {\scriptsize Loss};

\path [->, red] (C1) edge[dotted, bend right] node {\phantom{-}} (B1);
\node [rotate=90] at (\x - 2.1, 2) (Gradient1) {\scriptsize Gradient};

\draw [dotted,->, red] (-1.5,1.55) -- (1.5,1.55);
\node at (0,1.75) (Loss) {\scriptsize Weight update};

\end{tikzpicture}
\caption{Delayed weight update}
\label{fig:delayed_weight_update}
\end{subfigure}
\hspace{2.5em}
\begin{subfigure}{.25\textwidth}
\centering
\begin{tikzpicture}
\node [] at (0, 0) (alignment_node) {};
\node [] at (0, 6) (alignment_node) {};
\def\x{0}
\node[shape=circle,draw=black] (A) at (\x, 0) {$\bm{x}$};

\node[shape=circle,draw=black] (B2) at (\x + 1.5,1) {$\bm{x}_i$};
\node[shape=circle,draw=black] (B1) at (\x - 1.5,1) {$\bm{x}_j$};

\path [->] (A) edge node[below right] {$\mathcal{T}(\cdot)$} (B2);
\path [->] (A) edge node[below left] {$\mathcal{T}(\cdot)$} (B1);

\node[shape=circle,draw=black] (C2) at (\x + 1.5,3) {$\bm{r}_i$};
\node[shape=circle,draw=black] (C1) at (\x - 1.5,3) {$\bm{r}_j$};

\path [->] (B2) edge node[left] {$f_{\theta_2}(\cdot)$} (C2);
\path [->] (B1) edge node[right] {$f_{\theta_1}(\cdot)$} (C1);

\node[shape=circle,draw=black] (D2) at (\x + 1.5, 4.5) {$\bm{z}_i$};
\path [->] (C2) edge node[right] {$\text{proj}_2(\cdot)$} (D2);
\node[shape=circle,draw=black] (D1) at (\x - 1.5, 4.5) {$\bm{z}_j$};
\path [->] (C1) edge node[right] {$\text{proj}_1(\cdot)$} (D1);
\node[shape=circle,draw=black] (E1) at (\x - 1.5, 6) {$\bm{q}_j$};
\path [->] (D1) edge node[right] {$\text{pred}(\cdot)$} (E1);

\path [<->, blue] (E1) edge[dashed, bend left=30] node {\phantom{-}} (D2);
\node at (\x+0.6, 5.9) (Loss) {\scriptsize Loss};

\path [->, red] (E1) edge[dotted, bend right] node {\phantom{-}} (B1);
\node [rotate=90] at (\x - 2.5, 3.5) (Gradient1) {\scriptsize Gradient};
\end{tikzpicture}
\caption{Projector and predictor}
\label{fig:proj_pred}
\end{subfigure}
\hspace{2.5em}
\begin{subfigure}{.25\textwidth}
\centering
\begin{tikzpicture}
\node [] at (0, 0) (alignment_node) {};
\node [] at (0, 6) (alignment_node) {};
\def\x{0}

\node[shape=circle,draw=black] (B1) at (\x,0) {$\bm{x}$};

\node[shape=circle,draw=black] (C1) at (\x,2) {$\bm{r}$};

\path [->] (B1) edge node[right] {$f_{\theta}(\cdot)$} (C1);

\node[shape=circle,draw=black] (D1) at (\x, 3.5) {$\bm{z}$};
\path [->] (C1) edge node[right] {$\text{pred}(\cdot)$} (D1);

\node[shape=circle,draw=black] (E1) at (\x, 5) {$\hat{y}$};

\path [->, orange] (D1) edge[dashed, bend right=30] node {\phantom{-}} (E1);
\path [<->, blue] (D1) edge[dashed, bend left=30] node {\phantom{-}} (E1);

\path [->, red] (D1) edge[dotted, bend right] node {\phantom{-}} (B1);
\node [rotate=90] at (\x - 0.75, 1.75) (Gradient1) {\scriptsize Gradient};

\node [rotate=90] at (\x - 0.6, 4.25) (Loss) {\scriptsize Loss};
\node [rotate=-90] at (\x + 0.6, 4.25) (Loss) {\scriptsize Assign labels};
\end{tikzpicture}
\caption{Pseudo-labeling}
\label{fig:pseudo_label}
\end{subfigure}
\caption{Illustrations of some of the important concepts related to SSL described in Section~\ref{sec:important_concepts}.}
\label{fig:important_concepts}
\vspace{-1em}
\end{figure}

\section{Important concepts in self-supervised learning}
\label{sec:important_concepts}

In this section, we briefly describe a number of commonly used concepts that are relevant to the forthcoming SSL frameworks. Although these concepts were key elements of early individual SSL frameworks, newer frameworks make use of a mixture of them. 


\textbf{Notation}\,\textendash\,For clarity, we briefly detail the notation used to describe several core SSL concepts. Given an image $\bm{x} \in \mathbb{R}^p$ and its categorical association $\bm{y} \in \mathbb{R}^M$ sampled from a dataset $(\bm{x}, \bm{y}) \sim \mathcal{D}$, with $y_c = 1$ and $y_m = 0 \,, \forall \, m \in \{0,\ldots, M\} \char`\\ \{c\}$, let $f_\theta(\cdot)$ be an encoder (i.e., a feature extractor) that maps an image augmented with a stochastic augmentation function $\mathcal{T}(\cdot)$ to a set of features $\bm{r} \in \mathbb{R}^k$ using a neural network with parameters $\theta$. These features can then be mapped onto a set of projections $\bm{z}$ and predictions $\bm{q}$ using the $\text{proj}(\cdot)$ and  $\text{pred}(\cdot)$ functions, respectively. In this context, projectors and predictors are simply multi-layer perceptrons (MLP).


\textbf{Backbone network}\,\textendash\,In the context of SSL, the term \say{backbone} refers to the feature extractor(s) (i.e., $f_\theta(\cdot)$) that are trained with SSL frameworks. Typically, a backbone network is a task-agnostic DNN (e.g., a ResNet-50 without the final fully connected layer). The majority of the frameworks we will cover use either a variant of ResNet (e.g., vanilla ResNet-50, ResNext, or Wide ResNet) or, very recently, vision transformers as the backbone.

\textbf{SSL training and evaluation}\,\textendash\,In traditional supervised learning, the feature extractor (e.g., convolutional layers) and the predictor (e.g., linear layers that map features to classes) are trained at the same time. However, SSL is only concerned with the training of the feature extractor. After the SSL training is complete, the linear layer that maps the features to classes is trained separately.

In Figure~\ref{fig:ssl_arch}, we provide a simplified illustration of (left) SSL training and (right) linear evaluation. SSL frameworks are placed on top of backbone networks and are trained in conjunction with the backbone. After the SSL training is complete, the framework is discarded and only the trained backbone is used. Note that this backbone is merely a feature extractor. Then, depending on the problem at hand, a new layer that maps features to classes is initialized and trained. It is crucial to keep in mind that the SSL training is only concerned with the quality of features obtained from the feature extractor. As such, the majority (if not the entirety) of the forthcoming concepts as well as frameworks tackle feature extractor training. Nevertheless, for the sake of completeness, in Section~\ref{sec:evaluation}, we will also describe evaluation methods.

\textbf{Vision transformers}\,\textendash\,Vision transformers represent a novel deep learning paradigm that leverages the transformer architecture developed initially for NLP and applies it to image classification tasks.

\begin{minipage}{0.25\textwidth}
\phantom{-}
\\
\includegraphics[width=\textwidth]{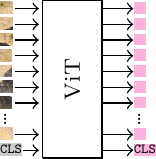}
\captionof{figure}{An illustration of ViT input-output relations.}
\label{fig:vit_architecture}
\end{minipage}
\begin{minipage}[t]{0.04\textwidth}  
\phantom{...........}  
\end{minipage}  
\begin{minipage}{0.71\textwidth}
ViT adopts a preprocessing step that involves partitioning the input image into non-overlapping patches, which are linearly embedded to create a sequence of tokens. The transformer encoder is then applied to these tokens, with the self-attention mechanism allowing the model to selectively focus on different patches and learn intricate correlation structures among them.
\vspace{0.5em} 
\\
An essential element of the ViT architecture is the \texttt{[CLS]} token, which is prepended to the input and subsequently leveraged for downstream classification tasks. However, in addition to the \texttt{[CLS]} token, ViTs also generate patch representation tokens that encapsulate information about the corresponding patch and its relationship with other patches, based on the attention mechanism (see Figure~\ref{fig:vit_architecture}). These representation tokens can be utilized for various MIM-based self-supervised tasks, which are relevant for generative SSL frameworks.
\end{minipage}

\textbf{Siamese networks}\,\textendash\,A form of dual-backbone networks called Siamese networks~\citep{siamese} consisting of two identical neural networks that share the same set of weights are popular architectures for SSL (see Figure~\ref{fig:siamese}). Although this type of networks was useful in solving a variety of problems~\citep{siamese_use3,siamese_use1,siamese_use2}, in the context of SSL, they are mostly employed to achieve consistency between representations when, for example, two instances of the same image are provided.

Apart from Siamese networks, a majority of SSL frameworks use dual backbones that may not share weights due to recently discovered beneficial properties. In such cases, the weights of one model are updated via backpropagation, while the weights of the other model can be updated using a variety of techniques which we discuss next.

\textbf{Stop-grad}\,\textendash\,Siamese networks generally propagate errors from both branches after the loss calculation. As illustrated in Figure~\ref{fig:stop-grad}, the term \say{stop-grad} refers to stopping the gradient flow from one branch of a dual-backbone network, while allowing this gradient flow to alter the weights of the other branch~\citep{simsiam}.

\textbf{Delayed weight updates}\,\textendash\,Assume a Siamese-like dual-backbone network where one branch is called the teacher and the other one the student. However, different than the Siamese architecture, weights of these models are not shared. In this scenario, delayed-weight updates refer to the idea of propagating the error through only one branch via backpropagation and updating the trainable parameters of the other branch via a predetermined rule (see Figure~\ref{fig:delayed_weight_update}). Popular implementations of this operation are \textit{Mean Teacher}~\citep{mean_teacher}, \textit{momentum encoding}~\citep{mocov1}, and \textit{exponential moving average}~\citep{byol}.

\textbf{Projection and prediction MLPs}\,\textendash\,The usage of multi-layer perceptrons in the form of projection and prediction heads following a feature extractor (e.g., a dual backbone) is acknowledged as a powerful technique that greatly improves the effectiveness of SSL methods~\citep{moco_v2}. We visualize this technique in Figure~\ref{fig:proj_pred}, as implemented in \texttt{BYOL} framework~\citep{byol}. Note that this visualization illustrates an asymmetric architecture but the asymmetry is not a necessity for projection/prediction MLPs.



\textbf{Negative samples}\,\textendash\,The InfoNCE loss (discussed in depth in Section~\ref{sec:lossfunctions}) aims at maximizing the similarity between representations of two augmentations of the same image, while minimizing the same metric across different images. In such cases, the \say{different} images are referred to as \textit{negative samples}~\citep{simclr}. This concept, which has been the focus of many research efforts (which we will discuss later on), will be particularly relevant for contrastive SSL~\citep{mocov1}.

\textbf{Memory bank}\,\textendash\,Given a set of $n$ images, $\texttt{x} = [\bm{x}_1,\ldots,\bm{x}_n]$, a memory bank refers to the simple idea of storing the corresponding image representations, as computed with $f_\theta(\mathtt{x}) = [\bm{r}_1, \ldots, \bm{r}_n]$, and to subsequently using this memory bank for various tasks (for example, to use the obtained image representations as negative samples in InfoNCE)~\citep{memory_bank,mocov1}.


\textbf{Pseudo-labeling}\,\textendash\,A number of SSL methods discussed below employ pseudo-labeling strategies to enable self-supervision~\citep{deepc,sela}. Such approaches can be visualized as shown in Figure~\ref{fig:pseudo_label}, where a label is assigned to an image based on its feature representation (through the use of, for example, K-means clustering) and where that label is then used to calculate a loss.

\subsection{Loss functions to train SSL frameworks}\label{sec:lossfunctions}

The forthcoming SSL frameworks utilize a wide range of loss functions to enable self-supervised training. Although the usage of these loss functions is often specific to certain frameworks, in this section, we will cover the most prominent losses that see common use across different frameworks.

\begin{minipage}{0.63\textwidth}
\textbf{Cross-entropy loss}\,\textendash\,Cross-entropy loss (CE) is a commonly used loss function in classification tasks which measures the difference between the predicted probabilities and the true probabilities of a categorical variable. Given a prediction $\hat{\bm{y}}$ for a $C$-class classification problem, CE for the class $t$ is calculated as follows:
\end{minipage}
\begin{minipage}[t]{0.04\textwidth}  
\phantom{...........}  
\end{minipage}  
\begin{minipage}[c]{0.32\textwidth}
\begin{equation*}
    \mathcal{L}_\text{CE} (\hat{\bm{y}}, t) =  -\log \frac
    {\exp(\hat{\bm{y}}_t)}
    {\sum_{c=0}^{C} \exp(\hat{\bm{y}}_c) } \,.\,\,
\end{equation*}
\end{minipage}

In clustering-based SSL, CE and its variants are mainly used with the target label $t$ being assigned via a self-labeling mechanism such as k-means clustering~\citep{deepc,sela,coke}. More recently, distillation-based SSL frameworks also make use of CE where the output of the student network is matched to that of the teacher~\citep{dino,obow,esvit}.

\begin{minipage}{0.63\textwidth}
\textbf{Cosine similarity}\,\textendash\,Cosine similarity measures the similarity between two non-zero vectors in a high-dimensional space, formalized as a dot product between $\ell_2$ normalized vectors $\bm{v}_1$ and $\bm{v}_2$ as follows:
\end{minipage}
\begin{minipage}[t]{0.04\textwidth}  
\phantom{...........}  
\end{minipage}  
\begin{minipage}[c]{0.32\textwidth}
\begin{equation*}
\cossim(\bm{v}_1, \bm{v}_2) = \frac{\bm{v}_1\,\cdot\, \bm{v}_2}{ \|\bm{v}_1\| \|\bm{v}_2\|} \,.\,\,\,
\end{equation*}
\end{minipage}

\begin{figure}[t!]
\centering
\begin{subfigure}{.49\textwidth}
\includegraphics[width=\textwidth]{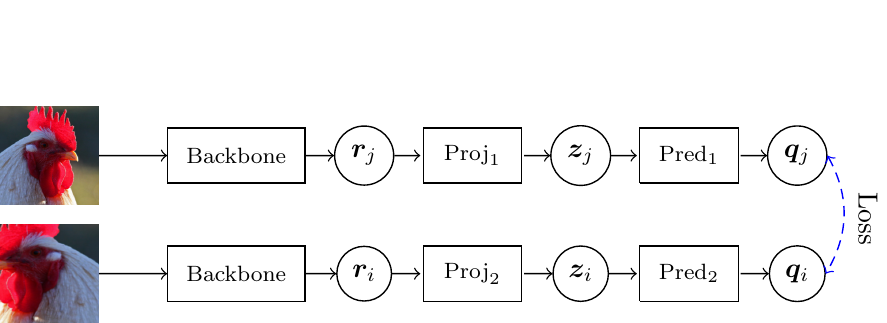}
\end{subfigure}
\begin{subfigure}{.49\textwidth}
\includegraphics[width=\textwidth]{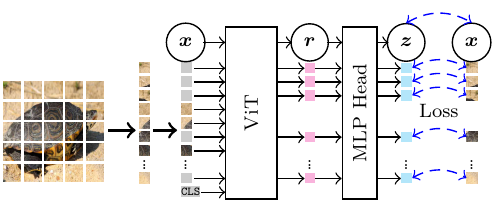}
\end{subfigure}
\caption{Illustrations of (left) dual-backbone discriminative and (right) MIM-based generative frameworks.}
\label{fig:ssl_overall_vis}
\end{figure}

In the context of SSL, cosine similarity is often employed in combination with noise-contrastive estimation (NCE) for contrastive-learning-based discriminative SSL frameworks. It is also employed by a number of prominent distillation networks to quantify representation similarity~\citep{byol,simsiam}. Given an image $\bm{x}$ and  two views $\bm{x}_{\{1,2\}} \sim \mathcal{T}(\bm{x})$ obtained with an augmentation $\mathcal{T}$, let $\bm{z}_{\{1,2\}}$ and $\bm{q}_{\{1,2\}}$ be the outputs of the projection and prediction layers, respectively, obtained by using a Siamese-like backbone similar to the one depicted in Figure~\ref{fig:ssl_overall_vis}. \texttt{SimSiam}, for example, then employs negative symmetric cosine similarity between projections and predictions defined as $- \frac{1}{2}\cossim(\bm{q}_1, \texttt{stop-grad}(\bm{z}_2)) - \frac{1}{2}\cossim(\bm{q}_2, \texttt{stop-grad(}\bm{z}_1))$ with \texttt{stop-grad}$(\cdot)$ referring to the stop-grad operation described above~\citep{simsiam}.

\textbf{Noise-contrastive estimation}\,\textendash\,A contrastive loss is a loss that has a low value when the two input images are similar and a large value when they are dissimilar~\citep{siamese_use3,early_contrastive_1}. A fundamental loss that enables contrastive training for image-based SSL is InfoNCE~\citep{infonce_2,cpc_infonce_1}, which is a modification of NCE~\citep{nce}. Following~\citet{simclr}, InfoNCE can be defined using $2n$ instances of $n$ images in a single batch: $\mathtt{x} = [\mathcal{T}(\bm{x}_1), \mathcal{T}(\bm{x}_1), \ldots, \mathcal{T}(\bm{x}_n), \mathcal{T}(\bm{x}_n)]$, with $\mathcal{T}(\cdot)$ a stochastic image augmentation function. In this scenario, the InfoNCE loss for a single positive pair is defined as follows:
\begin{equation}
    \mathcal{L}_\text{InfoNCE} (\mathtt{x}_{\{i,j\}}) =  -\log \frac
    {\exp(\cossim(\bm{r}_i, \bm{r}_j))}
    {\sum_{k=0}^{2n} \mathds{1}_{\{k \neq i\}} \exp(\cossim(\bm{r}_i, \bm{r}_k))} \,,
\end{equation}
where $f(\mathtt{x}_i) = \bm{r}_i$ denotes the feature representation of the $i$th data point. InfoNCE is the most employed loss function for SSL frameworks that use contrastive learning~\citep{simclr,mocov1}.

\textbf{Mean squared error}\,\textendash\,Defined as $\text{MSE} (\bm{v}, \hat{\bm{v}}) = \frac{1}{n} \sum_{i=1}^{n}(\bm{v}_i - \hat{\bm{v}}_i)^2$, the mean squared error (MSE) is employed in a number of prominent distillation-based SSL frameworks to measure feature alignment~\citep{byol,directPred,dino}. More recently, MSE has also been adopted to measure the correctness of reconstruction targets for MIM-based generative frameworks~\citep{mae,milan,spark}.

\textbf{Mean absolute error}\,\textendash\,Defined as $\text{MAE} (\bm{v}, \hat{\bm{v}}) = \frac{1}{n} \sum_{i=1}^{n} |\bm{v}_i - \hat{\bm{v}}_i|$, the mean absolute error (MAE) was a rarely used error measurement metric until the resurgence of MIM-based generative SSL, in which it is employed to measure the correctness of reconstruction targets~\citep{simmim,spark}.

\textbf{Information-maximization}\,\textendash\,Proposed by~\citet{wmse,barlow_twins,vicreg}, a unique method of self-supervision is to maximize the information content of the embeddings (i.e., projections/predictions). Compared to the previously discussed losses, losses that maximize information content of embeddings are not only unique but also much more complicated. 

For example, the loss of \texttt{VicReg}~\citep{vicreg} ---a popular information-maximization framework--- can be defined using two batches of $n$ image embeddings coming from two branches of a Siamese-like network, $\mathtt{q} = [\bm{q}_1, \ldots, \bm{q}_n]$ and $\mathtt{q}' = [\bm{q}'_1, \ldots, \bm{q}'_n]$. Then, the \texttt{VicReg} loss is defined as follows:
\begin{equation} \label{eq:loss_vicreg}
  \mathcal{L}_{\text{VIC}}(\mathtt{q}, \mathtt{q}') :=
  \lambda \underbrace{s(\mathtt{q}, \mathtt{q}')}_{\text{Invariance}} +
  \mu \underbrace{[v(\mathtt{q}) + v(\mathtt{q}')]}_{\text{Variance}} +
  \nu \underbrace{[c(\mathtt{q}) + c(\mathtt{q}')]}_{\text{Covariance}},
\end{equation}
where $\lambda$, $\mu$, and $\nu$ are hyperparameters, and the three constituent expressions in this complex loss function play the following role: (1) The \emph{invariance term} $s(\mathtt{q}, \mathtt{q}') = \frac{1}{n}\sum_{i = 1}^n \left\Vert \bm{q}_i - \bm{q}_i'\right\Vert_2^2$ aims to learn invariance to data transformations by making $\mathtt{q}$ and $\mathtt{q}'$ similar. (2) The \emph{variance term} $v(\mathtt{q})$ aims to prevent norm collapse by giving the components of $\mathtt{q}$ and $\mathtt{q}'$ a standard deviation equal to $\gamma$ (a fixed hyperparameter). It is defined as a hinge loss $v(\mathtt{q}) = \max(0, \gamma - S(\mathtt{q}, \epsilon))$, with $S(\mathtt{q}, \epsilon) = \sqrt{\text{Var}(\mathtt{q}) + \epsilon}$ the regularized standard deviation. (3) The \emph{covariance term} $c(\mathtt{q})$ strives to remove correlations between the different components of $\mathtt{q}$, and is given by the sum $\sum_{i \ne j} [C(\mathtt{q})]_{ij}^2$ over the off-diagonal elements of the $d$-dimensional covariance matrix $C(\mathtt{q})$.

\section{Self-supervised learning frameworks}
\label{sec:disc_ssl}
Although most recently proposed frameworks make use of a variety of techniques from both generative and discriminative SSL, the frameworks that we will discuss shortly can be typically categorized as either generative or discriminative. In the case when a framework leverages techniques that belong to multiple categories and may thus fall into more than one category, we adopt the designation used by its creators. Since most of the frameworks are known by their acronyms, we use their abbreviated names in the main text and provide their full names in Section~\ref{sec:abbrv} of the the appendix. 






\subsection{Discriminative SSL}
In terms of discriminative SSL, frameworks can roughly be grouped by their reliance on the following techniques: \textbf{clustering}, \textbf{contrastive learning}, \textbf{distillation} and \textbf{information-maximization}. In what follows, we detail discriminative SSL frameworks that fall under the aforementioned categories.

\subsubsection{Clustering}

Self-labeling via clustering is one of the most straightforward ways to achieve self-supervision, with clustering being one of the most popular methods for unsupervised learning~\citep{Bishop06a}. For neural networks, the usage of clustering-based methods for training can be traced back to the seminal works of~\citet{clustering_early_0}, \citet{clustering_early_1}, and \citet{clustering_early_2}, which paved the way for the use of such methods for SSL. Unfortunately, clustering-based methods have to solve a number of well-documented issues such as: (1) offline training that prevents their usage for large-scale data, (2) large clusters dominating the majority of the labels or small clusters leading to extremely granular labels, (3) empty clusters, (4) requiring knowledge about the number of clusters beforehand, and (5) trivial solutions where all data are gathered in a single cluster which causes the network to collapse~\citep{clustering_collapse_2,clustering_collapse}. Since these issues are fundamental problems of clustering, all of the clustering-based SSL methods have to tackle these problems in their own unique way when trying to perform self-supervision.

The pioneering work of~\citet{deepc} put forward \texttt{Deep Cluster}, one of the first clustering-based SSL methods that achieves results comparable to supervised models. This method solves the issues listed above with an offline training approach and by forcing a uniform distribution across clusters, both of which limit the usage of \texttt{Deep Cluster}. Following that, getting rid of the tricks applied in \texttt{Deep Cluster} became the primary focus of a number of subsequent studies, leading to improved clustering-based SSL methods such as \texttt{SeLA}~\citep{sela}, \texttt{Online Deep Cluster}~\citep{online_deep_clustering}, and \texttt{Self-Classifier}~\citep{self_classifier}. \texttt{SeLa} tackles the issue of model collapse by incorporating a more principled loss using the Sinkhorn-Knopp algorithm~\citep{sinkhorn_algo}. \texttt{Online Deep Clustering} on the other hand addresses the aforementioned offline training issue to enable online training for large datasets.

Conversely,~\citet{scan} argue that an end-to-end approach with online training may lead to various problems and propose an approach called \texttt{SCAN} that replaces the use of K-Means for the purpose of clustering with the use of an advanced neighbor search. When it comes to the state-of-the-art, the clustering-based method proposed in~\citet{swav}, known as \texttt{Swav}, which also leverages a number of contrastive elements, is currently considered to be the most stable and accurate approach. Table~\ref{tbl:ssl_info_clustering} provides a summarizing overview of several clustering-based SSL methods, detailing their unique traits.



\begin{table}[t!]
\centering
\caption{SSL frameworks that rely on \textbf{clustering}-based self-supervision and their unique properties.}
\scriptsize
\begin{tabular}{lll}
\cmidrule[1pt]{1-3}
SSL framework & Proposed by & Unique property \\
\cmidrule[0.5pt]{1-3}

\texttt{Deep Cluster} & \citet{deepc} & Avoids trivial solutions for clustering-based SSL \\

\texttt{Local Aggregation} & \citet{local_agg} & Local aggregation metric for soft cluster assignments \\

\texttt{Deeper Cluster} & \citet{deeperC} & Integrates rotation-based SSL into clustering \\

\texttt{SeLa} & \citet{sela} & Improves \texttt{Deep Cluster} with the Sinkhorn-Knopp algorithm \\

\texttt{SCAN} & \citet{scan} & Decouples feature learning and clustering using a two-step approach\\

\texttt{Deep Cluster-v2} & \citet{swav} & Incorporates various SSL improvements into \texttt{Deep Cluster} \\

\texttt{SeLa-v2} & \citet{swav} & Incorporates various SSL improvements into \texttt{SeLa} \\

\texttt{Swav} & \citet{swav} & Online clustering with consistency across assignments \\

\texttt{ODC} & \citet{online_deep_clustering} & Converts \texttt{Deep Cluster} into an online method \\

\texttt{CoKe} & \citet{coke} & Improves the clustering phase with an online constrained k-means method \\

\texttt{Self-Classifier} & \citet{self_classifier} & Single-stage end-to-end clustering combined with contrastive learning \\

\cmidrule[1pt]{1-3}
\end{tabular}
\label{tbl:ssl_info_clustering}
\vspace{-1.5em}
\end{table}

\begin{table}[t!]
\centering
\caption{SSL frameworks that rely on \textbf{contrastive learning}-based self-supervision and their unique properties.}
\scriptsize
\begin{tabular}{lll}
\cmidrule[1pt]{1-3}
SSL framework & Proposed by & Unique property \\
\cmidrule[0.5pt]{1-3}

\texttt{InstDist} (\texttt{NPID}) & \citet{instdist} & Non-parametric softmax calculation \\

\texttt{CPC} & \citet{cpc_infonce_1} & Usage of InfoNCE loss across multiple tasks \\

\texttt{DIM} & \citet{dim} & Measures representation quality with two novel losses (MINE and NDM) \\

\texttt{CPC-v2} & \citet{cpc_v2} & Improves \texttt{CPC} architecture and training \\

\texttt{AMDIM} & \citet{amdim} & Extends \texttt{DIM} for mixture-based representations \\

\texttt{CMC} & \citet{cmc} & Information-maximization across different sensory views \\

\texttt{MoCo} & \citet{mocov1} & SSL with momentum encoder and memory bank \\

\texttt{PIRL} & \citet{pirl} & Contrastive learning with jigsaw puzzles \\

\texttt{SimCLR} & \citet{simclr} & Usage of projection heads and new augmentations \\

\texttt{MoCo-v2} & \citet{moco_v2} & Improves \texttt{MoCo} with the design of \texttt{SimCLR} \\

\texttt{SimCLR-v2} & \citet{simclr_2} & Improves \texttt{SimCLR} with memory bank and deeper projector MLPs \\

\texttt{PCL} \& \texttt{PCL-v2}  & \citet{pcl} & Formulates contrastive learning with clustering using EM\\

\texttt{PIC} & \citet{pic} & One-branch parametric instance classification \\

\texttt{DCL} & \citet{negative_samples_1} & Negative sample section with a debiased contrastive objective \\

\texttt{LooC} & \citet{looc} & Learns transformation dependent and invariant representations \\

\texttt{G-SimCLR} & \citet{g_simclr} & SimCLR with negative sample selection using pseudo-labels \\

\texttt{ReLIC} & \citet{relic} & Imposes invariance constraints during SSL training \\

\texttt{AdCo} & \citet{adco} & Mixes self-trained negative adversaries into SSL \\

\texttt{DenseCL} & \citet{dense_cl} & Dense contrastive loss for SSL \\

\texttt{PixPro} & \citet{pix_pro} & PixContrast and PixPro losses for contrastive SSL \\

\texttt{MoCo-v3} & \citet{moco_v3} & Improves \texttt{MoCo-v2} with symmetrized loss and without a memory bank \\

\texttt{CLSA} & \citet{csla} & Usage of stronger augmentations for contrastive learning \\

\texttt{Truncated Triplet} & \citet{truncated_triplet} & Attempts to solve under- and over-clustering in contrastive learning\\

\texttt{NNCLR} & \citet{nnclr} &  Nearest-neighbors as positive samples in contrastive loss\\

\texttt{MoBY} & \citet{moby} & Combines design principles of \texttt{MoCo} and \texttt{BYOL} for transformers  \\

\texttt{DNC} & \citet{dnc} & Alternation of contrastive learning and clustering-based hard negative mining \\

\texttt{ReSSL} & \citet{ressl} & Maintains the relational consistency between different instances of images \\

\texttt{UniGrad} & \citet{uni_grad} & Unifies contrastive learning, distillation, and information-maximization \\

\texttt{ReLIC-v2} & \citet{relic_2} & Improves \texttt{ReLIC} with inductive biases to learn more informative representations \\

\texttt{SimCo} & \citet{negative_samples_2} & Simplifies \texttt{MoCo} with momentum removal \\

\texttt{SimMoCo} & \citet{negative_samples_2} & Simplifies \texttt{MoCo} with dictionary removal \\

\texttt{UniVIP} & \citet{uni_vip} & Scene-based SSL based on similarity, correlation, and discrimination \\

\texttt{Mugs} & \citet{mugs} & Explicitly learns multi-granular visual features\\

\texttt{CaCo} & \citet{caco} & Learns both positive and negative samples end-to-end with an encoder \\

\texttt{SMoG} & \citet{smog} & Replaces instance contrastive learning with group contrastive learning \\

\texttt{SiameseIM} & \citet{siameseim} & Instance discrimination using \texttt{UniGrad} and masked images \\

\cmidrule[1pt]{1-3}
\end{tabular}
\label{tbl:ssl_info_contrastive}
\vspace{-1.5em}
\end{table}

\vspace{-0.25em}
\subsubsection{Contrastive learning}
\vspace{-0.25em}

Contrastive learning with the InfoNCE loss (or an extension of it) is the most popular approach for self-supervision and also the one that received the most research contributions in the past years. Contrastive methods can be traced back to the works of~\citet{siamese} and \citet{siamese_use3}, but in terms of modern usage of SSL, \citet{instdist} and \citet{cpc_infonce_1} popularized this line of research by proposing \texttt{InstDist} and \texttt{CPC}, respectively. \citet{dim} and \citet{amdim} investigated different ways to measure representation quality for contrastive learning and proposed \texttt{DIM} and \texttt{AMDIM} respectively, while \citet{cmc} extended contrastive learning for multiple sensory inputs with \texttt{CMC}. After the aforementioned works contrastive SSL attracted significant research interest but it was the groundbreaking results obtained with \texttt{MoCo} which used memory banks with delayed weight updates that put contrastive SSL really into the spotlight~\citep{mocov1}. Shortly after,~\citet{simclr} proposed \texttt{SimCLR} and with it, further improved the state-of-the-art with the help of projection heads and strong augmentations and cemented the importance of contrastive self-supervision as a learning paradigm. Incorporating the enhancements of \texttt{SimCLR} into \texttt{MoCo}, \citet{moco_v2} proposed \texttt{MoCo-v2} and showed that there still exists a large margin for improvement. \citet{moco_v3} later introduced a third version of \texttt{MoCo}, exploring the usage of vision transformers as backbones. The reliable design of \texttt{MoCo} and its improved versions were the foundation of many subsequent contrastive SSL frameworks, such as \texttt{AdCo}~\citep{adco}, \texttt{MocHi}~\citep{mochi}, and \texttt{DenseCL}~\citep{dense_cl}.

While the above architectures mostly use dual backbones,~\citet{pic} proposed \texttt{PIC} and demonstrated the viability a single-branch backbone architecture for contrastive learning.~\citet{mochi} experimented with hard negative samples for improving the effectiveness of contrastive learning and~\citet{csla} demonstrated the usefulness of stronger augmentations. After the success of \texttt{Moco-v2} and \texttt{Moco-v3}, and with the increased availability of unique SSL methods, frameworks like \texttt{G-SimCLR}~\citep{g_simclr}, \texttt{MoBY}~\citep{moby}, \texttt{SimCo}, and \texttt{SimMoCo}~\citep{negative_samples_2}, which combine multiple SSL methods into a single one, gained traction. More recently, SSL frameworks such as \texttt{UniGrad}~\citep{uni_grad} claim to combine four self-supervision methodologies (clustering, contrastive, distillation, and information-maximization) into a single framework and to unify discriminative SSL training.

Although contrastive methods garnered more attention than clustering-based methods, they are also subject to a similar problem that needs to be mitigated: network collapse~\citep{dimensional_collapse_directCLR}. Contrastive methods prevent complete collapse of a network through the use of negative samples. However,~\citet{ssl_collapse} surprisingly demonstrated that contrastive SSL frameworks can suffer from another type of collapse, namely dimensional collapse, wherein representations collapse into a low-dimensional manifold. Given the importance of negative samples in preventing collapse in contrastive SSL, understanding the effects of negative samples and finding better sampling techniques became an active research topic shortly after~\citep{negative_samples_1,negative_samples_3,negative_samples_2}. A summarizing overview of several contrastive SSL frameworks can be found in Table~\ref{tbl:ssl_info_contrastive}.

\begin{table}[t!]
\centering
\caption{SSL frameworks that rely on \textbf{distillation}-based self-supervision and their unique properties.}
\scriptsize
\begin{tabular}{lll}
\cmidrule[1pt]{1-3}
SSL framework & Proposed by & Unique property \\
\cmidrule[0.5pt]{1-3}

\texttt{BYOL} & \citet{byol} & Avoids trivial solutions through network asymmetry \\

\texttt{SimSiam} & \citet{simsiam} & SSL with simple Siamese networks without negative samples \\

\texttt{OBoW} & \citet{obow} & Online bag-of-visual-words for SSL \\

\texttt{DirectPred} & \citet{directPred} & Adjusts linear predictor with a gradient-free approach \\

\texttt{SEED} & \citet{seed} & Knowledge distillation from large to small models \\

\texttt{DisCo} & \citet{disco} & Combines contrastive and distillation learning for lightweight models\\

\texttt{DINO} & \citet{dino} & Knowledge distillation with vision transformers \\

\texttt{EsViT} & \citet{esvit} & Multi-stage architectures with sparse self-attentions and region matching for efficient SSL\\

\texttt{BINGO} & \citet{bingo} & Distillation-based SSL for small-scale models \\

\texttt{TinyMIM} & \citet{tinymim} & Distillation to transfer knowledge from large MIM-based models to small models \\

\cmidrule[1pt]{1-3}
\end{tabular}
\label{tbl:ssl_info_distillation}
\vspace{-1.5em}
\end{table}

\subsubsection{Distillation}

Can the collapse of networks be prevented without the use of self-labeling or a contrastive loss that relies on negative samples? Through an asymmetric framework called \texttt{BYOL},~\citet{byol} demonstrated that neither of those techniques are necessary to achieve self-supervision when the proposed method relies on distillation~\citep{distilling_old}. The general idea behind distillation is to train a network (student) to predict representations of another one (teacher)~\citep{mean_teacher}. Shortly after the proposal of \texttt{BYOL},~\citet{simsiam} proposed \texttt{SimSiam}, a symmetric (Siamese) framework that  uses neither negative samples nor clustering, but leverages instead stop-grad and projection/prediction MLPs. This was followed by \texttt{OBOW}~\citep{obow}, in which the task is to reconstruct a bag-of-visual-words representation.

Similar to the trends witnessed for clustering and contrastive-learning, distillation-based SSL frameworks were experimentally combined with other frameworks in an attempt to obtain boosts in overall effectiveness. Frameworks such as \texttt{DisCo}~\citep{disco} and \texttt{MoBY}~\citep{moby} merged multiple frameworks together, while others tried to improve the effectiveness of established methods, such as \texttt{MSF}~\citep{msf} and \texttt{ORL}~\citep{orl}, improving upon \texttt{BYOL}.

How do distillation methods avoid network collapse? \citet{infonce_understand1} and \citet{byol_collapse_h1} argued that methods that incorporate batch statistics into training (e.g., batch normalization) aid \texttt{BYOL} (and potentially other distillation-based methods) in preventing collapse, but this hypothesis was promptly refuted by \citet{byol_works}. Recently, \citet{distilaltion_collapse} scrutinized \texttt{SimSiam} and found it to be highly sensitive to model size. Nevertheless, a definite answer to the way distillation-based SSL methods avoid collapse is not yet found. Table~\ref{tbl:ssl_info_distillation} provides a summarizing overview of several SSL frameworks that rely on distillation.

\begin{table}[t!]
\centering
\caption{SSL frameworks that rely on \textbf{information-maximization}-based self-supervision and their unique properties.}
\scriptsize
\begin{tabular}{lll}
\cmidrule[1pt]{1-3}
SSL framework & Proposed by & Unique property \\
\cmidrule[0.5pt]{1-3}

\texttt{WMSE} & \citet{wmse} & Whitening Mean Squared Error loss for information-maximization \\

\texttt{Barlow Twins} & \citet{barlow_twins} & SSL with redundancy reduction \\

\texttt{VicReg} & \citet{vicreg} & Variance-invariance-covariance regularization for avoiding collapse \\

\texttt{TWIST} & \citet{twist} & Theoretically explainable \texttt{TWIST} loss that avoids collapse \\

\texttt{TLDR} & \citet{tldr} & Improves \texttt{Barlow Twins} with \texttt{TLDR} encoder\\

\texttt{ARB} & \citet{arb} & Aligns feature representations with nearest orthonormal basis\\

\texttt{VicRegL} & \citet{vicregl} & Improves \texttt{VicReg} with location- and feature-based matching \\

\cmidrule[1pt]{1-3}
\end{tabular}
\label{tbl:ssl_info_maximization}
\vspace{-1.5em}
\end{table}

\subsubsection{Information-maximization}

The fourth and final discriminative self-supervision category we cover is information-maximization, having as primary idea the maximization of the information conveyed by decorrelated embeddings. Such approaches come with a number of advantages, in particular, they neither require negative samples nor require an asymmetric architecture to avoid collapse. Instead, they completely rely on innovative loss functions to avoid collapse. As a result, most of the frameworks that fall under this category can be characterized by the novel loss function that is used.

Information-maximization as a method for self-supervision was put forward by~\citet{wmse} and~\citet{barlow_twins}, where the former proposed \texttt{W-MSE} loss, which constrains the batch samples to dissipate in a spherical distribution, and where the latter (\texttt{Barlow Twins}) aims at making the normalized cross-correlation matrix of the embedding vectors to be close to the identity matrix.~\citet{vicreg} further improved the loss of \texttt{Barlow Twins} with the \texttt{VicReg} framework, proposing a loss based on variance, invariance, and covariance (described above in \eqref{eq:loss_vicreg}). Successor frameworks such as \texttt{TWIST}~\citep{twist}, \texttt{TLDR}~\citep{tldr}, and \texttt{ARB}~\citep{arb} followed the path paved by the previous frameworks and aim at improving the losses in different ways. Due to the complex nature of the losses used in information-maximization as a method for self-supervision, we refer the interested reader to the respective research papers underlying those frameworks. Table~\ref{tbl:ssl_info_maximization} provides a summarizing overview of several SSL frameworks that rely on information-maximization.

\begin{table}[t!]
\centering
\caption{\textbf{Enhancements} for existing discriminative SSL frameworks and their unique properties.}
\scriptsize
\begin{tabular}{lll}
\cmidrule[1pt]{1-3}
Module & Proposed by & Unique property \\
\cmidrule[0.5pt]{1-3}

\texttt{InfoMin} & \citet{info_min} & InfoMin principle and evaluation of augmentations \\

\texttt{InterCLR} & \citet{inter_clr} & Inter-image invariance for contrastive learning  \\

\texttt{HEXA} & \citet{hexa} & Proposes new data augmentation methods that are harder to predict \\

\texttt{MocHi} & \citet{mochi} & Hard negative image mixing approach \\

\texttt{ReSim} & \citet{resim} & Enhances SSL representations with region similarities \\

\texttt{MSF} & \citet{msf} & Enhances \texttt{BYOL} by shifting the embeddings to be close to the mean of its instances \\

\texttt{ORL} & \citet{orl} & Utilizes \texttt{BYOL} for object-level training \\

\texttt{CEB} & \citet{ceb} & Measures the amount of compression in the learned representations \\

\texttt{SEM} & \citet{byol_sem} & Employs simplicial embeddings to map unnormalized representations onto simplices \\

\texttt{ENS} & \citet{enh_ens} & Investigates optimal ensemble models for discriminative SSL frameworks \\

\texttt{MRCL} & \citet{mrcl} & Uses MIM as a method to avoid the discriminative information overfitting\\

\texttt{TS} & \citet{enh_ts} & Assists contrastive methods to learn group-wise features and instance-specific details \\

\texttt{ARCL} & \citet{enh_arcl} & Enhances contrastive learning with domain-invariant features representations \\

\texttt{MosRep} & \citet{wangmosaic} & Data augmentation strategy that enriches backgrounds of crops \\

\cmidrule[1pt]{1-3}
\end{tabular}
\label{tbl:ssl_modules}
\vspace{-1.5em}
\end{table}

\subsection{Enhancements for existing frameworks}

So far, we have covered a large number of discriminative SSL frameworks, all of which have consistently improved state-of-the-art results across various computer vision tasks. However, we have observed a trend that emerged towards the end of 2020: framework-agnostic enhancements. These modules are small tweaks to existing frameworks that can improve their performance in various ways, such as utilizing harder images/augmentations~\citep{mochi, hexa}, improving object-level representations~\citep{resim, orl}, or enabling optimal ensemble models~\citep{enh_ens}. For completeness, we have listed these modules separately in Table~\ref{tbl:ssl_modules}.

\subsection{Generative SSL}
\label{sec:generative_ssl}

From $2018$ onward, generative SSL was largely dismissed in favor of discriminative training methods with contrastive learning holding the prime spot for research~\citep{mocov1,simclr}. Popular discriminative frameworks such as \texttt{MoCo}, \texttt{SimCLR}, and \texttt{BYOL} were employed for a variety of unique tasks and were further improved with enhancements taken from each other~\citep{moco_v2,simclr_2}. In an unexpected turn of events, towards the end of $2021$, generative SSL came to dominate image-based self-supervised learning and became the primary research focus, dethroning contrastive learning as well as discriminative SSL~\citep{beit,ibot}. The advances in the field from the last quarter of $2021$ to the first quarter of $2023$ were so rapid that state-of-the-art results in generative SSL were improved on a monthly basis. This rapid expansion also came with a large category of unique approaches which resulted in frameworks of generative SSL becoming much less standardized as opposed to frameworks in discriminative SSL where the latter mostly contains straightforward Siamese-like dual-backbones as shown in Figure~\ref{fig:ssl_overall_vis}. In order to improve readability, we will group generative SSL frameworks into two categories: the ones that use generative adversarial networks (\texttt{GAN}s) and others that use a form of masked image modeling.

\begin{table}[t!]
\centering
\caption{SSL frameworks that rely on \textbf{GAN-based} generative self-supervision and their unique properties.}
\scriptsize
\begin{tabular}{lll}
\cmidrule[1pt]{1-3}
SSL framework & Proposed by & Unique property \\
\cmidrule[0.5pt]{1-3}

\texttt{BiGAN} & \citet{bigan} & Bidirectional \texttt{GAN} with additional encoder modela \\

\texttt{ALI} & \citet{ali} & A \texttt{GAN} framework with an additional inference model \\

\texttt{BigBiGAN} & \citet{bigbigan} & \texttt{BiGAN} with the generator of \texttt{BigGAN} \\

\texttt{SS-GAN} & \citet{ss_gan} & \texttt{GAN} with auxiliary rotation loss \\

\texttt{SS-GAN-LA} & \citet{ssgan_la} & \texttt{SS-GAN} with label augmentation \\

\texttt{Vit-VQGAN} & \citet{vit_vqgan} & \texttt{VQGAN} with label quantization and ViT backbone \\

\cmidrule[1pt]{1-3}
\end{tabular}
\label{tbl:ssl_info_gan}
\vspace{-1.5em}
\end{table}

\subsubsection{GAN-based generative SSL}

While the usage of generative neural networks can be traced back to the work of~\citet{hinton2006fast}, it was the seminal work of~\citet{gan} that popularized generative models with the newly proposed \texttt{GAN} framework. Since the work of \citet{gan}, numerous \texttt{GAN} variants were proposed  with some of them recently taking advantage of advances in SSL, such as incorporating rotation prediction~\citep{ss_gan}, jigsaw puzzles~\citep{deshufflegans1,deshufflegans2}, and self-labeling~\citep{gan_selflabel}. However, most of the research in the \texttt{GAN} space has primarily focused on enhancing the fidelity of images generated by the generator network, which is typically evaluated using metrics such as the Fr\'echet Inception Distance~\citep{frechlet_dist}. As a result, these studies largely ignore the discriminative network and lack comparative evaluations on downstream tasks, leaving them out of the scope of SSL. In what follows, we focus on those research efforts that evaluate the discriminative power of \texttt{GAN}s on downstream tasks.

With a unique twist to \texttt{GAN}s, \citet{bigan} proposed \texttt{BiGAN}, a \texttt{GAN} framework that contains an additional encoder network trained in conjunction with the generator and discriminator networks with the objective of inverting the generator. After the training is completed, this encoder can be used as a feature extractor for downstream tasks. Independently,~\citet{ali} proposed a generative framework called \texttt{ALI} that is almost identical to \texttt{BiGAN}. Leveraging the improved generator of \texttt{BigGAN}~\citep{biggan} in \texttt{BiGAN},~\citet{bigbigan} proposed \texttt{BigBiGAN} which comes with better downstream transferability results. Taking inspiration from the developments in the area of discriminative SSL,~\citet{ss_gan} proposed \texttt{SS-GAN} which exploits rotation as an auxiliary task to achieve self-supervision with \texttt{GAN}s. This framework was further improved with the addition of label augmentation by~\citet{ssgan_la}. One of the most recent approaches within \texttt{GAN}-based SSL is \texttt{ViT-VQGAN}~\citep{vit_vqgan} which improves \texttt{VQGAN}~\citep{vqgan} using \texttt{ViT} backbones.

Overall, the usage of \texttt{GAN}s in SSL has not become a mainstream method due to a number of \texttt{GAN}-related limitations, ranging from mode collapse to limitations related to scalability, as well as lack of flexibility in backbone networks.


\begin{table}[t!]
\centering
\caption{SSL frameworks that rely on \textbf{MIM-based} generative self-supervision and their unique properties.}
\scriptsize
\begin{tabular}{lll}
\cmidrule[1pt]{1-3}
SSL framework & Proposed by & Unique property \\
\cmidrule[0.5pt]{1-3}

\texttt{iGPT} & \citet{igpt} & MIM with 9-bit pixel clustering per patch \\
\texttt{BEiT} & \citet{beit} & Patch-based MIM with ViTs using offline \texttt{DALL-E} tokenizer  \\
\texttt{MAE} & \citet{mae} & MIM with autoencoders using lightweight ViT encoders and pixel-based reconstruction\\
\texttt{iBOT} & \citet{ibot} & \texttt{BEiT} with an online tokenizer trained using the \texttt{DINO} objective \\
\texttt{SimMIM} & \citet{simmim} & \texttt{BEiT} with a pixel reconstruction target without a tokenizer \\
\texttt{PeCO} & \citet{peco} & Proposes a new codebook to replace the \texttt{DALL-E} tokenizer \\
\texttt{MaskFeat} & \citet{masked_feature_modeling} & MIM training with HOG as a reconstruction target \\
\texttt{data2vec} & \citet{data2vec} & \texttt{BEiT} with a teacher-student model and stronger augmentations \\
\texttt{CAE} & \citet{cae} & \texttt{MAE} with \texttt{DALL-E} token target reconstruction \\
\texttt{CIM} & \citet{cim} & Corrupted image modeling for generative SSL with an additional discriminative objective \\
\texttt{MCMAE} & \citet{mcmae} & Multi-scale hybrid convolution-transformer for improved MIM performance \\
\texttt{ConMIM} & \citet{conmim} & Contrastive learning on MIM patches \\
\texttt{CMAE} & \citet{cmae} & MIM + contrastive learning with shifted image views \\
\texttt{SdAE} &  \citet{sdae} & Self-distillation with high-level feature reconstruction \\
\texttt{MILAN} & \citet{milan} & MIM with semantic-aware mask sampling and \texttt{CLIP}-assisted feature reconstruction \\
\texttt{BEiT-v2} & \citet{beit-v2} & \texttt{BEiT} with \texttt{CLIP} tokenizer as a teacher and patch aggregation strategy \\
\texttt{BEiT-v3} & \citet{beit-v3} & \texttt{BEiT-v2} with text fusion \\
\texttt{CAE-v2} & \citet{cae-v2} & \texttt{CAE} with \texttt{CLIP} tokenizer \\
\texttt{CAN} &  \citet{can} & Combines contrastive learning, MIM, and image denoising with symmetric backbones \\
\texttt{PCAE} & \citet{pcae} & Progressively drops reconstruction tokens in \texttt{MAE} for better speed/performance trade-off \\
\texttt{SparK} & \citet{spark} & MIM for convolutional neural networks \\
\texttt{MRMAE} & \citet{mrmae} & Uses pixels, \texttt{DINO} features, and \texttt{CLIP} features for reconstruction \\
\cmidrule[1pt]{1-3}
\end{tabular}
\label{tbl:ssl_info_mim}
\vspace{-1.5em}
\end{table}

\subsubsection{Generative SSL with masked image modeling}

Although only a couple of years have passed since the discovery of ViTs~\citep{vit}, these architectures have been shown to achieve state-of-the-art results in a variety of vision tasks. In their work,~\citet{vit} demonstrated the feasibility of SSL using MIM as a pretext task (although it was called masked patch prediction by the authors), where this pretext task is seamlessly supported by the patch-based image intake of ViTs. Developing this technique further, \texttt{BEiT} was one of the first frameworks that successfully employed MIM with vector quantized images and ViTs~\citep{beit}. It is important to note that \texttt{BEiT} does not directly predict the pixel values of the image but learns to predict discrete visual tokens which are created from image patches~\citep{image_tokenization}. \texttt{BEiT} uses the tokenizer --- a discrete variational autoencoder (dVAE) --- of \texttt{DALL-E}~\citep{tokenizer} which requires an offline training before training the final model. \texttt{BEiT} gave rise to \texttt{BEiT-v2}~\citep{beit-v2} and \texttt{BEiT-v3}~\citep{beit-v3}\footnote{\texttt{BEiT-v3} is a framework that fuses vision and text but we include it for completeness.} which obtain better results using the \texttt{CLIP} tokenizer~\citep{clip_tokenizer} with a patch aggregation strategy. Meanwhile, the requirement of an external tokenizer for \texttt{BEiT} was alleviated by \texttt{iBOT}~\citep{ibot} which introduced an online tokenizer trained with the distillation routine of \texttt{BYOL}, thus leveraging the advances made on the side of discriminative SSL.~\citet{simmim} got rid of the tokenizer of \texttt{BEiT} and proposed \texttt{SimMIM}, which directly operates over pixel values and predicts them. 

Narrowly predating \texttt{BEiT},~\citet{igpt} proposed \texttt{iGPT} by leveraging \texttt{GPT-2}~\citep{gpt_2} and adapted it to vision, which represents images with tokenized patches using a 9-bit color palette by clustering RGB pixels, and then training this model with the MLM objective of \texttt{BERT}~\citep{bert}. The primary difference between the training objective of \texttt{iGPT} (i.e., \texttt{BERT}-style MLM) and MIM of \texttt{BEiT} is that the latter directly uses image patches as an input, therefore not losing any pixel-level information.

With a unique take on MIM, \citet{mae} proposed \texttt{MAE}, an asymmetric autoencoder framework that directly learns to reconstruct image patches. What is unique to \texttt{MAE} is that its encoder (ViT) only processes unmasked patches (e.g., $25\%$ of all patches) without any tokenizer, making it much faster than the frameworks we have discussed thus far.~\citet{mae} also evaluated the effectiveness of different reconstruction targets and found no statistically significant difference between reconstructing \texttt{DALL-E} tokens and pixels, suggesting simple pixel reconstruction to be a viable reconstruction target. Building upon \texttt{MAE},~\citet{cae} proposed \texttt{CAE} which  comes with a latent contextual regressor and uses the \texttt{DALL-E} tokenizer, which was replaced in the next iteration of this framework (\texttt{CAE-v2})~\citep{cae-v2} by a \texttt{CLIP} tokenizer.

\begin{figure}[t!]
\floatbox[{\capbeside\thisfloatsetup{capbesideposition={right,top},capbesidewidth=8cm}}]{figure}[\FBwidth]
{\caption{ImageNet top-1 accuracy with \textbf{linear probing} on frozen representations for \textbf{discriminative SSL frameworks} is plotted against the number of parameters in the trained backbone. The connecting lines indicate different backbone networks trained with the same framework. In both figures, nodes with circles indicate CNN-based architectures, whereas triangles indicate transformer-based architectures. Note that a few frameworks with overlapping results are omitted from the figures, and that axis values are scaled independently to improve visual clarity.}
\label{fig:ssl_acc_vs_param}}
{\includegraphics[width=0.5\textwidth]{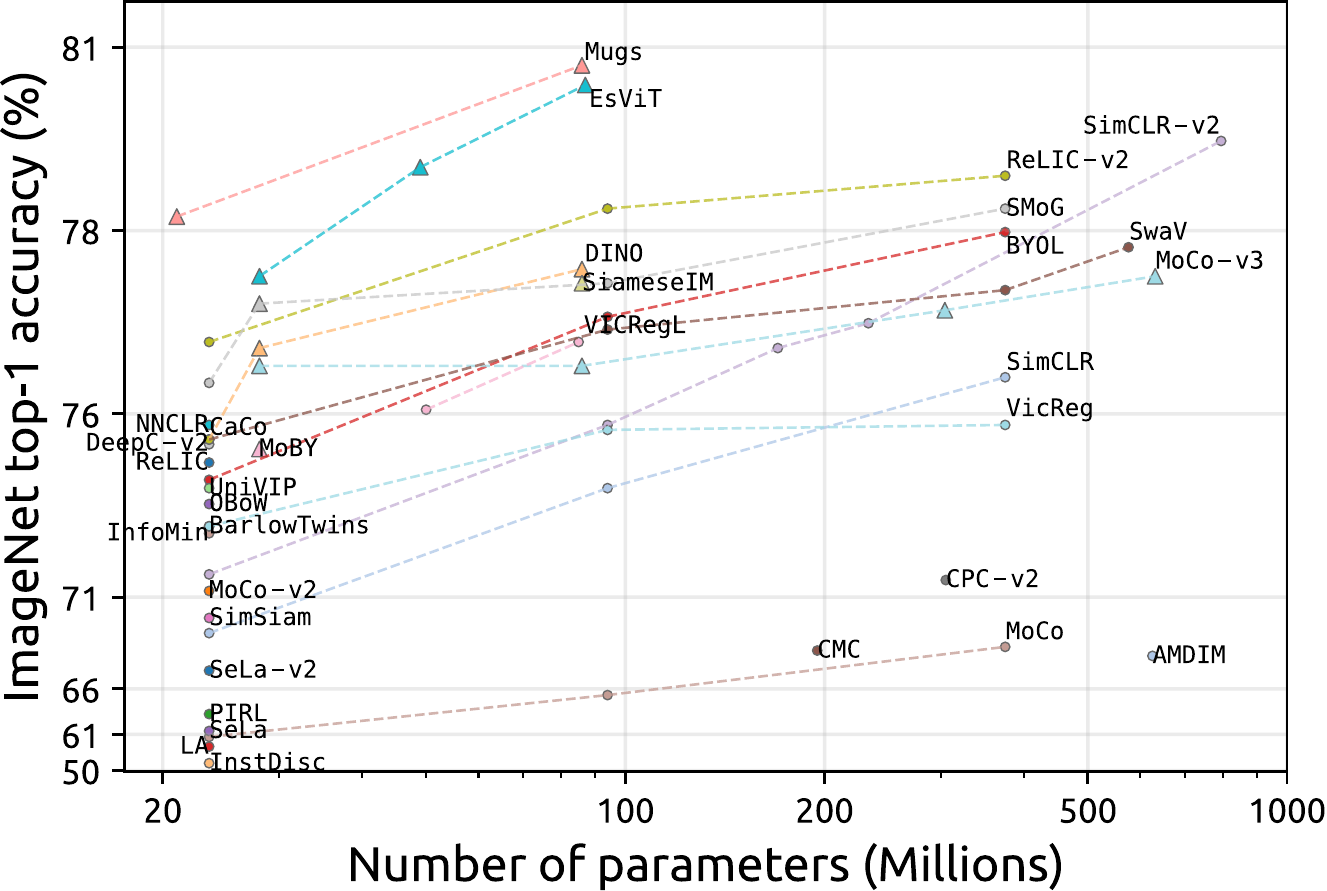}}
\end{figure}

At this point we believe it is important to reiterate that MIM was explored using different reconstruction targets such as (1) dVAE-based patch tokens~\citep{beit}, (2) clustering-based patch tokens~\citep{igpt}, and (3) pixel values~\citep{simmim}. Expanding this corpus,~\citet{masked_feature_modeling} proposed \texttt{MaskFeat} in which the task is to predict Histograms of Oriented Gradients (HOGs) --- a hand-crafted feature descriptor (see Figure~\ref{fig:mfp}) --- and argued that a broad spectrum of image features can be used as targets in MIM. Following their work, \texttt{SdAE}~\citep{sdae} and \texttt{MILAN}~\citep{milan} demonstrated the feasibility of reconstructing high-level features. For an overview of reconstruction targets for MIM-based generative SSL frameworks, see Table~\ref{tbl:reconstruction_tokens}.

\begin{minipage}[t]{0.25\textwidth}
\scriptsize
\captionof{table}{Reconstruction targets for generative SSL frameworks that use MIM.}
\label{tbl:reconstruction_tokens}
\begin{tabular}{ll}
\cmidrule[1pt]{1-2}
\text{Framework} & \text{Reconstruction} \\
\cmidrule[0.5pt]{1-2}
\texttt{iGPT} & \text{9-bit pixels} \\
\texttt{BEiT} & \text{\texttt{DALL-E} tokens} \\
\texttt{MAE} & \text{Pixels} \\
\texttt{iBOT} & \text{Distilled tokens} \\
\texttt{SimMIM} & \text{Pixels} \\
\texttt{PeCO} & \text{\texttt{PeCO} tokens} \\
\texttt{MaskFeat} & \text{HOG} \\
\texttt{data2vec} & \text{\texttt{DALL-E} tokens} \\
\texttt{CAE} & \text{\texttt{DALL-E} tokens} \\
\texttt{CIM} & \text{\texttt{DALL-E} tokens} \\
\texttt{MCMAE} & \text{Pixels} \\
\texttt{ConMIM} & \text{Patch features} \\
\texttt{CMAE} & \text{Pixels} \\
\texttt{SdAE} & \text{High-level features} \\
\texttt{MILAN} & \text{High-level features} \\
\texttt{BEiT-v2} & \text{\texttt{CLIP} tokens} \\
\texttt{BEiT-v3} & \text{\texttt{CLIP} tokens} \\
\texttt{CAE-v2} & \text{\texttt{CLIP} tokens} \\
\texttt{CAN} & \text{Pixels} \\
\texttt{PCAE} & \text{Pixels} \\
\texttt{SparK} & \text{Pixels} \\
\texttt{MRMAE} & \text{\texttt{CLIP} tokens} \\
\cmidrule[1pt]{1-2}
\end{tabular}
\end{minipage}
\begin{minipage}[t]{0.04\textwidth}  
\phantom{...........}  
\end{minipage}  
\begin{minipage}[t]{0.71\textwidth}
Experimenting with the input side~\citet{cim} proposed \texttt{CIM} where an auxiliary generator (in their use-case, \texttt{BEiT}) corrupts the input image and the proposed framework aims to (1) discriminate patches (classification for each patch) and (2) generate the original image.
\vspace{0.5em} 
\\
The current trend for generative SSL is to combine both generative and discriminative losses together to improve the quality of representations. In particular, contrastive learning has become a popular task to combine with MIM with frameworks such as \texttt{CAN}~\citep{can}, \texttt{CMAE}~\citep{cmae} and \texttt{ConMIM}~\citep{conmim} leveraging advances made on the side of contrastive SSL.
\vspace{0.5em} 
\\
All of the generative frameworks we have discussed thus far use some form of vision transformer as a backbone as opposed to the majority of the discriminative frameworks, which make use of ResNets. In an attempt to leverage masked convolutions, \citet{mcmae} proposed \texttt{MCMAE}, a framework that employs the hybrid convolution-transformer \texttt{MAE} which is able to learn discriminative representations. Very recently, \citet{spark} showed that classical (ResNets) and modern (ConvNext~\citep{convnext}) CNNs can be trained with MIM and achieve state-of-the-art results that can rival those of ViTs. The results obtained by~\citet{spark} indicate that ViTs, which have been considered a prerequisite for MIM, are not irreplaceable and that CNNs can still compete with ViTs in generative SSL.
\end{minipage}

\section{Evaluating SSL models}
\label{sec:evaluation}
\vspace{-0.25em}

As briefly noted in Section~\ref{sec:important_concepts}, the SSL frameworks covered thus far are concerned with the training of feature extractors that can extract robust and useful features from images. Regardless, those feature extractors must be evaluated for a fair comparison of performance, which is the focus of this section.

In the SSL literature, three types of evaluations are commonly used: (i) fine-tuning the entire model, (ii) linear evaluation, also known as linear probing or linear protocol, and (iii) K-nearest neighbors (KNN) evaluation using extracted features. A further distinction can be made based on the dataset the trained model is evaluated on: either (a) on the same dataset, typically ImageNet, that was used for the self-supervised training or (b) on different datasets to test downstream transferability.

\textbf{KNN evaluation}\,\textendash\,After the SSL training is complete, features of the training images are generated from the backbone and matched to their corresponding labels, thus creating a feature bank. Next, predictions are made for the test images based on the KNN labels of this feature bank. For KNN-based evaluation, following~\citet{instdist}, $k=200$ is often used. Although this form of evaluation was popular early on, the majority of recently proposed frameworks evaluate their models using linear evaluation or by fine-tuning.

\textbf{Linear evaluation}\,\textendash\,For this type of evaluation, all trainable parameters (e.g., weights) in the model are frozen and a new linear layer, which maps features to predictions, is introduced to the trained model. Then, only this linear layer is trained on the training set to achieve an optimal performance.

\textbf{Fine-tuning}\,\textendash\,For this type of evaluation, once again, a linear layer is introduced to the SSL-trained model which maps features to predictions. Then, in a supervised fashion, the entire model is (re)trained on the training dataset, after which an evaluation is performed on the test/validation set.

\textbf{Benchmarks}\,\textendash\,In order to provide an aggregate view of the field, we provide the benchmarking results below.

\begin{figure}[t!]
\centering
\begin{subfigure}{.47\textwidth}
\includegraphics[width=\textwidth]{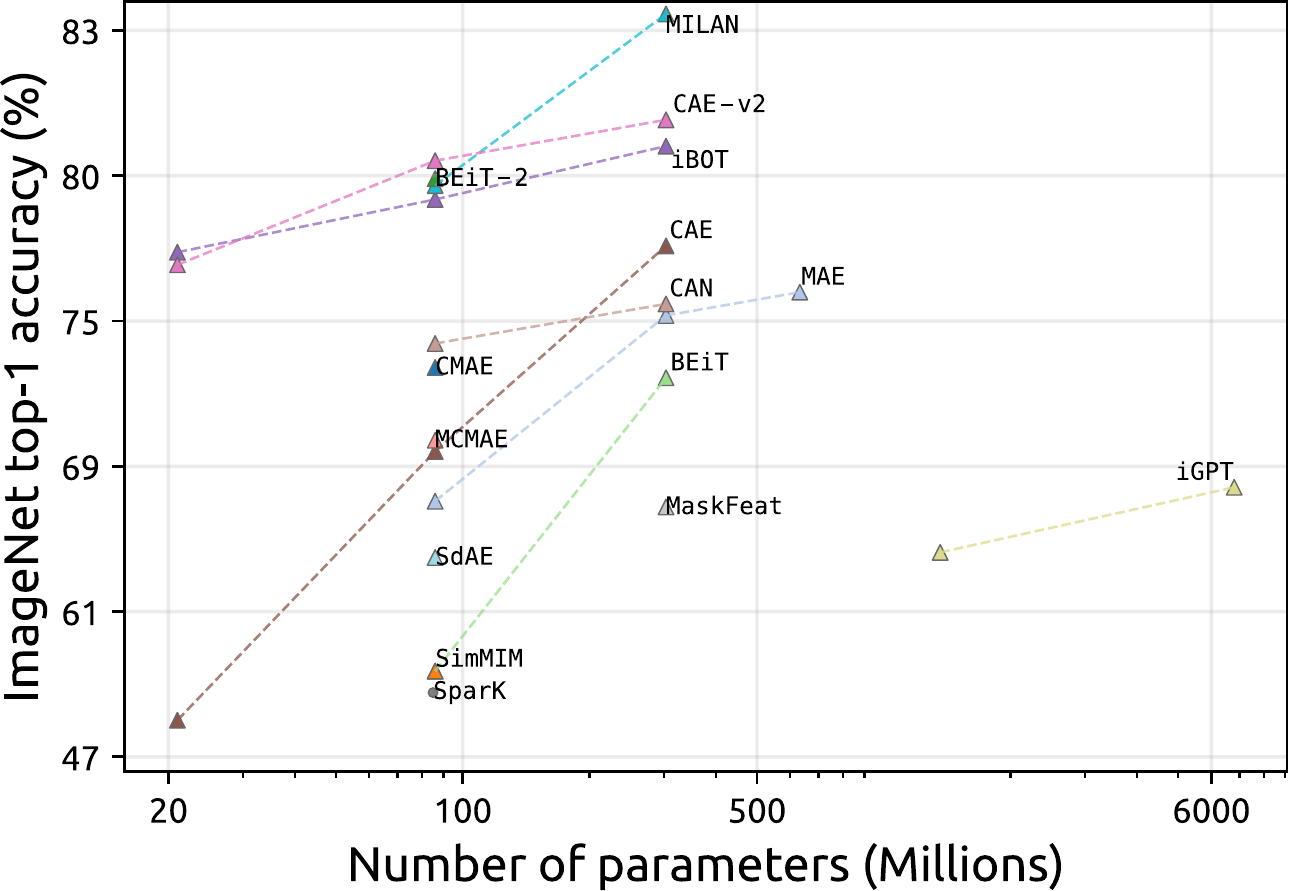}
\caption{Linear probing}
\end{subfigure}
\begin{subfigure}{.48\textwidth}
\includegraphics[width=\textwidth]{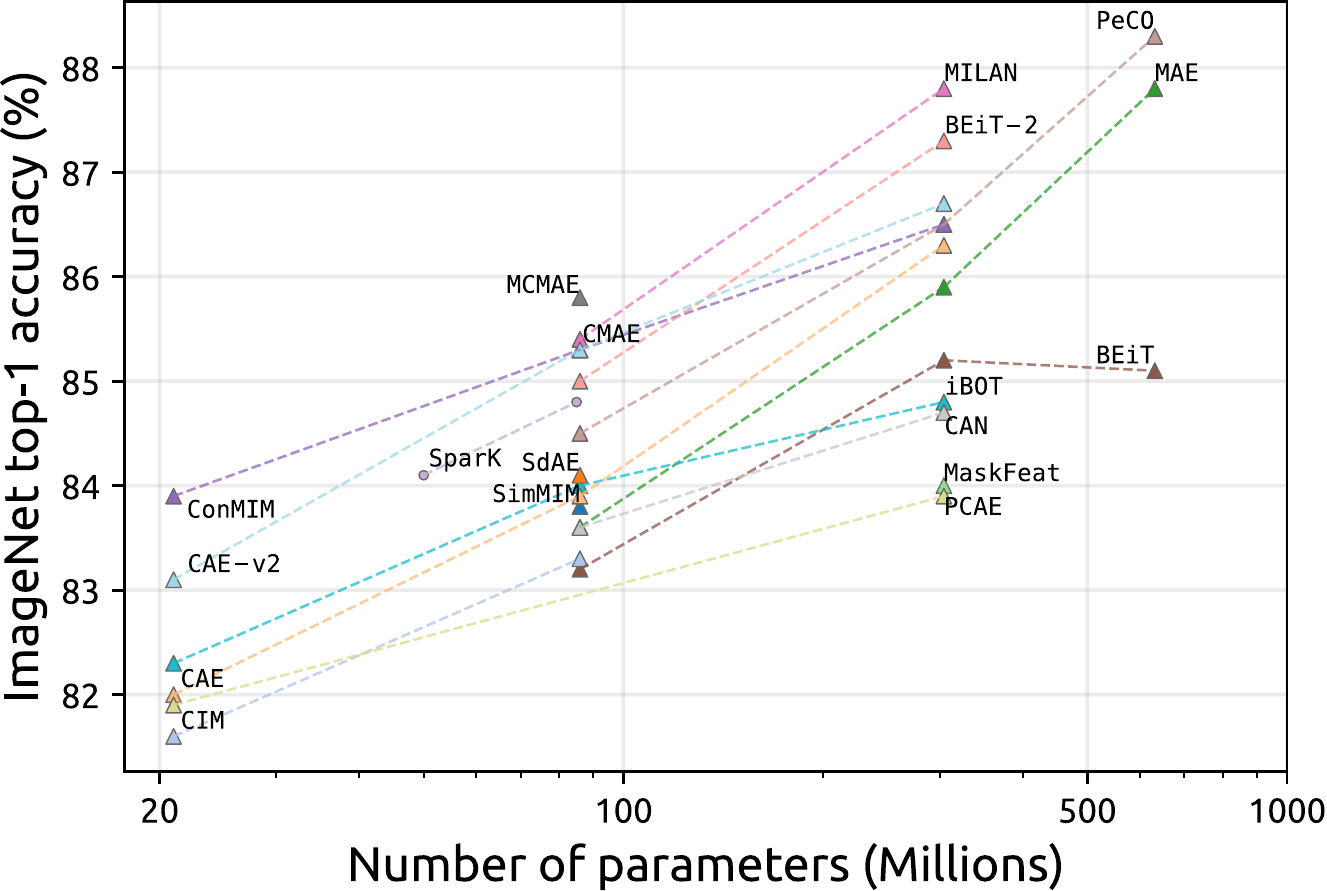}
\caption{Fine-tuning}
\end{subfigure}
\caption{The ImageNet Top-1 accuracy of backbones that are trained with MIM-based generative SSL frameworks are measured using (a) \textbf{linear probing} and (b) \textbf{fine-tuning}. The connecting lines indicate different backbone networks trained with the same framework. In both figures, nodes with circles indicate CNN-based architectures, whereas triangles indicate transformer-based architectures.
}
\label{fig:gen_ssl_acc_vs_param}
\end{figure}

\begin{itemize}
\item For the majority of the discriminative SSL frameworks covered in this survey, we provide a comparison of model size to linear probing accuracy on ImageNet in Figure~\ref{fig:ssl_acc_vs_param}.
\item For MIM-based generative SSL frameworks, we provide a comparison of model size to linear probing and fine-tuning accuracy on ImageNet in Figure~\ref{fig:gen_ssl_acc_vs_param}.
\item From Table~\ref{tbl:discriminative_dataset_usage} to Table~\ref{tbl:generative_dataset_usage}, we provide the datasets used in the respective papers of SSL frameworks. 
\item From Table~\ref{tbl:ssl_benchmark_clustering} to Table~\ref{tbl:ssl_benchmark_mim_based_fine_tun}, we provide benchmarks on ImageNet-1K.
\item In Table~\ref{tbl:ssl_benchmark_COCO}, we provide benchmarks on COCO.
\end{itemize}

\textbf{Evaluation preference for discriminative and generative frameworks}\,\textendash\,A noteworthy observation in the evaluation of self-supervised learning frameworks is that while most discriminative SSL frameworks tend to favor linear evaluation, the majority of generative SSL frameworks tend to prefer fine-tuning. This is primarily due to the poor results obtained with linear probing with generative frameworks that only use MIM as the pretext task (see \texttt{MAE}, \texttt{BEiT}, \texttt{SimMIM}, \texttt{iGPT} in Figure~\ref{fig:gen_ssl_acc_vs_param}). When discriminative elements such as contrastive learning, distillation, or the \texttt{CLIP} tokenizer which is trained contrastively, are used, linear probing accuracy shows a dramatic increase (see \texttt{MILAN}, \texttt{iBOT}, \texttt{CAE-v2}, and \texttt{BEiT-v2} in Figure~\ref{fig:gen_ssl_acc_vs_param}).

\vspace{-0.5em}
\section{Availability and comparability of SSL frameworks}
\label{sec:code}
\vspace{-0.25em}

Most of the frameworks covered in Section~\ref{sec:disc_ssl} perform experiments on ImageNet, COCO~\citep{coco}, and Pascal VOC~\citep{pascal_voc}, thus enabling straightforward benchmarking and comparability. Moreover, many SSL frameworks come with implementations and trained models that are publicly available, contributing to speeding up research on SSL. For example, the availability and the straightforward adoptability of the \texttt{MoCo} framework enabled a number of follow-up studies that used the code of \texttt{MoCo}~\citep{mochi,adco,dense_cl}. For the SSL frameworks covered in this survey, in Table~\ref{tbl:ssl_publication_info}, Table~\ref{tbl:ssl_publication_info_enh}, and Table~\ref{tbl:ssl_publication_info_gen} we provide details on the availability of official implementation as well as trained models.

Apart from the availability of official implementations, the availability of third-party repositories also accelerated the adoption of SSL, enabling unified experimentation. Alas, not all third-party repositories are up to date, and some of them have already been abandoned. In Table~\ref{tbl:ssl_repository}, we provide a number of useful SSL repositories that have been updated within the third quarter of 2022.

\vspace{-0.5em}
\section{Conclusions, current trends, and directions for future research}
\label{sec:conclusions}

In this survey, we reviewed general-purpose frameworks that use images for SSL training, with the goal of bringing interested researchers up to speed with the field of SSL. In what follows, we highlight a number of directions for future research that are ripe for contribution.

\textbf{Theoretical understanding of the requirements of SSL}\,\textendash\,As detailed in Section~\ref{sec:disc_ssl}, the successful implementation of discriminative self-supervised learning frameworks requires several prerequisites. To this end, several studies have investigated the efficacy of these requirements, covering topics such as the necessity of negative samples~\citep{mochi,inter_clr}, the importance of image augmentations~\citep{aug_future_1,aug_future_2}, and architectural tricks~\citep{simsiam}. Furthermore, a number of research efforts have attempted to explain the underlying mechanisms behind collapse avoidance~\citep{garrido2022duality,chenintra_vicreg}. Because they were the earliest self-supervision methods, clustering~\citep{exp4} and contrastive learning~\citep{exp1,exp2,exp3} have received significant attention in terms of theoretical contributions. However, other self-supervision paradigms are areas where theoretical explanations are still lacking and are open for further research efforts.




\textbf{Domain- and task-specific SSL}\,\textendash\,The majority of the frameworks covered in this survey are task-agnostic and evaluate their performance on the ImageNet dataset and a number of various downstream tasks focusing on natural images. However, the effectiveness of these models on natural image datasets may not necessarily generalize to other datasets that contain different image modalities or to other tasks. Therefore, investigating the effectiveness of SSL frameworks that leverage the unique characteristics of data in other domains such as medical imaging~\citep{ramesh2022dissecting,chen2023colo} or other tasks such as classification in the wild~\citep{ssl_wild,dnc}, object detection~\citep{object_ssl,uni_vip}, pose estimation~\citep{chen2023liftedcl}, action detection as well as human-object interaction~\citep{masked_feature_modeling,shah2023halp} represents a relevant area of research.

\textbf{Calibration, interpretability, and adversarial robustness}\,\textendash\,Initial findings suggest that models trained using SSL exhibit distinct properties for robustness and interpretability when compared to models trained via supervised learning~\citep{ssl_calib,ssl_robust}. However, many of the beneficial and detrimental effects of self-supervised training on downstream tasks remain unclear.

\textbf{Efficient SSL}\,\textendash\,The training of SSL models demands a substantial amount of computational resources in comparison to supervised learning. For instance, as reported by~\citet{moco_v3}, the training of \texttt{MoCo-v3} with a vision transformer backbone requires approximately 625 TPU days. Consequently, SSL has significantly increased the computational demands of DNN training. This observation explains why a vast majority of the contributions to the frameworks discussed in this survey have at least one author with an industry affiliation (see Table~\ref{tbl:ssl_publication_info} to Table~\ref{tbl:ssl_publication_info_gen}). Moreover, the majority of these contributions ($>80\%$) come from industry labs such as Facebook AI Research, Microsoft Research, DeepMind, Google Research, SenseTime, ByteDance, and Huawei. To mitigate the high costs of training, researchers have started exploring techniques for efficient training and evaluation~\citep{esvit,efficient_ssl_evaluation,mae}. Despite the progress made, there is still a considerable amount of work to be done in this field.

\textbf{KNN-based evaluation, linear probing, or fine-tuning?}\,\textendash\,As we mentioned in Section~\ref{sec:evaluation}, most generative SSL frameworks prefer to use fine-tuning as the method of evaluation while discriminative frameworks prefer to use linear probing. In favor of those two, KNN-based evaluation has been mostly abandoned. \newline \citet{simsiam} and~\citet{mae} argue that there is no correlation between the accuracy of linear probing and fine-tuning or downstream transferability. ~\citet{mae} further argues that linear probing misses the opportunity to utilize strong but non-linear features, and this sentiment is repeated by a number of follow-up research efforts~\citep{conmim,cae}. The most recent research effort on this topic is the work of~\citep{cl_vs_mim} where the authors argue that models trained with MIM exhibit a bias towards texture whereas contrastive learning leads to a bias towards shape, suggesting that this may be the explanation for the difference in linear probing accuracy. Nevertheless, thorough investigations on SSL model evaluations and convincing explanations for their differences are largely missing.

\textbf{On the usage of tokenizers in SSL}\,\textendash\,From Table~\ref{tbl:reconstruction_tokens} it can be seen that a number of MIM-based generative SSL frameworks make use of a previously trained tokenizer in order to enhance learned representations. In particular,~\citep{beit-v2,mrmae,cae-v2} demonstrate that the usage of \texttt{CLIP} tokenizers (either as-is or as a teacher for distillation) boosts the performance of frameworks considerably. However, using a tokenizer that is trained on a large corpus of images to demonstrate state-of-the-art results against other frameworks that do not have access to this level of supervision has recently attracted criticism from the community~\citep{beit_review}. We believe that the investigation of the efficacy and necessity of tokenizers in generative SSL is an area that is largely unexplored.

\textbf{Moving forward: generative or discriminative SSL?}\,\textendash\,Although recent results obtained with generative SSL seem to favor generative approaches over discriminative ones, it is important to note that generative SSL has not only benefited from the discovery of vision transformers, but also from advancements made in the area of discriminative SSL. The most recent comparative studies suggest that the answer to the discriminative versus generative question is not that straightforward and that both approaches have their own merits and limitations~\citep{cl_vs_mim}. As mentioned, some of the newly proposed generative frameworks also leverage discriminative features such as contrastive learning~\citep{conmim,cae-v2} or distillation~\citep{ibot,beit-v2}. Therefore, we speculate that this trend will continue and that the newly proposed frameworks in the upcoming years will leverage improvements from both sides in order to further improve the state-of-the-art.

\textbf{Cadence of research in SSL and the extent of this survey}\,\textendash\,Before we conclude our survey, we would like to briefly discuss the cadence of research in SSL and the breadth of topics covered in this survey.

It is widely recognized that the field of machine learning has experienced an unprecedented growth in research and development over the past decade, particularly following the groundbreaking results achieved with AlexNet~\citep{Alexnet}. During this period, significant advances have been made not only in the architectural design of deep neural networks, but also in optimizers, training routines, normalization techniques, data augmentation, and various other areas. While these improvements have steadily advanced the state-of-the-art in supervised learning, self-supervised learning has also benefited from the majority of these advancements from the beginning. Thanks to the integration of these advancements into SSL from its early stages, as well as the availability of computational resources, the state-of-the-art in SSL has improved rapidly since 2018. The pace of research has been so fast that some frameworks have been improved upon even before their predecessors got published (e.g., \texttt{CAE}~\citep{cae} and \texttt{CAE-v2}~\citep{cae-v2}, \texttt{BEiT-v2}~\citep{beit-v2} and \texttt{BEiT-v3}~\citep{beit-v3}).

Given the aforementioned observation, to cover the most up-to-date research efforts, we have decided to include papers that have not yet been published in conference proceedings or journals. Consequently, the majority of the papers cited in this survey, from $2021$ onward, are preprints. At the time of submitting this survey, the latest conference surveyed for relevant papers was the 11th International Conference on Learning Representations (ICLR), held in May 2023, from which a portion of the papers cited in this survey have been rejected, withdrawn, or published~\citep{wangmosaic,cim,cl_vs_mim,cae,beit-v2,chenintra_vicreg,mrmae,can,pcae,enh_ts,enh_arcl,chen2023liftedcl,cl_vs_mim}.

\vspace{-0.5em}
\section*{Funding Information}
\vspace{-0.5em}
This work was supported by a grant from the Special Research Fund (BOF) of Ghent University (BOF/STA/202109/039).

\clearpage
\appendix

{\centering
\textbf{\Large{Appendix for:}}\\
\textbf{\Large{Know Your Self-supervised Learning:}}\\
\textbf{\Large{A Survey on Image-based Generative and Discriminative Training}}\\
}
\phantom{-}
\\
The content of appendix is detailed below.
\begin{itemize}
    \item \textbf{A list of abbreviations} is provided in Section~\ref{sec:abbrv}.
    
    \item \textbf{Metadata} for the frameworks such as the primary affiliation of authors, publication date, and source code as well as availability of trained models are provided for:
        \begin{itemize}
        \item Discriminative SSL frameworks in Table~\ref{tbl:ssl_publication_info}
        \item Enhancements to existing SSL frameworks in Table~\ref{tbl:ssl_publication_info_enh}
        \item Generative SSL frameworks in Table~\ref{tbl:ssl_publication_info_gen}
        \end{itemize}

    \item \textbf{Repositories} that are useful for vision-based SSL are listed in Table~\ref{tbl:ssl_repository}.

    \item \textbf{Datasets} used for the evaluation in the respective papers of the frameworks are provided for:
        \begin{itemize}
        \item Discriminative SSL frameworks in Table~\ref{tbl:discriminative_dataset_usage}
        \item Enhancements to discriminative SSL frameworks in Table~\ref{tbl:enhancement_dataset_usage}
        \item Generative SSL frameworks in Table~\ref{tbl:generative_dataset_usage}
        \end{itemize}

    \item \textbf{Benchmarks} on ImageNet-1K are provided for:
    \begin{itemize}
        \item Clustering-based SSL frameworks in Table~\ref{tbl:ssl_benchmark_clustering}
        \item Contrastive-learning-based SSL frameworks in Table~\ref{tbl:ssl_benchmark_contrastive}
        \item Distillation-based SSL frameworks in Table~\ref{tbl:ssl_benchmark_distillation}
        \item Information-maximization-based SSL frameworks in Table~\ref{tbl:ssl_benchmark_infomax}
        \item Enhancements to discriminative SSL frameworks in Table~\ref{tbl:ssl_benchmark_enhancements}
        \item \texttt{GAN}-based SSL frameworks in Table~\ref{tbl:ssl_benchmark_gan_based}
        \item MIM-based SSL frameworks in Table~\ref{tbl:ssl_benchmark_mim_based_fine_tun}
    \end{itemize}

    \item \textbf{Benchmarks} on COCO for all frameworks are provided in Table~\ref{tbl:ssl_benchmark_COCO}.
\end{itemize}

\clearpage
\section{List of abbreviations}
\label{sec:abbrv}

\begin{minipage}{0.5\textwidth}
{\scriptsize
\underline{\footnotesize Clustering frameworks}
\begin{longtable}[l]{ l l }
\texttt{SeLa} & Self labeling \\
\texttt{SCAN} & Semantic clustering by adopting nearest neighbors \\
\texttt{Swav} & Swapping assignments between multiple \\
~ & views of the same image \\
\texttt{ODC} & Online deep clustering \\
\texttt{CoKe} & Constrained K-means \\
\end{longtable}

\underline{\footnotesize Contrastive frameworks}
\begin{longtable}[l]{ l l }
\texttt{InstDist} & Instance discrimination \\
\texttt{NPID} & Non-parametric instance discrimination \\
\texttt{CPC} & Contrastive predictive coding \\
\texttt{DIM} & Deep InfoMax \\
\texttt{AMDIM} & Augmented multi-scale \texttt{DIM} \\
\texttt{CMC} & Contrastive multi-view coding  \\
\texttt{MoCo} & Momentum contrast \\
\texttt{PIRL} & Pretext-invariant representation learning \\
\texttt{SimCLR} & A simple framework for contrastive learning \\
\texttt{PCL} & Prototypical contrastive learning \\
\texttt{PIC} & Parametric instance classification \\
\texttt{DCL} & Debiased contrastive learning \\
\texttt{LooC} & Leave-one-out contrastive learning\\
\texttt{G-SimCLR} & Self-supervised contrastive learning with \\
~ & guided projection \\
\texttt{ReLIC} & Representation learning via\\
~ & invariant causal mechanisms \\
\texttt{AdCo} & Adversarial contrast\\
\texttt{DenseCL} & Dense contrastive learning \\
\texttt{PixPro} & Pixel-level consistency propagation \\
\texttt{CLSA} & Contrastive learning with \\
~ & stronger augmentations \\
\texttt{NNCLR} & Nearest-neighbor contrastive learning \\
~ & of visual representations \\
\texttt{MoBY} & \texttt{\blu{Mo}Co} + \texttt{\blu{BY}OL} \\
\texttt{DNC} & Divide and contrast \\
\texttt{ReSSL} & Relational self-supervised learning \\
\texttt{UniGrad} & A unified gradient framework \\
\texttt{SimCo} & Simplified \texttt{MoCo} without momentum\\
\texttt{SimMoCo} & Simplified \texttt{MoCo} \\
\texttt{UniVIP} & A unified framework for self-supervised\\
~ & visual pre-training \\
\texttt{Mugs} &  Multi-granular self-supervised learning\\
\texttt{CaCo} & Cooperative-adversarial contrastive learning \\
\texttt{SMoG} & Synchronous momentum grouping \\
\texttt{SiameseIM} & Siamese image modelling \\

\end{longtable}

\underline{\footnotesize Distillation frameworks}
\begin{longtable}[l]{ l l }
\texttt{BYOL} & Build your-own latent \\
\texttt{SimSiam} & Simple Siamese representation learning networks \\
\texttt{OBoW} & Online bag-of-visual-words \\
\texttt{DirectPred} & Direct linear predictor \\
\texttt{SEED} & Self-supervised distillation for visual representation \\
\texttt{DisCo} & Distilled contrastive learning \\
\texttt{DINO} & Self-distillation with no labels \\
\texttt{EsVit} & Efficient self-supervised vision transformer \\
\texttt{BINGO} & Bag of instances aggregation \\
\texttt{TinyMIM} & Tiny \texttt{MIM} \\

\end{longtable}

\underline{\footnotesize Information-maximization frameworks}
\begin{longtable}[l]{ l l }
\texttt{WMSE} & Whitening mean squared error \\
\texttt{VicReg} & Variance-invariance-covariance regularization \\
\texttt{TWIST} & Twin class distribution estimation \\
\texttt{TLDR} & Twin learning for dimensionality reduction \\
\texttt{ARB} & Align representations with base \\
\texttt{VicRegL} & \texttt{VicReg} with local visual features\\

\end{longtable}
}
\end{minipage}
\begin{minipage}{0.5\textwidth}
{\scriptsize
\underline{\footnotesize Enhancement modules}
\begin{longtable}[l]{ l l }
\texttt{InfoMin} & Mutual information principle \\
\texttt{InterCLR} & Inter-image contrastive learning \\
\texttt{HEXA}  & Hard examples \\
\texttt{MocHi} & Mixing of contrastive hard negatives\\
\texttt{Resim} & Region similarity representation learning \\
\texttt{MSF} & Mean shift for self-supervised learning \\
\texttt{ORL} & Object-level representation learning  \\
\texttt{CEB} & Conditional entropy bottleneck \\
\texttt{SEM} & Simplicial embeddings \\
\texttt{ENS} & Ensemble self-supervised learning \\
\texttt{MRCL} & Masked reconstruction contrastive learning\\
\texttt{TS} & Temperature schedules\\
\texttt{ARCL} & Augmentation-robust contrastive learning \\
\texttt{MosRep} & Mosaic representation learning framework \\

\end{longtable}

\underline{\footnotesize GAN-based frameworks}
\begin{longtable}[l]{ l l }
\texttt{BiGAN} & Bidirectional \texttt{GAN} \\

\texttt{ALI} & Adversarially learned inference \\

\texttt{BigBiGAN} & \texttt{BiGAN} with \texttt{BigGAN} generator\\

\texttt{SS-GAN} & Self-supervised \texttt{GAN} \\

\texttt{SS-GAN-LA} & \texttt{SS-GAN} with label augmentation\\

\texttt{VQGAN} & Vector-quantized \texttt{GAN} \\

\texttt{Vit-VQGAN} & ViT-based \texttt{VQGAN} \\
\end{longtable}

\underline{\footnotesize MIM-based frameworks}
\begin{longtable}[l]{ l l }
\texttt{iGPT} &  Image \texttt{GPT}\\
\texttt{BEiT} &  Bidirectional encoder representation from\\
~ & image transformers\\
\texttt{MAE} &  Masked autoencoders\\
\texttt{iBOT} &  Image \texttt{BERT} pre-training with online tokenizer\\
\texttt{SimMIM} &  A simple framework for MIM \\
\texttt{PeCO} &  Perceptual codebook \\ 
\texttt{MaskFeat} &  Masked feature prediction \\
\texttt{CAE} &  Context autoencoder \\
\texttt{CIM} &  Corrupted image modeling\\
\texttt{MCMAE} &  Masked convolution meets \texttt{MAE}\\
\texttt{ConMIM} &  Denoising contrast masked image modeling \\
\texttt{CMAE} &  Contrastive \texttt{MAE}\\
\texttt{SdAE} &  Self-distillated masked autoencoder\\
\texttt{MILAN} & Masked image pretraining on\\
~ & language assisted representation \\
\texttt{CAN} & Contrastive learning, masked autoencoders,\\
~ & and noise prediction \\
\texttt{PCAE} & Progressively compressed autoencoder\\
\texttt{SparK} & Sparse masked modeling\\
\texttt{MRMAE} & Mimic before reconstruct \texttt{MAE}\\
\end{longtable}

\underline{\footnotesize Others}
\begin{longtable}[l]{ l l }
\texttt{BERT} &  Bidirectional encoder representations from transformers \\
\texttt{GPT} & Generative pre-trained transformer \\
\phantom{--} \\
\phantom{--} \\
\phantom{--} \\
\phantom{--} \\
\phantom{--} \\
\phantom{--} \\
\phantom{--} \\
\phantom{--} \\
\phantom{--} \\
\phantom{--} \\
\phantom{--} \\
\phantom{--} \\
\phantom{--} \\
\phantom{--} \\

\end{longtable}
}
\end{minipage}

\clearpage
\section{Metadata for frameworks}

\begin{table}[htbp!]
\centering
\caption{Publication information as well as implementation details for discriminative SSL frameworks covered in this survey.}
\tiny
\begin{tabular}{ccccccc}
\cmidrule[1pt]{1-7}
SSL & Primary & Publication & Experiments on  & Downstream  & Official & Trained  \\
framework & affiliation & date & on ImageNet 1K & experiments & implementation & models \\
\cmidrule[0.5pt]{1-7}

\texttt{Deep Cluster} & Facebook AI Research & Mar 2019 & Yes & Yes & \href{https://github.com/facebookresearch/deepcluster}{Available} & \href{https://github.com/facebookresearch/deepcluster}{Available} \\

\texttt{Local Aggregation} & Stanford University & Apr 2019 & Yes & Yes & \href{https://github.com/neuroailab/LocalAggregation}{Available} & \red{Not available} \\

\texttt{Deeper Cluster} & Facebook AI Research & Aug 2019 & Yes & Yes & \href{https://github.com/facebookresearch/DeeperCluster}{Available} & \href{https://github.com/facebookresearch/DeeperCluster}{Available} \\

\texttt{SeLa} & University of Oxford & Nov 2019 & Yes & Yes & \href{https://github.com/yukimasano/self-label}{Available}  & \href{https://github.com/yukimasano/self-label}{Available} \\

\texttt{SCAN} & KU Leuven & Jul 2020 & Yes & \red{No} & \href{https://github.com/wvangansbeke/Unsupervised-Classification}{Available} & \href{https://github.com/wvangansbeke/Unsupervised-Classification}{Available} \\

\texttt{Deep Cluster-v2} & Facebook AI Research & Jun 2020 & Yes & Yes & \href{https://github.com/facebookresearch/swav}{Available} & \href{https://github.com/facebookresearch/swav}{Available} \\
\end{tabular}
\label{tbl:ssl_publication_info}
\end{table}

\clearpage

\begin{table}[t!]
\centering
\caption{Publication information as well as implementation details for enhancements proposed to existing SSL frameworks covered in this survey.}
\scriptsize
\begin{tabular}{ccccccc}
\cmidrule[1pt]{1-7}
SSL & Primary & Publication & Experiments on  & Downstream  & Official & Trained  \\
framework & affiliation & date & on ImageNet 1K & experiments & implementation & models \\
\cmidrule[0.5pt]{1-7}

\texttt{InfoMin} & MIT & May 2020 & Yes & Yes & \href{https://github.com/HobbitLong/PyContrast}{Available} & \href{https://github.com/HobbitLong/PyContrast}{Available} \\

\texttt{InterCLR} & Nanyang Technological Univ. & Aug 2020 & Yes & Yes & \red{Not available} & \red{Not available}  \\

\texttt{HEXA} & Microsoft Research & Dec 2020 & Yes & Yes & \red{Not available} & \red{Not available}  \\

\texttt{MocHi} & Naver Labs EU & Oct 2020 & Yes & Yes & \red{Not available} & \href{https://europe.naverlabs.com/research/computer-vision/mochi/#pretrainedmodels}{Available} \\

\texttt{ReSim} & UC Berkeley & Mar 2021 & Yes & Yes & \href{https://github.com/Tete-Xiao/ReSim}{Available} & \href{https://github.com/Tete-Xiao/ReSim}{Available} \\

\texttt{MSF} & University of Maryland & May 2021 & Yes & Yes & \href{https://github.com/UMBCvision/MSF}{Available} & \href{https://github.com/UMBCvision/MSF}{Available} \\
\end{tabular}
\label{tbl:ssl_publication_info_enh}
\end{table}

\begin{table}[t!]
\centering
\caption{Publication information as well as implementation details for generative SSL frameworks covered in this survey.}
\scriptsize
\begin{tabular}{ccccccc}
\cmidrule[1pt]{1-7}
SSL & Primary & Publication & Experiments on  & Downstream  & Official & Trained  \\
framework & affiliation & date & on ImageNet 1K & experiments & implementation & models \\
\cmidrule[0.5pt]{1-7}

\texttt{BiGAN} & University of California & May 2016 & Yes & Yes & \href{https://github.com/jeffdonahue/bigan}{Available} & \href{https://github.com/jeffdonahue/bigan}{Available}\\

\texttt{BigBiGAN} & DeepMind &  Jul 2019  & Yes & Yes & \href{https://colab.research.google.com/github/tensorflow/hub/blob/master/examples/colab/bigbigan_with_tf_hub.ipynb}{Available} & \href{https://tfhub.dev/s?publisher=deepmind&q=bigbigan}{Available}\\

\texttt{ALI} & MILA & Jun 2016  & No & Yes & \href{https://github.com/IshmaelBelghazi/ALI}{Available} & \href{https://github.com/IshmaelBelghazi/ALI}{Available}\\

\texttt{SS-GAN} & University of California & Nov 2018  & Yes & Yes & \href{https://github.com/google/compare_gan/tree/master/compare_gan}{Available} & \red{Not available} \\

\texttt{SS-GAN-LA} & University of Chinese AoS & Oct 2021 & \red{No} & Yes & \href{https://github.com/liang-hou/ssgan-la}{Available} & \href{https://github.com/liang-hou/ssgan-la}{Available}\\

\texttt{Vit-VQGAN} & Google Research &  Oct 2021  & Yes & Yes & \red{Not available} & \red{Not available}\\
\end{tabular}
\label{tbl:ssl_publication_info_gen}
\end{table}

\begin{table}[t!]
\centering
\caption{Github repositories related to SSL, their maintainer, and purpose.}
\scriptsize
\begin{tabular}{lll}
\cmidrule[1pt]{1-3}
Repository name & Maintainer & Purpose \\
\cmidrule[0.5pt]{1-3}

\href{https://github.com/jason718/awesome-self-supervised-learning}{Awesome SSL}  & Independent & A comprehensive reading list for SSL \\

\href{https://github.com/vturrisi/solo-learn}{solo-learn} & Independent~\citep{solo_learn} & SSL frameworks, benchmarking, and model zoo \\

\href{https://github.com/facebookresearch/vissl}{VISSL} & Facebook~\citep{vissl} & SSL frameworks, benchmarking, and model zoo \\

\href{https://github.com/open-mmlab/mmselfsup}{MMSelfSup}  & OpenMMLab~\citep{mmselfsup} & SSL frameworks, benchmarking, and model zoo  \\

\href{https://github.com/lightly-ai/lightly}{Lightly} & Lightly.ai~\citep{lightly} & SSL frameworks and benchmarking \\

\href{https://github.com/alibaba/EasyCV}{EasyCV} & Alibaba~\citep{easy_cv} & SSL frameworks and benchmarking \\

\href{https://github.com/microsoft/Semi-supervised-learning}{Unified SSL Benchmark} & Microsoft~\citep{usb_microsoft} & SSL frameworks and benchmarking \\

\cmidrule[1pt]{1-3}
\end{tabular}
\label{tbl:ssl_repository}
\end{table}

\clearpage

\section{Framework dataset usage}

\begin{table}[htbp!]
\centering
\caption{Employed datasets for experiments used in the research articles of the discriminative SSL frameworks. The third column summarizes the evaluated tasks in the respective papers of frameworks: (C)lassification, (D)etection/localization, and (S)egmentation.}
\tiny
\begin{tabular}{rll}
\cmidrule[1pt]{1-3}
SSL & Used & \multirow{2}{*}{Tasks} \\
framework & datasets & ~ \\
\cmidrule[0.5pt]{1-3}

\texttt{Deep Cluster} & ImageNet-1k, Pascal VOC, Places, YFCC100M & C, D, S\\

\texttt{Local Aggregation} &  ImageNet-1k, Pascal VOC & C, D\\

\texttt{Deeper Cluster} & ImageNet-1k, Pascal VOC, Places & C, D, S  \\ 

\texttt{SeLa} &  ImageNet-1k, CIFAR-100, CIFAR-10, Pascal VOC, SVHN & C, D\\

\texttt{SCAN} &  ImageNet-1k, CIFAR-100, CIFAR-10, STL-10 & C \\

\texttt{Deep Cluster-v2} &   ImageNet-1k  & C\\
\end{tabular}
\label{tbl:discriminative_dataset_usage}
\end{table}

\begin{table}[htbp!]
\centering
\caption{Employed datasets for experiments used in the research articles of the enhancements to discriminative SSL frameworks. The third column summarizes the evaluated tasks in the respective papers of frameworks: (C)lassification, (D)etection/localization, and (S)egmentation.}
\scriptsize
\begin{tabular}{rll}
\cmidrule[1pt]{1-3}
SSL & Used & \multirow{2}{*}{Tasks} \\
framework & datasets & ~  \\
\cmidrule[0.5pt]{1-3}

\texttt{InfoMin} & ImageNet-1k, COCO, Pascal VOC, Colorful Moving-MNIST & C, D, S \\

\texttt{InterCLR} &  ImageNet-1k, Pascal VOC, Places & C, D\\

\texttt{HEXA} &  ImageNet-1k, CIFAR-100, CIFAR-10, Pascal VOC & C, D \\

\texttt{MocHi} & ImageNet-1k, ImageNet-100, COCO, Pascal VOC & C, D, S\\

\texttt{ReSim} & ImageNet-1k, ImageNet-100, COCO, Pascal VOC & D, S \\ 

\multirow{2}{*}{\texttt{MSF}} & ImageNet-1k, CIFAR-100, CIFAR-10, Pascal VOC, Food, SUN397, Cars & \multirow{2}{*}{C, D}\\ 
~ & Aircraft, DTD, Pets, Caltech-101, Flower & ~ \\

\texttt{ORL} & ImageNet-1k, COCO, Pascal VOC, Places, iNat & C, D, S\\

\multirow{2}{*}{\texttt{CEB}} &  ImageNet-1k, ImageNet-A, ImageNet-C, ImageNet-R & \multirow{2}{*}{C} \\
~ & ImageNet-v2, ImageNet-Vid, YouTube-BB, ObjectNet & ~ \\

\multirow{2}{*}{\texttt{SEM}} &  ImageNet-1k, ImageNet-A, ImageNet-C, ImageNet-R & \multirow{2}{*}{C} \\
~ & ImageNet-v2, CIFAR-100, CIFAR-10, Food, Sun397, DTD, Flower  & ~ \\

\multirow{2}{*}{\texttt{ENS}} & ImageNet-1k, CIFAR-100, CIFAR-10, Food, SUN397, Cars & \multirow{2}{*}{C} \\
~ & DTD, Pets, Caltech-101, Flower & ~ \\

\texttt{MRCL} &  ImageNet-1k, COCO, ADE20k & C, D, S\\

\texttt{TS} &  ImageNet-100, CIFAR-100, CIFAR-10 & C \\

\multirow{2}{*}{\texttt{ARCL}} &  ImageNet-1k,  CIFAR-100, CIFAR-10,  Food,  SUN397 & \multirow{2}{*}{C}\\ 
~ & Cars, Aircraft, DTD, Pets, Caltech-101, Flower & ~ \\

\multirow{2}{*}{\texttt{MosRep}} & ImageNet-1k, ImageNet-100,  CIFAR-100, CIFAR-10, COCO & \multirow{2}{*}{C, D, S} \\
~ & Food, Cars, DTD, Pets, Caltech-101, Flower & ~ \\

\cmidrule[1pt]{1-3}
\end{tabular}
\label{tbl:enhancement_dataset_usage}
\end{table}

\begin{table}[htbp!]
\centering
\caption{Employed datasets for experiments used in the research articles of the generative SSL frameworks. The third column summarizes the evaluated tasks in the respective papers of frameworks: (C)lassification, (D)etection/localization, and (S)egmentation.}
\scriptsize
\begin{tabular}{rll}
\cmidrule[1pt]{1-3}
SSL & Used & \multirow{2}{*}{Tasks} \\
framework & datasets & ~  \\
\cmidrule[0.5pt]{1-3}

\texttt{BigBiGAN} & ImageNet-1k & C\\

\texttt{BiGAN} &  ImageNet-1k, Pascal VOC, MNIST &  C, D, S\\

\texttt{ALI} &  Tiny ImageNet, CIFAR-10, SVHN, CelebA & C\\

\texttt{SS-GAN} & ImageNet-1k, CIFAR-10, CelebA-HQ, LSUN-Bedroom & C\\

\texttt{SS-GAN-LA} & Tiny-ImageNet, CIFAR-10, STL-10, CelebA & C \\

\texttt{Vit-VQGAN} &  ImageNet-1k, CelebA-HQ, FFHQ & C \\
\end{tabular}
\label{tbl:generative_dataset_usage}
\end{table}

\clearpage

\section{ImageNet benchmarks}

\begin{table}[htbp!]
\centering
\caption{ImageNet-1k linear probing and fine-tuning benchmarks for \textbf{clustering}-based SSL frameworks.}
\scriptsize
\begin{tabular}{cccccc}
\cmidrule[1pt]{1-6}
SSL & Backbone & SSL & Fine-tuning & Linear probing & Additional \\
framework & network & epochs & accuracy & accuracy & notes \\
\cmidrule[0.5pt]{1-6}

\texttt{Deep Cluster} & AlexNet & 500 & - & 39.8 &  Used conv4 output \\

\cmidrule[0.5pt]{1-6}

\texttt{LA} & ResNet-50 & 200 & - & 60.2 & - \\


\cmidrule[0.5pt]{1-6}

\texttt{Deeper Cluster} & VGG-16 & 100 & - & 48.4 & -\\

\cmidrule[0.5pt]{1-6}

\texttt{SeLA} & ResNet-50 & 90 & - & 61.5 & - \\

\cmidrule[0.5pt]{1-6}

\texttt{SCAN} & ResNet-50 & 90 & - & 39.9 & Unsupervised evaluation \\

\cmidrule[0.5pt]{1-6}

\texttt{Deep Cluster-v2} & ResNet-50 & 400 & - & 74.3 &   2x160 + 4x96 crops \\
\end{tabular}
\label{tbl:ssl_benchmark_clustering}
\end{table}

\begin{table}[htbp!]
\centering
\caption{ImageNet-1k linear probing and fine-tuning benchmarks for \textbf{contrastive-learning}-based SSL frameworks.}
\scriptsize
\begin{tabular}{cccccc}
\cmidrule[1pt]{1-6}
SSL & Backbone & SSL & Fine-tuning & Linear probing & Additional \\
framework & network & epochs & accuracy & accuracy & notes\\
\cmidrule[0.5pt]{1-6}

\texttt{InstDist (NPID)} & ResNet-50 & 200 & - & 54.0 & Using conv5 \\

\cmidrule[0.5pt]{1-6}

\texttt{CPC} & ResNet-v2 101 & 130 & - & 48.7 & - \\

\cmidrule[0.5pt]{1-6}

\texttt{CPCv2} & ResNet-50 & 200 & - & 61.8 & - \\

\cmidrule[0.5pt]{1-6}

\texttt{AMDIM} & ResNet-50 & 150 & - & 68.1 & Large \texttt{AMDIM} model \\

\cmidrule[0.5pt]{1-6}

\texttt{CMC} & ResNet-50 & 200 & - & 66.2 &  RandAugment \\

\cmidrule[0.5pt]{1-6}

\texttt{MoCo} & ResNet-50 & 200 & 77.3 & 60.6 & - \\
\end{tabular}
\label{tbl:ssl_benchmark_contrastive}
\end{table}

\begin{table}[htbp!]
\centering
\caption{ImageNet-1k linear probing and fine-tuning benchmarks for \textbf{distillation}-based SSL frameworks.}
\scriptsize
\begin{tabular}{ccccccc}
\cmidrule[1pt]{1-6}
SSL & Backbone & SSL & Fine-tuning & Linear probing & Additional \\
framework & network & epochs & accuaracy & accuracy & notes\\
\cmidrule[0.5pt]{1-6}

\texttt{BYOL} & ResNet-50 & 1000 & 77.7 & 74.3 &  - \\
\texttt{BYOL} & ResNet-50 ($\times 2$) & 1000 & - & 77.4 &  - \\
\texttt{BYOL} & ResNet-50 ($\times 4$) & 1000 & - & 78.6 &  - \\
\texttt{BYOL} & ResNet-200 ($\times 2$) & 1000 & - & 79.6 &   - \\

\cmidrule[0.5pt]{1-6}

\texttt{SimSiam} & ResNet-50 & 800 & - & 71.3  & - \\

\cmidrule[0.5pt]{1-6}

\texttt{OBoW} & ResNet-50 & 200 & - & 73.8   & - \\

\cmidrule[0.5pt]{1-6}

\texttt{DirectPred} & ResNet-50 & 300 & - & 72.4   & - \\

\cmidrule[0.5pt]{1-6}

\texttt{SEED} & ResNet-34 & 800 & - & 58.5 & Distilled from ResNet-50 \\

\cmidrule[0.5pt]{1-6}

\texttt{DisCo}  & ResNet-34 & 800 & - & 62.5  & Distilled from ResNet-50 \\

\cmidrule[0.5pt]{1-6}

\texttt{DINO} & ResNet-50 & 300 & - & 75.3  & - \\
\texttt{DINO} & ViT-S & 300 & 81.5 & 77.0  & - \\
\texttt{DINO} & ViT-B & 300 & 82.8 & 78.2  & - \\

\cmidrule[0.5pt]{1-6}

\texttt{EsViT} & Swin-T & 300 & - & 78.1 & - \\
\texttt{EsViT} & Swin-S & 300 & - & 79.5 & - \\
\texttt{EsViT} & Swin-B & 300 & - & 80.4 & - \\

\cmidrule[0.5pt]{1-6}

\texttt{BINGO} & ResNet-34 & 200 & - & 66.1 & Distilled from ResNet-50 \\

\cmidrule[0.5pt]{1-6}

\texttt{TinyMIM} & ViT-S & 300 & 83.0 & - & Distilled from Vit-B \\
\texttt{TinyMIM} & ViT-B & 300 & 85.0 & - & Distilled from Vit-L \\

\cmidrule[1pt]{1-6}
\end{tabular}
\label{tbl:ssl_benchmark_distillation}
\end{table}

\begin{table}[htbp!]
\centering
\caption{ImageNet-1k linear probing and fine-tuning benchmarks for \textbf{information-maximization}-based SSL frameworks.}
\scriptsize
\begin{tabular}{cccccc}
\cmidrule[1pt]{1-6}
SSL & Backbone & SSL & Fine-tuning & Linear probing & Additional \\
framework & network & epochs & accuracy & accuracy & notes\\
\cmidrule[0.5pt]{1-6}

\texttt{WMSE} & ResNet-50 & 100 & - & 69.4 & $d=4$, corresponding to 6 positive pairs \\
\texttt{WMSE} & ResNet-50 & 400 & - & 72.5 & $d=4$, corresponding to 6 positive pairs \\

\cmidrule[0.5pt]{1-6}

\texttt{Barlow Twins} & ResNet-50 & 1000 & - & 73.2 & - \\

\cmidrule[0.5pt]{1-6}

\texttt{VicReg} & ResNet-50 & 1000 & - & 73.2 &  - \\

\cmidrule[0.5pt]{1-6}

\texttt{TWIST} & ResNet-50 & 800 & - & 75.5 &  Multi-crop \\
\texttt{TWIST} & ViT-B/16 & 300 & 82.8 & 78.4 &  - \\

\cmidrule[0.5pt]{1-6}

\texttt{TLDR} & ViT-S/16 & 100 & - & 74.8 & - \\

\cmidrule[0.5pt]{1-6}

\texttt{ARB} & ResNet-50 & 100 & - & 68.2 &  - \\

\cmidrule[0.5pt]{1-6}

\texttt{VicRegL} & ConvNext-S & 150 & - & 75.9 & - \\
\texttt{VicRegL} & ConvNext-B & 150 & - & 77.1 & - \\

\cmidrule[1pt]{1-6}
\end{tabular}
\label{tbl:ssl_benchmark_infomax}
\end{table}

\begin{table}[htbp!]
\centering
\caption{ImageNet-1k linear probing and fine-tuning benchmarks for \textbf{enhancements} to discriminative SSL frameworks.}
\scriptsize
\begin{tabular}{cccccc}
\cmidrule[1pt]{1-6}
SSL & Backbone & SSL & Fine-tuning & Linear probing & Additional \\
framework & network & epochs & accuracy & accuracy & notes\\
\cmidrule[0.5pt]{1-6}

\texttt{InfoMin} & ResNet-50 & 800 & - & 73.0 & - \\
\cmidrule[0.5pt]{1-6}

\texttt{InterCLR} & ResNet-50 + NPID-v2 & 1000 & - & 69.6 & - \\
\texttt{InterCLR} & ResNet-50 + BYOL & 1000 & - & 74.5 & - \\
\cmidrule[0.5pt]{1-6}

\texttt{HEXA} &  ResNet-50 + MoCo-v2 & 800 & 75.7 & 71.7 & - \\
\texttt{HEXA} & ResNet-50 + DeepCluster-v2 & 800 & 78.3 & 75.5 & 8-crops \\
\cmidrule[0.5pt]{1-6}

\texttt{MocHi} & ResNet-50 + MoCo-v2 & 1000 & - & 70.6 & - \\
\end{tabular}
\label{tbl:ssl_benchmark_enhancements}
\end{table}

\begin{table}[htbp!]
\centering
\caption{ImageNet-1k linear probing and fine-tuning benchmarks for \textbf{GAN}-based SSL frameworks.}
\scriptsize
\begin{tabular}{cccccc}
\cmidrule[1pt]{1-6}
SSL & Backbone & SSL & Fine-tuning & Linear probing & Additional \\
framework & network & epochs & accuracy & accuracy & notes\\
\cmidrule[0.5pt]{1-6}
\texttt{BiGAN}  & BB & 400 & - & 56.2  & - \\

\cmidrule[0.5pt]{1-6}

\texttt{BigBiGAN}  & ResNet-50 & 800 & 76.3 & 55.4  & - \\
\texttt{BigBiGAN}  & ResNet-50 ($\times 4$) & 800 & 76.6 & 60.8 & -\\

\cmidrule[0.5pt]{1-6}

\texttt{SS-GAN}  & ResNet & 80 & - & 38  & -\\

\cmidrule[0.5pt]{1-6}

\texttt{Vit-VQGAN}  & ViT-B & 100 & - & 65.1  & - \\
\texttt{Vit-VQGAN}  & ViT-L & 100 & - & 73.2   & - \\

\cmidrule[1pt]{1-6}
\end{tabular}
\label{tbl:ssl_benchmark_gan_based}
\end{table}

\begin{table}[htbp!]
\centering
\caption{ImageNet-1k linear probing and fine-tuning benchmarks for \textbf{MIM}-based generative SSL frameworks. ``SSL epochs'' denotes the number of epochs for the SSL training.}
\scriptsize
\begin{tabular}{cccccc}
\cmidrule[1pt]{1-6}
SSL & Backbone & SSL & Fine tuning & Linear probing & Additional \\
framework & network & epochs & accuracy & accuracy & notes\\
\cmidrule[0.5pt]{1-6}

\texttt{iGPT} & GPT-L & $\sim$ 100 & 72.6 & 65.2 & - \\
\texttt{iGPT} & GPT-XL & $\sim$ 100 & - & 68.7 & - \\
\texttt{iGPT} & GPT-XL & $\sim$ 100 & - & 72.0 & Concatenation of five layers\\
\cmidrule[0.5pt]{1-6}

\texttt{BEiT} & ViT-B & 800 & 83.2 & 56.7 & - \\
\texttt{BEiT} & ViT-L & 300 & 85.2 & 73.5 & -\\
\texttt{BEiT} & ViT-H & 300 & 85.1 & - & -\\
\end{tabular}
\label{tbl:ssl_benchmark_mim_based_fine_tun}
\end{table}

\clearpage
\section{COCO benchmarks}

\begin{table}[htbp!]
\centering
\caption{Downstream transferability results on COCO dataset.}
\scriptsize
\begin{tabular}{ccccc}
\cmidrule[1pt]{1-5}
SSL & Backbone & $\text{AP}^{b}$ & $\text{AP}^{m}$ & Additional \\
framework & network & performance & performance & notes\\
\cmidrule[0.5pt]{1-5}

\texttt{Swav} & ResNet-50 & 41.6 & 37.8 & - \\
\texttt{Coke} & ResNet-50 & 40.9 & 37.2 & - \\
\texttt{Self-Classifier} & ResNet-50 & 41.5 & 36.1 & -\\

\cmidrule[0.5pt]{1-5}

\texttt{InstDist} & ResNet-50 & 37.4 & 34.1 & - \\
\texttt{MoCo} & ResNet-50 & 40.9 & 35.5 & - \\
\texttt{PIRL} & ResNet-50 & 38.5 & 34.0 & - \\
\end{tabular}
\label{tbl:ssl_benchmark_COCO}
\end{table}

\clearpage

\bibliography{main}

\begin{thebibliography}{233}
\providecommand{\natexlab}[1]{#1}
\providecommand{\url}[1]{\texttt{#1}}
\expandafter\ifx\csname urlstyle\endcsname\relax
  \providecommand{\doi}[1]{doi: #1}\else
  \providecommand{\doi}{doi: \begingroup \urlstyle{rm}\Url}\fi

\bibitem[Albelwi(2022)]{albelwi2022_survey}
Saleh Albelwi.
\newblock Survey on self-supervised learning: auxiliary pretext tasks and
  contrastive learning methods in imaging.
\newblock \emph{Entropy}, 24\penalty0 (4):\penalty0 551, 2022.

\bibitem[Amrani \& Bronstein(2021)Amrani and Bronstein]{self_classifier}
Elad Amrani and Alex Bronstein.
\newblock Self-supervised classification network.
\newblock \emph{arXiv preprint arXiv:2103.10994}, 2021.

\bibitem[Asano et~al.(2019)Asano, Rupprecht, and Vedaldi]{sela}
Yuki~Markus Asano, Christian Rupprecht, and Andrea Vedaldi.
\newblock Self-labelling via simultaneous clustering and representation
  learning.
\newblock \emph{arXiv preprint arXiv:1911.05371}, 2019.

\bibitem[Assran et~al.(2023)Assran, Balestriero, Duval, Bordes, Misra,
  Bojanowski, Vincent, Rabbat, and Ballas]{exp4}
Mido Assran, Randall Balestriero, Quentin Duval, Florian Bordes, Ishan Misra,
  Piotr Bojanowski, Pascal Vincent, Michael Rabbat, and Nicolas Ballas.
\newblock The hidden uniform cluster prior in self-supervised learning.
\newblock In \emph{The Eleventh International Conference on Learning
  Representations}, 2023.
\newblock URL \url{https://openreview.net/forum?id=04K3PMtMckp}.

\bibitem[Bachman(2019)]{microsoft_ssl}
Philip Bachman.
\newblock Going meta: learning algorithms and the self-supervised machine.
\newblock In \emph{Microsoft AI Podcast}, 2019.
\newblock URL \url{https://www.youtube.com/watch?v=CSjWb3gcZJ4}.

\bibitem[Bachman et~al.(2019)Bachman, Hjelm, and Buchwalter]{amdim}
Philip Bachman, R~Devon Hjelm, and William Buchwalter.
\newblock Learning representations by maximizing mutual information across
  views.
\newblock \emph{Advances in Neural Information Processing Systems}, 32, 2019.

\bibitem[Baevski et~al.(2022)Baevski, Hsu, Xu, Babu, Gu, and Auli]{data2vec}
Alexei Baevski, Wei-Ning Hsu, Qiantong Xu, Arun Babu, Jiatao Gu, and Michael
  Auli.
\newblock Data2vec: A general framework for self-supervised learning in speech,
  vision and language.
\newblock In \emph{International Conference on Machine Learning}, pp.\
  1298--1312. PMLR, 2022.

\bibitem[Bao et~al.(2021)Bao, Dong, Piao, and Wei]{beit}
Hangbo Bao, Li~Dong, Songhao Piao, and Furu Wei.
\newblock {BEiT}: Bert pre-training of image transformers.
\newblock \emph{arXiv preprint arXiv:2106.08254}, 2021.

\bibitem[Bardes et~al.(2021)Bardes, Ponce, and LeCun]{vicreg}
Adrien Bardes, Jean Ponce, and Yann LeCun.
\newblock {VICReg}: Variance-invariance-covariance regularization for
  self-supervised learning.
\newblock \emph{arXiv preprint arXiv:2105.04906}, 2021.

\bibitem[Bardes et~al.(2022)Bardes, Ponce, and LeCun]{vicregl}
Adrien Bardes, Jean Ponce, and Yann LeCun.
\newblock {VICRegL}: Self-supervised learning of local visual features.
\newblock \emph{arXiv preprint arXiv:2210.01571}, 2022.

\bibitem[Baykal \& Unal(2020)Baykal and Unal]{deshufflegans1}
Gulcin Baykal and Gozde Unal.
\newblock Deshufflegan: A self-supervised gan to improve structure learning.
\newblock In \emph{2020 IEEE International Conference on Image Processing
  (ICIP)}, pp.\  708--712. IEEE, 2020.

\bibitem[Baykal et~al.(2022)Baykal, Ozcelik, and Unal]{deshufflegans2}
Gulcin Baykal, Furkan Ozcelik, and Gozde Unal.
\newblock Exploring deshufflegans in self-supervised generative adversarial
  networks.
\newblock \emph{Pattern Recognition}, 122:\penalty0 108244, 2022.

\bibitem[Becker \& Hinton(1992)Becker and Hinton]{ssl_early}
Suzanna Becker and Geoffrey~E Hinton.
\newblock Self-organizing neural network that discovers surfaces in random-dot
  stereograms.
\newblock \emph{Nature}, 355\penalty0 (6356):\penalty0 161--163, 1992.

\bibitem[Bengio et~al.(2013)Bengio, Courville, and
  Vincent]{representation_learning}
Yoshua Bengio, Aaron Courville, and Pascal Vincent.
\newblock Representation learning: A review and new perspectives.
\newblock \emph{IEEE transactions on pattern analysis and machine
  intelligence}, 35\penalty0 (8):\penalty0 1798--1828, 2013.

\bibitem[Bertalmio et~al.(2000)Bertalmio, Sapiro, Caselles, and
  Ballester]{bertalmio2000image}
Marcelo Bertalmio, Guillermo Sapiro, Vincent Caselles, and Coloma Ballester.
\newblock Image inpainting.
\newblock In \emph{Proceedings of the 27th annual conference on Computer
  graphics and interactive techniques}, pp.\  417--424, 2000.

\bibitem[Bertinetto et~al.(2016)Bertinetto, Valmadre, Henriques, Vedaldi, and
  Torr]{siamese_use1}
Luca Bertinetto, Jack Valmadre, Joao~F Henriques, Andrea Vedaldi, and Philip~HS
  Torr.
\newblock Fully-convolutional siamese networks for object tracking.
\newblock In \emph{European Conference on Computer Vision}, pp.\  850--865.
  Springer, 2016.

\bibitem[Bishop(2006)]{Bishop06a}
Christopher~M. Bishop.
\newblock \emph{{Pattern Recognition and Machine Learning}}.
\newblock Springer, 2006.

\bibitem[Brock et~al.(2018)Brock, Donahue, and Simonyan]{biggan}
Andrew Brock, Jeff Donahue, and Karen Simonyan.
\newblock Large scale {GAN} training for high fidelity natural image synthesis.
\newblock \emph{arXiv preprint arXiv:1809.11096}, 2018.

\bibitem[Bromley et~al.(1993)Bromley, Guyon, LeCun, S{\"a}ckinger, and
  Shah]{siamese}
Jane Bromley, Isabelle Guyon, Yann LeCun, Eduard S{\"a}ckinger, and Roopak
  Shah.
\newblock Signature verification using a ``siamese'' time delay neural network.
\newblock \emph{Advances in Neural Information Processing Systems}, 6, 1993.

\bibitem[Brown et~al.(2020)Brown, Mann, Ryder, Subbiah, Kaplan, Dhariwal,
  Neelakantan, Shyam, Sastry, Askell, et~al.]{gpt3}
Tom Brown, Benjamin Mann, Nick Ryder, Melanie Subbiah, Jared~D Kaplan, Prafulla
  Dhariwal, Arvind Neelakantan, Pranav Shyam, Girish Sastry, Amanda Askell,
  et~al.
\newblock Language models are few-shot learners.
\newblock \emph{Advances in Neural Information Processing Systems},
  33:\penalty0 1877--1901, 2020.

\bibitem[Cao et~al.(2020)Cao, Xie, Liu, Lin, Zhang, and Hu]{pic}
Yue Cao, Zhenda Xie, Bin Liu, Yutong Lin, Zheng Zhang, and Han Hu.
\newblock Parametric instance classification for unsupervised visual feature
  learning.
\newblock \emph{Advances in Neural Information Processing Systems},
  33:\penalty0 15614--15624, 2020.

\bibitem[Carlucci et~al.(2019)Carlucci, D'Innocente, Bucci, Caputo, and
  Tommasi]{puzzle_generalization}
Fabio~M Carlucci, Antonio D'Innocente, Silvia Bucci, Barbara Caputo, and
  Tatiana Tommasi.
\newblock Domain generalization by solving jigsaw puzzles.
\newblock In \emph{Proceedings of the IEEE/CVF Conference on Computer Vision
  and Pattern Recognition}, pp.\  2229--2238, 2019.

\bibitem[Caron et~al.(2018)Caron, Bojanowski, Joulin, and Douze]{deepc}
Mathilde Caron, Piotr Bojanowski, Armand Joulin, and Matthijs Douze.
\newblock Deep clustering for unsupervised learning of visual features.
\newblock In \emph{Proceedings of the European Conference on Computer Vision
  (ECCV)}, pp.\  132--149, 2018.

\bibitem[Caron et~al.(2019)Caron, Bojanowski, Mairal, and Joulin]{deeperC}
Mathilde Caron, Piotr Bojanowski, Julien Mairal, and Armand Joulin.
\newblock Unsupervised pre-training of image features on non-curated data.
\newblock In \emph{Proceedings of the IEEE/CVF International Conference on
  Computer Vision}, pp.\  2959--2968, 2019.

\bibitem[Caron et~al.(2020)Caron, Misra, Mairal, Goyal, Bojanowski, and
  Joulin]{swav}
Mathilde Caron, Ishan Misra, Julien Mairal, Priya Goyal, Piotr Bojanowski, and
  Armand Joulin.
\newblock Unsupervised learning of visual features by contrasting cluster
  assignments.
\newblock \emph{Advances in Neural Information Processing Systems},
  33:\penalty0 9912--9924, 2020.

\bibitem[Caron et~al.(2021)Caron, Touvron, Misra, J{\'e}gou, Mairal,
  Bojanowski, and Joulin]{dino}
Mathilde Caron, Hugo Touvron, Ishan Misra, Herv{\'e} J{\'e}gou, Julien Mairal,
  Piotr Bojanowski, and Armand Joulin.
\newblock Emerging properties in self-supervised vision transformers.
\newblock In \emph{Proceedings of the IEEE/CVF International Conference on
  Computer Vision}, pp.\  9650--9660, 2021.

\bibitem[Celebi \& Aydin(2016)Celebi and Aydin]{unsupervsed_survey}
M~Emre Celebi and Kemal Aydin.
\newblock \emph{Unsupervised learning algorithms}.
\newblock Springer, 2016.

\bibitem[Chakraborty et~al.(2020)Chakraborty, Gosthipaty, and Paul]{g_simclr}
Souradip Chakraborty, Aritra~Roy Gosthipaty, and Sayak Paul.
\newblock {G-SimCLR}: Self-supervised contrastive learning with guided
  projection via pseudo labelling.
\newblock In \emph{2020 International Conference on Data Mining Workshops
  (ICDMW)}, pp.\  912--916. IEEE, 2020.

\bibitem[Charpiat et~al.(2008)Charpiat, Hofmann, and
  Sch{\"o}lkopf]{image_colorization_no_dnn_2}
Guillaume Charpiat, Matthias Hofmann, and Bernhard Sch{\"o}lkopf.
\newblock Automatic image colorization via multimodal predictions.
\newblock In \emph{European Conference on Computer Vision}, pp.\  126--139.
  Springer, 2008.

\bibitem[Chen et~al.(2017)Chen, Fisch, Weston, and Bordes]{wiki_data}
Danqi Chen, Adam Fisch, Jason Weston, and Antoine Bordes.
\newblock Reading {W}ikipedia to answer open-domain questions.
\newblock \emph{arXiv preprint arXiv:1704.00051}, 2017.

\bibitem[Chen et~al.(2020{\natexlab{a}})Chen, Radford, Child, Wu, Jun, Luan,
  and Sutskever]{igpt}
Mark Chen, Alec Radford, Rewon Child, Jeffrey Wu, Heewoo Jun, David Luan, and
  Ilya Sutskever.
\newblock Generative pretraining from pixels.
\newblock In \emph{International conference on machine learning}, pp.\
  1691--1703. PMLR, 2020{\natexlab{a}}.

\bibitem[Chen et~al.(2023{\natexlab{a}})Chen, Cai, Cai, Yu, Qian, and
  Xiang]{chen2023colo}
Qingzhong Chen, Shilun Cai, Crystal Cai, Zefang Yu, Dahong Qian, and Suncheng
  Xiang.
\newblock Colo-scrl: Self-supervised contrastive representation learning for
  colonoscopic video retrieval.
\newblock \emph{arXiv preprint arXiv:2303.15671}, 2023{\natexlab{a}}.

\bibitem[Chen(2020)]{google_ssl}
Ting Chen.
\newblock Advancing self-supervised and semi-supervised learning with {SimCLR}.
\newblock In \emph{Google AI Blog}, 2020.
\newblock URL
  \url{https://ai.googleblog.com/2020/04/advancing-self-supervised-and-semi.html}.

\bibitem[Chen et~al.(2019)Chen, Zhai, Ritter, Lucic, and Houlsby]{ss_gan}
Ting Chen, Xiaohua Zhai, Marvin Ritter, Mario Lucic, and Neil Houlsby.
\newblock Self-supervised {GAN}s via auxiliary rotation loss.
\newblock In \emph{Proceedings of the IEEE/CVF conference on computer vision
  and pattern recognition}, pp.\  12154--12163, 2019.

\bibitem[Chen et~al.(2020{\natexlab{b}})Chen, Kornblith, Norouzi, and
  Hinton]{simclr}
Ting Chen, Simon Kornblith, Mohammad Norouzi, and Geoffrey Hinton.
\newblock A simple framework for contrastive learning of visual
  representations.
\newblock In \emph{International Conference on Machine Learning}, pp.\
  1597--1607. PMLR, 2020{\natexlab{b}}.

\bibitem[Chen et~al.(2020{\natexlab{c}})Chen, Kornblith, Swersky, Norouzi, and
  Hinton]{simclr_2}
Ting Chen, Simon Kornblith, Kevin Swersky, Mohammad Norouzi, and Geoffrey~E
  Hinton.
\newblock Big self-supervised models are strong semi-supervised learners.
\newblock \emph{Advances in Neural Information Processing Systems},
  33:\penalty0 22243--22255, 2020{\natexlab{c}}.

\bibitem[Chen et~al.(2022{\natexlab{a}})Chen, Ding, Wang, Xin, Mo, Wang, Han,
  Luo, Zeng, and Wang]{cae}
Xiaokang Chen, Mingyu Ding, Xiaodi Wang, Ying Xin, Shentong Mo, Yunhao Wang,
  Shumin Han, Ping Luo, Gang Zeng, and Jingdong Wang.
\newblock Context autoencoder for self-supervised representation learning.
\newblock \emph{arXiv preprint arXiv:2202.03026}, 2022{\natexlab{a}}.

\bibitem[Chen \& He(2021)Chen and He]{simsiam}
Xinlei Chen and Kaiming He.
\newblock Exploring simple siamese representation learning.
\newblock In \emph{Proceedings of the IEEE/CVF Conference on Computer Vision
  and Pattern Recognition}, pp.\  15750--15758, 2021.

\bibitem[Chen et~al.(2020{\natexlab{d}})Chen, Fan, Girshick, and He]{moco_v2}
Xinlei Chen, Haoqi Fan, Ross Girshick, and Kaiming He.
\newblock Improved baselines with momentum contrastive learning.
\newblock \emph{arXiv preprint arXiv:2003.04297}, 2020{\natexlab{d}}.

\bibitem[Chen et~al.(2021)Chen, Xie, and He]{moco_v3}
Xinlei Chen, Saining Xie, and Kaiming He.
\newblock An empirical study of training self-supervised vision transformers.
\newblock In \emph{Proceedings of the IEEE/CVF International Conference on
  Computer Vision}, pp.\  9640--9649, 2021.

\bibitem[Chen et~al.(2022{\natexlab{b}})Chen, Liu, Jiang, Zhang, Dai, Xiong,
  and Tian]{sdae}
Yabo Chen, Yuchen Liu, Dongsheng Jiang, Xiaopeng Zhang, Wenrui Dai, Hongkai
  Xiong, and Qi~Tian.
\newblock Sdae: Self-distillated masked autoencoder.
\newblock In \emph{Computer Vision--ECCV 2022: 17th European Conference, Tel
  Aviv, Israel, October 23--27, 2022, Proceedings, Part XXX}, pp.\  108--124.
  Springer, 2022{\natexlab{b}}.

\bibitem[Chen et~al.(2023{\natexlab{b}})Chen, Bardes, Li, and
  LeCun]{chenintra_vicreg}
Yubei Chen, Adrien Bardes, Zengyi Li, and Yann LeCun.
\newblock Intra-instance {VICReg}: Bag of self-supervised image patch embedding
  explains the performance.
\newblock 2023{\natexlab{b}}.
\newblock URL \url{https://openreview.net/forum?id=J923QzIz8Sh}.

\bibitem[Chen et~al.(2023{\natexlab{c}})Chen, Li, Wang, and
  Yang]{chen2023liftedcl}
Ziwei Chen, Qiang Li, Xiaofeng Wang, and Wankou Yang.
\newblock Lifted{CL}: Lifting contrastive learning for human-centric
  perception.
\newblock In \emph{The Eleventh International Conference on Learning
  Representations}, 2023{\natexlab{c}}.

\bibitem[Cheng et~al.(2015)Cheng, Yang, and Sheng]{improve_coloring_1}
Zezhou Cheng, Qingxiong Yang, and Bin Sheng.
\newblock Deep colorization.
\newblock In \emph{Proceedings of the IEEE international conference on computer
  vision}, pp.\  415--423, 2015.

\bibitem[Chicco(2021)]{siamese_use2}
Davide Chicco.
\newblock Siamese neural networks: An overview.
\newblock \emph{Artificial Neural Networks}, pp.\  73--94, 2021.

\bibitem[Chopra et~al.(2005)Chopra, Hadsell, and LeCun]{siamese_use3}
Sumit Chopra, Raia Hadsell, and Yann LeCun.
\newblock Learning a similarity metric discriminatively, with application to
  face verification.
\newblock In \emph{2005 IEEE Computer Society Conference on Computer Vision and
  Pattern Recognition (CVPR'05)}, volume~1, pp.\  539--546. IEEE, 2005.

\bibitem[Chuang et~al.(2020)Chuang, Robinson, Lin, Torralba, and
  Jegelka]{negative_samples_1}
Ching-Yao Chuang, Joshua Robinson, Yen-Chen Lin, Antonio Torralba, and Stefanie
  Jegelka.
\newblock Debiased contrastive learning.
\newblock \emph{Advances in Neural Information Processing Systems},
  33:\penalty0 8765--8775, 2020.

\bibitem[Clevert et~al.(2015)Clevert, Unterthiner, and
  Hochreiter]{ELU_activation}
Djork-Arn{\'e} Clevert, Thomas Unterthiner, and Sepp Hochreiter.
\newblock {Fast And accurate deep network learning by exponential linear units
  (Elus)}.
\newblock \emph{arXiv preprint arXiv:1511.07289}, 2015.

\bibitem[Coates \& Ng(2012)Coates and Ng]{clustering_early_1}
Adam Coates and Andrew~Y Ng.
\newblock Learning feature representations with k-means.
\newblock In \emph{Neural Networks: Tricks of the Trade}, pp.\  561--580.
  Springer, 2012.

\bibitem[Coates et~al.(2011)Coates, Ng, and Lee]{clustering_early_0}
Adam Coates, Andrew Ng, and Honglak Lee.
\newblock An analysis of single-layer networks in unsupervised feature
  learning.
\newblock In \emph{Proceedings of the fourteenth international conference on
  artificial intelligence and statistics}, pp.\  215--223. JMLR Workshop and
  Conference Proceedings, 2011.

\bibitem[Contributors(2022)]{easy_cv}
EasyCV Contributors.
\newblock {EasyCV}.
\newblock \url{https://github.com/alibaba/EasyCV}, 2022.

\bibitem[Contributors(2021)]{mmselfsup}
MMSelfSup Contributors.
\newblock {MMSelfSup}: {OpenMMLab} self-supervised learning toolbox and
  benchmark.
\newblock \url{https://github.com/open-mmlab/mmselfsup}, 2021.

\bibitem[Cuturi(2013)]{sinkhorn_algo}
Marco Cuturi.
\newblock Sinkhorn distances: Lightspeed computation of optimal transport.
\newblock \emph{Advances in Neural Information Processing Systems}, 26, 2013.

\bibitem[da~Costa et~al.(2022)da~Costa, Fini, Nabi, Sebe, and
  Ricci]{solo_learn}
Victor Guilherme~Turrisi da~Costa, Enrico Fini, Moin Nabi, Nicu Sebe, and Elisa
  Ricci.
\newblock solo-learn: A library of self-supervised methods for visual
  representation learning.
\newblock \emph{Journal of Machine Learning Research}, 23\penalty0
  (56):\penalty0 1--6, 2022.
\newblock URL \url{http://jmlr.org/papers/v23/21-1155.html}.

\bibitem[Devlin et~al.(2018)Devlin, Chang, Lee, and Toutanova]{bert}
Jacob Devlin, Ming-Wei Chang, Kenton Lee, and Kristina Toutanova.
\newblock Bert: Pre-training of deep bidirectional transformers for language
  understanding.
\newblock \emph{arXiv preprint arXiv:1810.04805}, 2018.

\bibitem[Doersch et~al.(2015)Doersch, Gupta, and Efros]{ssl_context_pred}
Carl Doersch, Abhinav Gupta, and Alexei~A Efros.
\newblock Unsupervised visual representation learning by context prediction.
\newblock In \emph{Proceedings of the IEEE international conference on computer
  vision}, pp.\  1422--1430, 2015.

\bibitem[Donahue \& Simonyan(2019)Donahue and Simonyan]{bigbigan}
Jeff Donahue and Karen Simonyan.
\newblock Large scale adversarial representation learning.
\newblock \emph{Advances in neural information processing systems}, 32, 2019.

\bibitem[Donahue et~al.(2016)Donahue, Kr{\"a}henb{\"u}hl, and Darrell]{bigan}
Jeff Donahue, Philipp Kr{\"a}henb{\"u}hl, and Trevor Darrell.
\newblock Adversarial feature learning.
\newblock \emph{arXiv preprint arXiv:1605.09782}, 2016.

\bibitem[Dong et~al.(2021)Dong, Bao, Zhang, Chen, Zhang, Yuan, Chen, Wen, and
  Yu]{peco}
Xiaoyi Dong, Jianmin Bao, Ting Zhang, Dongdong Chen, Weiming Zhang, Lu~Yuan,
  Dong Chen, Fang Wen, and Nenghai Yu.
\newblock Peco: Perceptual codebook for bert pre-training of vision
  transformers.
\newblock \emph{arXiv preprint arXiv:2111.12710}, 2021.

\bibitem[Dosovitskiy et~al.(2014)Dosovitskiy, Springenberg, Riedmiller, and
  Brox]{ssl_discriminative_1}
Alexey Dosovitskiy, Jost~Tobias Springenberg, Martin Riedmiller, and Thomas
  Brox.
\newblock Discriminative unsupervised feature learning with convolutional
  neural networks.
\newblock \emph{Advances in Neural Information Processing Systems}, 27, 2014.

\bibitem[Dosovitskiy et~al.(2020)Dosovitskiy, Beyer, Kolesnikov, Weissenborn,
  Zhai, Unterthiner, Dehghani, Minderer, Heigold, Gelly, et~al.]{vit}
Alexey Dosovitskiy, Lucas Beyer, Alexander Kolesnikov, Dirk Weissenborn,
  Xiaohua Zhai, Thomas Unterthiner, Mostafa Dehghani, Matthias Minderer, Georg
  Heigold, Sylvain Gelly, et~al.
\newblock An image is worth 16x16 words: Transformers for image recognition at
  scale.
\newblock \emph{arXiv preprint arXiv:2010.11929}, 2020.

\bibitem[Dumoulin et~al.(2016)Dumoulin, Belghazi, Poole, Mastropietro, Lamb,
  Arjovsky, and Courville]{ali}
Vincent Dumoulin, Ishmael Belghazi, Ben Poole, Olivier Mastropietro, Alex Lamb,
  Martin Arjovsky, and Aaron Courville.
\newblock Adversarially learned inference.
\newblock \emph{arXiv preprint arXiv:1606.00704}, 2016.

\bibitem[Dwibedi et~al.(2021)Dwibedi, Aytar, Tompson, Sermanet, and
  Zisserman]{nnclr}
Debidatta Dwibedi, Yusuf Aytar, Jonathan Tompson, Pierre Sermanet, and Andrew
  Zisserman.
\newblock With a little help from my friends: Nearest-neighbor contrastive
  learning of visual representations.
\newblock In \emph{Proceedings of the IEEE/CVF International Conference on
  Computer Vision}, pp.\  9588--9597, 2021.

\bibitem[Efros(2019)]{gelato_ssl}
Alyosha Efros.
\newblock The gelato bet.
\newblock 2019.
\newblock URL \url{https://people.eecs.berkeley.edu/~efros/gelato_bet.html}.

\bibitem[Ermolov et~al.(2021)Ermolov, Siarohin, Sangineto, and Sebe]{wmse}
Aleksandr Ermolov, Aliaksandr Siarohin, Enver Sangineto, and Nicu Sebe.
\newblock Whitening for self-supervised representation learning.
\newblock In \emph{International Conference on Machine Learning}, pp.\
  3015--3024. PMLR, 2021.

\bibitem[Esser et~al.(2021)Esser, Rombach, and Ommer]{vqgan}
Patrick Esser, Robin Rombach, and Bjorn Ommer.
\newblock Taming transformers for high-resolution image synthesis.
\newblock In \emph{Proceedings of the IEEE/CVF conference on computer vision
  and pattern recognition}, pp.\  12873--12883, 2021.

\bibitem[Everingham et~al.(2007)Everingham, Van~Gool, Williams, Winn, and
  Zisserman]{pascal_voc}
M.~Everingham, L.~Van~Gool, C.~K.~I. Williams, J.~Winn, and A.~Zisserman.
\newblock The {PASCAL} {V}isual {O}bject {C}lasses {C}hallenge 2007 {(VOC2007)}
  {R}esults.
\newblock
  http://www.pascal-network.org/challenges/VOC/voc2007/workshop/index.html,
  2007.

\bibitem[Fang et~al.(2023)Fang, Dong, Bao, Wang, and Wei]{cim}
Yuxin Fang, Li~Dong, Hangbo Bao, Xinggang Wang, and Furu Wei.
\newblock Corrupted image modeling for self-supervised visual pre-training.
\newblock In \emph{The Eleventh International Conference on Learning
  Representations}, 2023.
\newblock URL \url{https://openreview.net/forum?id=09hVcSDkea}.

\bibitem[Fang et~al.(2021)Fang, Wang, Wang, Zhang, Yang, and Liu]{seed}
Zhiyuan Fang, Jianfeng Wang, Lijuan Wang, Lei Zhang, Yezhou Yang, and Zicheng
  Liu.
\newblock Seed: Self-supervised distillation for visual representation.
\newblock \emph{arXiv preprint arXiv:2101.04731}, 2021.

\bibitem[Fetterman \& Albrecht(2020)Fetterman and Albrecht]{byol_collapse_h1}
Abe Fetterman and Josh Albrecht.
\newblock Understanding self-supervised and contrastive learning with
  ``bootstrap your own latent'' ({BYOL}).
\newblock In \emph{Generally Intelligent AI Blog}, 2020.
\newblock URL
  \url{https://generallyintelligent.ai/blog/2020-08-24-understanding-self-supervised-contrastive-learning/}.

\bibitem[Gao et~al.(2022)Gao, Ma, Li, Dai, and Qiao]{mcmae}
Peng Gao, Teli Ma, Hongsheng Li, Jifeng Dai, and Yu~Qiao.
\newblock {ConvMAE}: Masked convolution meets masked autoencoders.
\newblock \emph{arXiv preprint arXiv:2205.03892}, 2022.

\bibitem[Gao et~al.(2023)Gao, Zhang, Li, Li, and Qiao]{mrmae}
Peng Gao, Renrui Zhang, Hongyang Li, Hongsheng Li, and Yu~Qiao.
\newblock Mimic before reconstruct: Enhance masked autoencoders with feature
  mimicking, 2023.
\newblock URL \url{https://openreview.net/forum?id=UoBJm4V21md}.

\bibitem[Gao et~al.(2021)Gao, Zhuang, Li, Cheng, Guo, Huang, Ji, and
  Sun]{disco}
Yuting Gao, Jia-Xin Zhuang, Ke~Li, Hao Cheng, Xiaowei Guo, Feiyue Huang,
  Rongrong Ji, and Xing Sun.
\newblock Disco: Remedy self-supervised learning on lightweight models with
  distilled contrastive learning.
\newblock \emph{arXiv preprint arXiv:2104.09124}, 2021.

\bibitem[Garrido et~al.(2022{\natexlab{a}})Garrido, Balestriero, Najman, and
  Lecun]{efficient_ssl_evaluation}
Quentin Garrido, Randall Balestriero, Laurent Najman, and Yann Lecun.
\newblock Rankme: Assessing the downstream performance of pretrained
  self-supervised representations by their rank.
\newblock \emph{arXiv preprint arXiv:2210.02885}, 2022{\natexlab{a}}.

\bibitem[Garrido et~al.(2022{\natexlab{b}})Garrido, Chen, Bardes, Najman, and
  Lecun]{garrido2022duality}
Quentin Garrido, Yubei Chen, Adrien Bardes, Laurent Najman, and Yann Lecun.
\newblock On the duality between contrastive and non-contrastive
  self-supervised learning.
\newblock \emph{arXiv preprint arXiv:2206.02574}, 2022{\natexlab{b}}.

\bibitem[Gidaris et~al.(2018)Gidaris, Singh, and Komodakis]{rotation_ssl}
Spyros Gidaris, Praveer Singh, and Nikos Komodakis.
\newblock Unsupervised representation learning by predicting image rotations.
\newblock \emph{arXiv preprint arXiv:1803.07728}, 2018.

\bibitem[Gidaris et~al.(2021)Gidaris, Bursuc, Puy, Komodakis, Cord, and
  P{\'e}rez]{obow}
Spyros Gidaris, Andrei Bursuc, Gilles Puy, Nikos Komodakis, Matthieu Cord, and
  Patrick P{\'e}rez.
\newblock Online bag-of-visual-words generation for unsupervised representation
  learning.
\newblock \emph{arXiv preprint arXiv:2012.11552}, 2021.

\bibitem[Gogna \& Majumdar(2016)Gogna and Majumdar]{ssl_autoencoder}
Anupriya Gogna and Angshul Majumdar.
\newblock Semi supervised autoencoder.
\newblock In \emph{International Conference on Neural Information Processing},
  pp.\  82--89. Springer, 2016.

\bibitem[Goodfellow et~al.(2020)Goodfellow, Pouget-Abadie, Mirza, Xu,
  Warde-Farley, Ozair, Courville, and Bengio]{gan}
Ian Goodfellow, Jean Pouget-Abadie, Mehdi Mirza, Bing Xu, David Warde-Farley,
  Sherjil Ozair, Aaron Courville, and Yoshua Bengio.
\newblock Generative adversarial networks.
\newblock \emph{Communications of the ACM}, 63\penalty0 (11):\penalty0
  139--144, 2020.

\bibitem[Goyal et~al.(2021{\natexlab{a}})Goyal, Caron, Lefaudeux, Xu, Wang,
  Pai, Singh, Liptchinsky, Misra, Joulin, et~al.]{ssl_wild}
Priya Goyal, Mathilde Caron, Benjamin Lefaudeux, Min Xu, Pengchao Wang, Vivek
  Pai, Mannat Singh, Vitaliy Liptchinsky, Ishan Misra, Armand Joulin, et~al.
\newblock Self-supervised pretraining of visual features in the wild.
\newblock \emph{arXiv preprint arXiv:2103.01988}, 2021{\natexlab{a}}.

\bibitem[Goyal et~al.(2021{\natexlab{b}})Goyal, Duval, Reizenstein, Leavitt,
  Xu, Lefaudeux, Singh, Reis, Caron, Bojanowski, Joulin, and Misra]{vissl}
Priya Goyal, Quentin Duval, Jeremy Reizenstein, Matthew Leavitt, Min Xu,
  Benjamin Lefaudeux, Mannat Singh, Vinicius Reis, Mathilde Caron, Piotr
  Bojanowski, Armand Joulin, and Ishan Misra.
\newblock {VISSL}.
\newblock \url{https://github.com/facebookresearch/vissl}, 2021{\natexlab{b}}.

\bibitem[Grill et~al.(2020)Grill, Strub, Altch{\'e}, Tallec, Richemond,
  Buchatskaya, Doersch, Avila~Pires, Guo, Gheshlaghi~Azar, et~al.]{byol}
Jean-Bastien Grill, Florian Strub, Florent Altch{\'e}, Corentin Tallec, Pierre
  Richemond, Elena Buchatskaya, Carl Doersch, Bernardo Avila~Pires, Zhaohan
  Guo, Mohammad Gheshlaghi~Azar, et~al.
\newblock Bootstrap your own latent -- a new approach to self-supervised
  learning.
\newblock \emph{Advances in Neural Information Processing Systems},
  33:\penalty0 21271--21284, 2020.

\bibitem[Gutmann \& Hyv{\"a}rinen(2010)Gutmann and Hyv{\"a}rinen]{nce}
Michael Gutmann and Aapo Hyv{\"a}rinen.
\newblock Noise-contrastive estimation: A new estimation principle for
  unnormalized statistical models.
\newblock In \emph{Proceedings of the thirteenth international conference on
  artificial intelligence and statistics}, pp.\  297--304. JMLR Workshop and
  Conference Proceedings, 2010.

\bibitem[Hadsell et~al.(2006)Hadsell, Chopra, and LeCun]{early_contrastive_1}
Raia Hadsell, Sumit Chopra, and Yann LeCun.
\newblock Dimensionality reduction by learning an invariant mapping.
\newblock In \emph{2006 IEEE Computer Society Conference on Computer Vision and
  Pattern Recognition (CVPR'06)}, volume~2, pp.\  1735--1742. IEEE, 2006.

\bibitem[Hamilton et~al.(2017)Hamilton, Ying, and Leskovec]{reddit_data}
Will Hamilton, Zhitao Ying, and Jure Leskovec.
\newblock Inductive representation learning on large graphs.
\newblock \emph{Advances in Neural Information Processing Systems}, 30, 2017.

\bibitem[He et~al.(2016)He, Zhang, Ren, and Sun]{resnet}
Kaiming He, Xiangyu Zhang, Shaoqing Ren, and Jian Sun.
\newblock {Deep residual learning for image recognition}.
\newblock In \emph{Proceedings of the IEEE/CVF Conference on Computer Vision
  and Pattern Recognition}, 2016.

\bibitem[He et~al.(2020)He, Fan, Wu, Xie, and Girshick]{mocov1}
Kaiming He, Haoqi Fan, Yuxin Wu, Saining Xie, and Ross Girshick.
\newblock Momentum contrast for unsupervised visual representation learning.
\newblock In \emph{Proceedings of the IEEE/CVF Conference on Computer Vision
  and Pattern Recognition}, pp.\  9729--9738, 2020.

\bibitem[He et~al.(2022)He, Chen, Xie, Li, Doll{\'a}r, and Girshick]{mae}
Kaiming He, Xinlei Chen, Saining Xie, Yanghao Li, Piotr Doll{\'a}r, and Ross
  Girshick.
\newblock Masked autoencoders are scalable vision learners.
\newblock In \emph{Proceedings of the IEEE/CVF Conference on Computer Vision
  and Pattern Recognition}, pp.\  16000--16009, 2022.

\bibitem[Henaff(2020)]{cpc_v2}
Olivier Henaff.
\newblock Data-efficient image recognition with contrastive predictive coding.
\newblock In \emph{International Conference on Machine Learning}, pp.\
  4182--4192. PMLR, 2020.

\bibitem[Hendrycks et~al.(2019)Hendrycks, Mazeika, Kadavath, and
  Song]{ssl_calib}
Dan Hendrycks, Mantas Mazeika, Saurav Kadavath, and Dawn Song.
\newblock Using self-supervised learning can improve model robustness and
  uncertainty.
\newblock \emph{Advances in Neural Information Processing Systems}, 32, 2019.

\bibitem[Heusel et~al.(2017)Heusel, Ramsauer, Unterthiner, Nessler, and
  Hochreiter]{frechlet_dist}
Martin Heusel, Hubert Ramsauer, Thomas Unterthiner, Bernhard Nessler, and Sepp
  Hochreiter.
\newblock {GAN}s trained by a two time-scale update rule converge to a local
  {N}ash equilibrium.
\newblock \emph{Advances in neural information processing systems}, 30, 2017.

\bibitem[Hinton et~al.(2015)Hinton, Vinyals, Dean, et~al.]{distilling_old}
Geoffrey Hinton, Oriol Vinyals, Jeff Dean, et~al.
\newblock Distilling the knowledge in a neural network.
\newblock \emph{arXiv preprint arXiv:1503.02531}, 2\penalty0 (7), 2015.

\bibitem[Hinton et~al.(2006)Hinton, Osindero, and Teh]{hinton2006fast}
Geoffrey~E Hinton, Simon Osindero, and Yee-Whye Teh.
\newblock A fast learning algorithm for deep belief nets.
\newblock \emph{Neural computation}, 18\penalty0 (7):\penalty0 1527--1554,
  2006.

\bibitem[Hjelm et~al.(2018)Hjelm, Fedorov, Lavoie-Marchildon, Grewal, Bachman,
  Trischler, and Bengio]{dim}
R~Devon Hjelm, Alex Fedorov, Samuel Lavoie-Marchildon, Karan Grewal, Phil
  Bachman, Adam Trischler, and Yoshua Bengio.
\newblock Learning deep representations by mutual information estimation and
  maximization.
\newblock \emph{arXiv preprint arXiv:1808.06670}, 2018.

\bibitem[Hou et~al.(2021)Hou, Shen, Cao, and Cheng]{ssgan_la}
Liang Hou, Huawei Shen, Qi~Cao, and Xueqi Cheng.
\newblock Self-supervised {GAN}s with label augmentation.
\newblock \emph{Advances in Neural Information Processing Systems},
  34:\penalty0 13019--13031, 2021.

\bibitem[Hou et~al.(2022)Hou, Sun, Chen, Xie, and Kung]{milan}
Zejiang Hou, Fei Sun, Yen-Kuang Chen, Yuan Xie, and Sun-Yuan Kung.
\newblock Milan: Masked image pretraining on language assisted representation.
\newblock \emph{arXiv preprint arXiv:2208.06049}, 2022.

\bibitem[Howard(2020)]{fastai_ssl}
Jeremy Howard.
\newblock Self-supervised learning and computer vision.
\newblock In \emph{fast.ai Blog}, 2020.
\newblock URL \url{https://www.fast.ai/2020/01/13/self_supervised/}.

\bibitem[Hu et~al.(2021)Hu, Wang, Hu, and Qi]{adco}
Qianjiang Hu, Xiao Wang, Wei Hu, and Guo-Jun Qi.
\newblock Adco: Adversarial contrast for efficient learning of unsupervised
  representations from self-trained negative adversaries.
\newblock In \emph{Proceedings of the IEEE/CVF Conference on Computer Vision
  and Pattern Recognition}, pp.\  1074--1083, 2021.

\bibitem[Hu et~al.(2023)Hu, LIU, Zhou, Wang, and Huang]{exp1}
Tianyang Hu, Zhili LIU, Fengwei Zhou, Wenjia Wang, and Weiran Huang.
\newblock Your contrastive learning is secretly doing stochastic neighbor
  embedding.
\newblock In \emph{The Eleventh International Conference on Learning
  Representations}, 2023.
\newblock URL \url{https://openreview.net/forum?id=XFSCKELP3bp}.

\bibitem[Hua et~al.(2021)Hua, Wang, Xue, Ren, Wang, and Zhao]{ssl_collapse}
Tianyu Hua, Wenxiao Wang, Zihui Xue, Sucheng Ren, Yue Wang, and Hang Zhao.
\newblock On feature decorrelation in self-supervised learning.
\newblock In \emph{Proceedings of the IEEE/CVF International Conference on
  Computer Vision}, pp.\  9598--9608, 2021.

\bibitem[Huang et~al.(2022)Huang, Jin, Lu, Hou, Cheng, Fu, Shen, and
  Feng]{cmae}
Zhicheng Huang, Xiaojie Jin, Chengze Lu, Qibin Hou, Ming-Ming Cheng, Dongmei
  Fu, Xiaohui Shen, and Jiashi Feng.
\newblock Contrastive masked autoencoders are stronger vision learners.
\newblock \emph{arXiv preprint arXiv:2207.13532}, 2022.

\bibitem[Hui(2013)]{ssl_discriminative_3}
Ka~Yu Hui.
\newblock Direct modeling of complex invariances for visual object features.
\newblock In \emph{International Conference on Machine Learning}, pp.\
  352--360. PMLR, 2013.

\bibitem[ICLR(2023)]{beit_review}
ICLR.
\newblock {ICLR 2023 Reviews for: BEiT v2: Masked Image Modeling with
  Vector-Quantized Visual Tokenizers}.
\newblock \url{https://openreview.net/forum?id=VB75Pi89p7}, 2023.
\newblock Accessed: 2023-04-04.

\bibitem[Iizuka et~al.(2016)Iizuka, Simo-Serra, and
  Ishikawa]{improve_coloring_2}
Satoshi Iizuka, Edgar Simo-Serra, and Hiroshi Ishikawa.
\newblock Let there be color! {J}oint end-to-end learning of global and local
  image priors for automatic image colorization with simultaneous
  classification.
\newblock \emph{ACM Transactions on Graphics (ToG)}, 35\penalty0 (4):\penalty0
  1--11, 2016.

\bibitem[Ioffe \& Szegedy(2015)Ioffe and Szegedy]{batch_norm}
Sergey Ioffe and Christian Szegedy.
\newblock {Batch normalization: accelerating deep network training by reducing
  internal covariate shift}.
\newblock \emph{CoRR}, abs/1502.03167, 2015.

\bibitem[Jain et~al.(2022)Jain, Salman, Khaddaj, Wong, Park, and
  Madry]{transfer_learning_brittle}
Saachi Jain, Hadi Salman, Alaa Khaddaj, Eric Wong, Sung~Min Park, and
  Aleksander Madry.
\newblock A data-based perspective on transfer learning.
\newblock \emph{arXiv preprint arXiv:2207.05739}, 2022.

\bibitem[Jing et~al.(2021)Jing, Vincent, LeCun, and
  Tian]{dimensional_collapse_directCLR}
Li~Jing, Pascal Vincent, Yann LeCun, and Yuandong Tian.
\newblock Understanding dimensional collapse in contrastive self-supervised
  learning.
\newblock \emph{arXiv preprint arXiv:2110.09348}, 2021.

\bibitem[Johnson et~al.(2023)Johnson, Hanchi, and Maddison]{exp2}
Daniel~D. Johnson, Ayoub~El Hanchi, and Chris~J. Maddison.
\newblock Contrastive learning can find an optimal basis for approximately
  view-invariant functions.
\newblock In \emph{The Eleventh International Conference on Learning
  Representations}, 2023.
\newblock URL \url{https://openreview.net/forum?id=AjC0KBjiMu}.

\bibitem[Joulin et~al.(2016)Joulin, Maaten, Jabri, and
  Vasilache]{clustering_collapse}
Armand Joulin, Laurens van~der Maaten, Allan Jabri, and Nicolas Vasilache.
\newblock Learning visual features from large weakly supervised data.
\newblock In \emph{European Conference on Computer Vision}, pp.\  67--84.
  Springer, 2016.

\bibitem[Kalantidis et~al.(2020)Kalantidis, Sariyildiz, Pion, Weinzaepfel, and
  Larlus]{mochi}
Yannis Kalantidis, Mert~Bulent Sariyildiz, Noe Pion, Philippe Weinzaepfel, and
  Diane Larlus.
\newblock Hard negative mixing for contrastive learning.
\newblock \emph{Advances in Neural Information Processing Systems},
  33:\penalty0 21798--21809, 2020.

\bibitem[Kalantidis et~al.(2021)Kalantidis, Lassance, Almazan, and
  Larlus]{tldr}
Yannis Kalantidis, Carlos Lassance, Jon Almazan, and Diane Larlus.
\newblock {TLDR}: Twin learning for dimensionality reduction.
\newblock \emph{arXiv preprint arXiv:2110.09455}, 2021.

\bibitem[Kanazawa et~al.(2016)Kanazawa, Jacobs, and Chandraker]{ssl_geometry_1}
Angjoo Kanazawa, David~W Jacobs, and Manmohan Chandraker.
\newblock Warpnet: Weakly supervised matching for single-view reconstruction.
\newblock In \emph{Proceedings of the IEEE/CVF Conference on Computer Vision
  and Pattern Recognition}, pp.\  3253--3261, 2016.

\bibitem[Khan et~al.(2022)Khan, AlBarri, and
  Manzoor]{khan2022contrastive_survey}
Adnan Khan, Sarah AlBarri, and Muhammad~Arslan Manzoor.
\newblock Contrastive self-supervised learning: a survey on different
  architectures.
\newblock In \emph{2022 2nd International Conference on Artificial Intelligence
  (ICAI)}, pp.\  1--6. IEEE, 2022.

\bibitem[Kingma \& Ba(2014)Kingma and Ba]{adaptive_momentum}
Diederik~P. Kingma and Jimmy Ba.
\newblock {Adam: {{A}} method for stochastic optimization}.
\newblock \emph{CoRR}, abs/1412.6980, 2014.

\bibitem[Kingma \& Welling(2013)Kingma and Welling]{vae}
Diederik~P Kingma and Max Welling.
\newblock Auto-encoding variational bayes.
\newblock \emph{arXiv preprint arXiv:1312.6114}, 2013.

\bibitem[Koohpayegani et~al.(2021)Koohpayegani, Tejankar, and Pirsiavash]{msf}
Soroush~Abbasi Koohpayegani, Ajinkya Tejankar, and Hamed Pirsiavash.
\newblock Mean shift for self-supervised learning.
\newblock In \emph{Proceedings of the IEEE/CVF International Conference on
  Computer Vision}, pp.\  10326--10335, 2021.

\bibitem[Krizhevsky \& Hinton(2009)Krizhevsky and Hinton]{cifar}
Alex Krizhevsky and Geoffrey Hinton.
\newblock {Learning multiple layers of features from tiny images}.
\newblock Technical report, Citeseer, 2009.

\bibitem[Krizhevsky et~al.(2012)Krizhevsky, Sutskever, and Hinton]{Alexnet}
Alex Krizhevsky, Ilya Sutskever, and Geoffrey~E Hinton.
\newblock {ImageNet classification with deep convolutional neural networks}.
\newblock In \emph{Advances in Neural Information Processing Systems}, 2012.

\bibitem[Kukleva et~al.(2023)Kukleva, B{\"o}hle, Schiele, Kuehne, and
  Rupprecht]{enh_ts}
Anna Kukleva, Moritz B{\"o}hle, Bernt Schiele, Hilde Kuehne, and Christian
  Rupprecht.
\newblock Temperature schedules for self-supervised contrastive methods on
  long-tail data.
\newblock \emph{arXiv preprint arXiv:2303.13664}, 2023.

\bibitem[Larsson et~al.(2016)Larsson, Maire, and
  Shakhnarovich]{colorization_ssl_2}
Gustav Larsson, Michael Maire, and Gregory Shakhnarovich.
\newblock Learning representations for automatic colorization.
\newblock In \emph{European Conference on Computer Vision}, pp.\  577--593.
  Springer, 2016.

\bibitem[Larsson et~al.(2017)Larsson, Maire, and
  Shakhnarovich]{colorization_ssl_1}
Gustav Larsson, Michael Maire, and Gregory Shakhnarovich.
\newblock Colorization as a proxy task for visual understanding.
\newblock In \emph{Proceedings of the IEEE/CVF Conference on Computer Vision
  and Pattern Recognition}, pp.\  6874--6883, 2017.

\bibitem[Lavoie et~al.(2022)Lavoie, Tsirigotis, Schwarzer, Vani, Noukhovitch,
  Kawaguchi, and Courville]{byol_sem}
Samuel Lavoie, Christos Tsirigotis, Max Schwarzer, Ankit Vani, Michael
  Noukhovitch, Kenji Kawaguchi, and Aaron Courville.
\newblock Simplicial embeddings in self-supervised learning and downstream
  classification.
\newblock \emph{arXiv preprint arXiv:2204.00616}, 2022.

\bibitem[LeCun(2016)]{lecun_cake}
Yann LeCun.
\newblock Predictive learning.
\newblock In \emph{NIPS 2016}, 2016.
\newblock URL \url{https://www.youtube.com/watch?v=Ount2Y4qxQo&t=1150s}.

\bibitem[LeCun(2019)]{lecun_ssl_facebook}
Yann LeCun.
\newblock I now call it ``self-supervised learning''.
\newblock 2019.
\newblock URL
  \url{https://www.facebook.com/yann.lecun/posts/10155934004262143}.

\bibitem[LeCun(2020)]{lecun_ssl}
Yann LeCun.
\newblock Self-supervised learning.
\newblock In \emph{Proceedings of the Thirty-Fourth AAAI Conference on
  Artificial Intelligence, AAAI, Invited Talk}, 2020.
\newblock URL
  \url{https://drive.google.com/file/d/1r-mDL4IX_hzZLDBKp8_e8VZqD7fOzBkF/view}.

\bibitem[LeCun \& Misra(2020)LeCun and Misra]{facebook_ssl}
Yann LeCun and Ishan Misra.
\newblock Self-supervised learning: The dark matter of intelligence.
\newblock In \emph{Facebook AI Blog}, 2020.
\newblock URL
  \url{https://ai.facebook.com/blog/self-supervised-learning-the-dark-matter-of-intelligence/}.

\bibitem[LeCun et~al.(1998)LeCun, Bottou, Bengio, and
  Haffner]{lecun1998gradient}
Yann LeCun, L{\'e}on Bottou, Yoshua Bengio, and Patrick Haffner.
\newblock {Gradient-based learning applied to document recognition}.
\newblock \emph{Proceedings of the IEEE}, 1998.

\bibitem[Lee et~al.(2021)Lee, Arnab, Guadarrama, Canny, and Fischer]{ceb}
Kuang-Huei Lee, Anurag Arnab, Sergio Guadarrama, John Canny, and Ian Fischer.
\newblock Compressive visual representations.
\newblock \emph{Advances in Neural Information Processing Systems},
  34:\penalty0 19538--19552, 2021.

\bibitem[Li et~al.(2022{\natexlab{a}})Li, Efros, and
  Pathak]{distilaltion_collapse}
Alexander~C Li, Alexei~A Efros, and Deepak Pathak.
\newblock Understanding collapse in non-contrastive siamese representation
  learning.
\newblock In \emph{European Conference on Computer Vision}, pp.\  490--505.
  Springer, 2022{\natexlab{a}}.

\bibitem[Li et~al.(2020{\natexlab{a}})Li, Li, Zhang, Peng, Zhou, and Gao]{hexa}
Chunyuan Li, Xiujun Li, Lei Zhang, Baolin Peng, Mingyuan Zhou, and Jianfeng
  Gao.
\newblock Self-supervised pre-training with hard examples improves visual
  representations.
\newblock \emph{arXiv preprint arXiv:2012.13493}, 2020{\natexlab{a}}.

\bibitem[Li et~al.(2021{\natexlab{a}})Li, Yang, Zhang, Gao, Xiao, Dai, Yuan,
  and Gao]{esvit}
Chunyuan Li, Jianwei Yang, Pengchuan Zhang, Mei Gao, Bin Xiao, Xiyang Dai,
  Lu~Yuan, and Jianfeng Gao.
\newblock Efficient self-supervised vision transformers for representation
  learning.
\newblock \emph{arXiv preprint arXiv:2106.09785}, 2021{\natexlab{a}}.

\bibitem[Li et~al.(2023)Li, Wang, Zhang, Chen, Jiang, Dai, Li, Xiong, and
  Tian]{pcae}
Jin Li, Yaoming Wang, Xiaopeng Zhang, Yabo Chen, Dongsheng Jiang, Wenrui Dai,
  Chenglin Li, Hongkai Xiong, and Qi~Tian.
\newblock Progressively compressed auto-encoder for self-supervised
  representation learning.
\newblock In \emph{The Eleventh International Conference on Learning
  Representations}, 2023.
\newblock URL \url{https://openreview.net/forum?id=8T4qmZbTkW7}.

\bibitem[Li et~al.(2020{\natexlab{b}})Li, Zhou, Xiong, and Hoi]{pcl}
Junnan Li, Pan Zhou, Caiming Xiong, and Steven~CH Hoi.
\newblock Prototypical contrastive learning of unsupervised representations.
\newblock \emph{arXiv preprint arXiv:2005.04966}, 2020{\natexlab{b}}.

\bibitem[Li et~al.(2021{\natexlab{b}})Li, Liu, Wang, Liu, and Zeng]{puzzle_gan}
Ru~Li, Shuaicheng Liu, Guangfu Wang, Guanghui Liu, and Bing Zeng.
\newblock {JigsawGAN}: Auxiliary learning for solving jigsaw puzzles with
  generative adversarial networks.
\newblock \emph{IEEE Transactions on Image Processing}, 31:\penalty0 513--524,
  2021{\natexlab{b}}.

\bibitem[Li et~al.(2022{\natexlab{b}})Li, Zhu, Yang, Li, Zhao, Chen, Chen, Xie,
  Wu, Zhao, et~al.]{uni_vip}
Zhaowen Li, Yousong Zhu, Fan Yang, Wei Li, Chaoyang Zhao, Yingying Chen,
  Zhiyang Chen, Jiahao Xie, Liwei Wu, Rui Zhao, et~al.
\newblock Univip: A unified framework for self-supervised visual pre-training.
\newblock In \emph{Proceedings of the IEEE/CVF Conference on Computer Vision
  and Pattern Recognition}, pp.\  14627--14636, 2022{\natexlab{b}}.

\bibitem[Lin et~al.(2014)Lin, Maire, Belongie, Hays, Perona, Ramanan,
  Doll{\'a}r, and Zitnick]{coco}
Tsung-Yi Lin, Michael Maire, Serge Belongie, James Hays, Pietro Perona, Deva
  Ramanan, Piotr Doll{\'a}r, and C~Lawrence Zitnick.
\newblock {Microsoft Coco: Common Objects In Context}.
\newblock In \emph{Proceedings of the IEEE European Conference on Computer
  Vision}, pp.\  740--755. Springer, 2014.

\bibitem[Liu et~al.(2022{\natexlab{a}})Liu, Mao, Wu, Feichtenhofer, Darrell,
  and Xie]{convnext}
Zhuang Liu, Hanzi Mao, Chao-Yuan Wu, Christoph Feichtenhofer, Trevor Darrell,
  and Saining Xie.
\newblock A convnet for the 2020s.
\newblock In \emph{Proceedings of the IEEE/CVF Conference on Computer Vision
  and Pattern Recognition}, pp.\  11976--11986, 2022{\natexlab{a}}.

\bibitem[Liu et~al.(2022{\natexlab{b}})Liu, Li, Han, Guo, and Nie]{mrcl}
Ziwen Liu, Bonan Li, Congying Han, Tiande Guo, and Xuecheng Nie.
\newblock Masked reconstruction contrastive learning with information
  bottleneck principle.
\newblock \emph{arXiv preprint arXiv:2211.09013}, 2022{\natexlab{b}}.

\bibitem[Loshchilov \& Hutter(2017)Loshchilov and Hutter]{cosine_anneal}
Ilya Loshchilov and Frank Hutter.
\newblock {Decoupled weight decay regularization}.
\newblock \emph{arXiv preprint arXiv:1711.05101}, 2017.

\bibitem[Luan et~al.(2007)Luan, Wen, Cohen-Or, Liang, Xu, and
  Shum]{image_colorization_no_dnn_1}
Qing Luan, Fang Wen, Daniel Cohen-Or, Lin Liang, Ying-Qing Xu, and Heung-Yeung
  Shum.
\newblock Natural image colorization.
\newblock In \emph{Proceedings of the 18th Eurographics conference on Rendering
  Techniques}, pp.\  309--320, 2007.

\bibitem[Lu{\v{c}}i{\'c} et~al.(2019)Lu{\v{c}}i{\'c}, Tschannen, Ritter, Zhai,
  Bachem, and Gelly]{gan_selflabel}
Mario Lu{\v{c}}i{\'c}, Michael Tschannen, Marvin Ritter, Xiaohua Zhai, Olivier
  Bachem, and Sylvain Gelly.
\newblock High-fidelity image generation with fewer labels.
\newblock In \emph{International conference on machine learning}, pp.\
  4183--4192. PMLR, 2019.

\bibitem[Mishra et~al.(2021)Mishra, Shah, Bansal, Jagannatha, Sharma, Jacobs,
  and Krishnan]{object_ssl}
Shlok Mishra, Anshul Shah, Ankan Bansal, Abhyuday Jagannatha, Abhishek Sharma,
  David Jacobs, and Dilip Krishnan.
\newblock Object-aware cropping for self-supervised learning.
\newblock \emph{arXiv preprint arXiv:2112.00319}, 2021.

\bibitem[Mishra et~al.(2023)Mishra, Robinson, Chang, Jacobs, Kuo, Sarna,
  Maschinot, and Krishnan]{can}
Shlok~Kumar Mishra, Joshua~David Robinson, Huiwen Chang, David Jacobs, Weicheng
  Kuo, Aaron Sarna, Aaron Maschinot, and Dilip Krishnan.
\newblock {CAN}: A simple, efficient and scalable contrastive masked
  autoencoder framework for learning visual representations, 2023.
\newblock URL \url{https://openreview.net/forum?id=qmV_tOHp7B9}.

\bibitem[Misra \& Maaten(2020)Misra and Maaten]{pirl}
Ishan Misra and Laurens van~der Maaten.
\newblock Self-supervised learning of pretext-invariant representations.
\newblock In \emph{Proceedings of the IEEE/CVF Conference on Computer Vision
  and Pattern Recognition}, pp.\  6707--6717, 2020.

\bibitem[Mitrovic et~al.(2020)Mitrovic, McWilliams, Walker, Buesing, and
  Blundell]{relic}
Jovana Mitrovic, Brian McWilliams, Jacob Walker, Lars Buesing, and Charles
  Blundell.
\newblock Representation learning via invariant causal mechanisms.
\newblock \emph{arXiv preprint arXiv:2010.07922}, 2020.

\bibitem[Netzer et~al.(2011)Netzer, Wang, Coates, Bissacco, Wu, and Ng]{svhn}
Yuval Netzer, Tao Wang, Adam Coates, Alessandro Bissacco, Bo~Wu, and Andrew~Y
  Ng.
\newblock {Reading digits in natural images with unsupervised feature
  learning}.
\newblock 2011.

\bibitem[Noroozi \& Favaro(2016)Noroozi and Favaro]{puzzle_ssl}
Mehdi Noroozi and Paolo Favaro.
\newblock Unsupervised learning of visual representations by solving jigsaw
  puzzles.
\newblock In \emph{European Conference on Computer Vision}, pp.\  69--84.
  Springer, 2016.

\bibitem[Noroozi et~al.(2017)Noroozi, Pirsiavash, and Favaro]{ssl_counting}
Mehdi Noroozi, Hamed Pirsiavash, and Paolo Favaro.
\newblock Representation learning by learning to count.
\newblock In \emph{Proceedings of the IEEE international conference on computer
  vision}, pp.\  5898--5906, 2017.

\bibitem[Novotny et~al.(2018)Novotny, Albanie, Larlus, and
  Vedaldi]{ssl_geometry_3}
David Novotny, Samuel Albanie, Diane Larlus, and Andrea Vedaldi.
\newblock Self-supervised learning of geometrically stable features through
  probabilistic introspection.
\newblock In \emph{Proceedings of the IEEE/CVF Conference on Computer Vision
  and Pattern Recognition}, pp.\  3637--3645, 2018.

\bibitem[Oord et~al.(2018)Oord, Li, and Vinyals]{cpc_infonce_1}
Aaron van~den Oord, Yazhe Li, and Oriol Vinyals.
\newblock Representation learning with contrastive predictive coding.
\newblock \emph{arXiv preprint arXiv:1807.03748}, 2018.

\bibitem[Pang et~al.(2022)Pang, Zhang, Li, Cai, and Lu]{smog}
Bo~Pang, Yifan Zhang, Yaoyi Li, Jia Cai, and Cewu Lu.
\newblock Unsupervised visual representation learning by synchronous momentum
  grouping.
\newblock In \emph{Computer Vision--ECCV 2022: 17th European Conference, Tel
  Aviv, Israel, October 23--27, 2022, Proceedings, Part XXX}, pp.\  265--282.
  Springer, 2022.

\bibitem[Pang et~al.(2020)Pang, Yang, Hospedales, Xiang, and
  Song]{puzzle_retrieval}
Kaiyue Pang, Yongxin Yang, Timothy~M Hospedales, Tao Xiang, and Yi-Zhe Song.
\newblock Solving mixed-modal jigsaw puzzle for fine-grained sketch-based image
  retrieval.
\newblock In \emph{Proceedings of the IEEE/CVF Conference on Computer Vision
  and Pattern Recognition}, pp.\  10347--10355, 2020.

\bibitem[Park et~al.(2023)Park, Kim, Heo, Kim, and Yun]{cl_vs_mim}
Namuk Park, Wonjae Kim, Byeongho Heo, Taekyung Kim, and Sangdoo Yun.
\newblock What do self-supervised vision transformers learn?
\newblock In \emph{The Eleventh International Conference on Learning
  Representations}, 2023.

\bibitem[Pathak et~al.(2016)Pathak, Krahenbuhl, Donahue, Darrell, and
  Efros]{inpainting_ssl}
Deepak Pathak, Philipp Krahenbuhl, Jeff Donahue, Trevor Darrell, and Alexei~A
  Efros.
\newblock Context encoders: Feature learning by inpainting.
\newblock In \emph{Proceedings of the IEEE/CVF Conference on Computer Vision
  and Pattern Recognition}, pp.\  2536--2544, 2016.

\bibitem[Peng et~al.(2022)Peng, Dong, Bao, Ye, and Wei]{beit-v2}
Zhiliang Peng, Li~Dong, Hangbo Bao, Qixiang Ye, and Furu Wei.
\newblock {BEiT} v2: Masked image modeling with vector-quantized visual
  tokenizers.
\newblock \emph{arXiv preprint arXiv:2208.06366}, 2022.

\bibitem[Qian et~al.(2022)Qian, Xu, Hu, Li, and Jin]{coke}
Qi~Qian, Yuanhong Xu, Juhua Hu, Hao Li, and Rong Jin.
\newblock Unsupervised visual representation learning by online constrained
  k-means.
\newblock In \emph{Proceedings of the IEEE/CVF Conference on Computer Vision
  and Pattern Recognition}, pp.\  16640--16649, 2022.

\bibitem[Radford et~al.(2019)Radford, Wu, Child, Luan, Amodei, Sutskever,
  et~al.]{gpt_2}
Alec Radford, Jeffrey Wu, Rewon Child, David Luan, Dario Amodei, Ilya
  Sutskever, et~al.
\newblock Language models are unsupervised multitask learners.
\newblock \emph{OpenAI blog}, 1\penalty0 (8):\penalty0 9, 2019.

\bibitem[Radford et~al.(2021)Radford, Kim, Hallacy, Ramesh, Goh, Agarwal,
  Sastry, Askell, Mishkin, Clark, et~al.]{clip_tokenizer}
Alec Radford, Jong~Wook Kim, Chris Hallacy, Aditya Ramesh, Gabriel Goh,
  Sandhini Agarwal, Girish Sastry, Amanda Askell, Pamela Mishkin, Jack Clark,
  et~al.
\newblock Learning transferable visual models from natural language
  supervision.
\newblock In \emph{International Conference on Machine Learning}, pp.\
  8748--8763. PMLR, 2021.

\bibitem[Ramesh et~al.(2021)Ramesh, Pavlov, Goh, Gray, Voss, Radford, Chen, and
  Sutskever]{tokenizer}
Aditya Ramesh, Mikhail Pavlov, Gabriel Goh, Scott Gray, Chelsea Voss, Alec
  Radford, Mark Chen, and Ilya Sutskever.
\newblock Zero-shot text-to-image generation.
\newblock In \emph{International Conference on Machine Learning}, pp.\
  8821--8831. PMLR, 2021.

\bibitem[Ramesh et~al.(2022)Ramesh, Srivastav, Alapatt, Yu, Murali, Sestini,
  Nwoye, Hamoud, Fleurentin, Exarchakis, et~al.]{ramesh2022dissecting}
Sanat Ramesh, Vinkle Srivastav, Deepak Alapatt, Tong Yu, Aditya Murali, Luca
  Sestini, Chinedu~Innocent Nwoye, Idris Hamoud, Antoine Fleurentin, Georgios
  Exarchakis, et~al.
\newblock Dissecting self-supervised learning methods for surgical computer
  vision.
\newblock \emph{arXiv preprint arXiv:2207.00449}, 2022.

\bibitem[Ren et~al.(2023)Ren, Wei, Zhang, and Hu]{tinymim}
Sucheng Ren, Fangyun Wei, Zheng Zhang, and Han Hu.
\newblock {TinyMIM}: An empirical study of distilling {MIM} pre-trained models.
\newblock \emph{arXiv preprint arXiv:2301.01296}, 2023.

\bibitem[Ren \& Lee(2018)Ren and Lee]{ssl_syntethic}
Zhongzheng Ren and Yong~Jae Lee.
\newblock Cross-domain self-supervised multi-task feature learning using
  synthetic imagery.
\newblock In \emph{Proceedings of the IEEE/CVF Conference on Computer Vision
  and Pattern Recognition}, pp.\  762--771, 2018.

\bibitem[Richemond et~al.(2020)Richemond, Grill, Altch{\'e}, Tallec, Strub,
  Brock, Smith, De, Pascanu, Piot, et~al.]{byol_works}
Pierre~H Richemond, Jean-Bastien Grill, Florent Altch{\'e}, Corentin Tallec,
  Florian Strub, Andrew Brock, Samuel Smith, Soham De, Razvan Pascanu, Bilal
  Piot, et~al.
\newblock {BYOL} works even without batch statistics.
\newblock \emph{arXiv preprint arXiv:2010.10241}, 2020.

\bibitem[Robinson et~al.(2020)Robinson, Chuang, Sra, and
  Jegelka]{negative_samples_3}
Joshua Robinson, Ching-Yao Chuang, Suvrit Sra, and Stefanie Jegelka.
\newblock Contrastive learning with hard negative samples.
\newblock \emph{arXiv preprint arXiv:2010.04592}, 2020.

\bibitem[Rocco et~al.(2017)Rocco, Arandjelovic, and Sivic]{ssl_geometry_2}
Ignacio Rocco, Relja Arandjelovic, and Josef Sivic.
\newblock Convolutional neural network architecture for geometric matching.
\newblock In \emph{Proceedings of the IEEE/CVF Conference on Computer Vision
  and Pattern Recognition}, pp.\  6148--6157, 2017.

\bibitem[Ruan et~al.(2022)Ruan, Singh, Morningstar, Alemi, Ioffe, Fischer, and
  Dillon]{enh_ens}
Yangjun Ruan, Saurabh Singh, Warren Morningstar, Alexander~A Alemi, Sergey
  Ioffe, Ian Fischer, and Joshua~V Dillon.
\newblock Weighted ensemble self-supervised learning.
\newblock \emph{arXiv preprint arXiv:2211.09981}, 2022.

\bibitem[Russakovsky et~al.(2015)Russakovsky, Deng, Su, Krause, Satheesh, Ma,
  Huang, Karpathy, Khosla, Bernstein, Berg, and Fei-Fei]{ILSVRC15:rus}
Olga Russakovsky, Jia Deng, Hao Su, Jonathan Krause, Sanjeev Satheesh, Sean Ma,
  Zhiheng Huang, Andrej Karpathy, Aditya Khosla, Michael Bernstein,
  Alexander~C. Berg, and Li~Fei-Fei.
\newblock {ImageNet large scale visual recognition challenge}.
\newblock \emph{International Journal of Computer Vision}, 115\penalty0
  (3):\penalty0 211--252, 2015.

\bibitem[Shah et~al.(2023)Shah, Roy, Shah, Mishra, Jacobs, Cherian, and
  Chellappa]{shah2023halp}
Anshul Shah, Aniket Roy, Ketul Shah, Shlok~Kumar Mishra, David Jacobs, Anoop
  Cherian, and Rama Chellappa.
\newblock {HaLP}: Hallucinating latent positives for skeleton-based
  self-supervised learning of actions.
\newblock \emph{arXiv preprint arXiv:2304.00387}, 2023.

\bibitem[Simonyan \& Zisserman(2015)Simonyan and Zisserman]{VGG}
Karen Simonyan and Andrew Zisserman.
\newblock {Very deep convolutional networks for large-scale image recognition}.
\newblock \emph{International Conference on Learning Representations}, 2015.

\bibitem[Sohn(2016)]{infonce_2}
Kihyuk Sohn.
\newblock Improved deep metric learning with multi-class n-pair loss objective.
\newblock \emph{Advances in Neural Information Processing Systems}, 29, 2016.

\bibitem[Sohn \& Lee(2012)Sohn and Lee]{ssl_discriminative_2}
Kihyuk Sohn and Honglak Lee.
\newblock Learning invariant representations with local transformations.
\newblock \emph{arXiv preprint arXiv:1206.6418}, 2012.

\bibitem[Susmelj et~al.(2020)Susmelj, Heller, Wirth, Prescott, and
  et~al.]{lightly}
Igor Susmelj, Matthias Heller, Philipp Wirth, Jeremy Prescott, and Malte~Ebner
  et~al.
\newblock Lightly.
\newblock \emph{GitHub. Note: https://github.com/lightly-ai/lightly}, 2020.

\bibitem[Szegedy et~al.(2015)Szegedy, Liu, Jia, Sermanet, Reed, Anguelov,
  Erhan, Vanhoucke, and Rabinovich]{Inceptionv1}
Christian Szegedy, Wei Liu, Yangqing Jia, Pierre Sermanet, Scott Reed, Dragomir
  Anguelov, Dumitru Erhan, Vincent Vanhoucke, and Andrew Rabinovich.
\newblock {Going deeper with convolutions}.
\newblock In \emph{Proceedings of the IEEE/CVF Conference on Computer Vision
  and Pattern Recognition}, pp.\  1--9, 2015.

\bibitem[Szegedy et~al.(2016)Szegedy, Vanhoucke, Ioffe, Shlens, and
  Wojna]{inceptionv3}
Christian Szegedy, Vincent Vanhoucke, Sergey Ioffe, Jon Shlens, and Zbigniew
  Wojna.
\newblock {Rethinking the inception architecture for computer vision}.
\newblock In \emph{Proceedings of the IEEE/CVF Conference on Computer Vision
  and Pattern Recognition}, 2016.

\bibitem[Tan et~al.(2018)Tan, Sun, Kong, Zhang, Yang, and
  Liu]{transfer_learning_1}
Chuanqi Tan, Fuchun Sun, Tao Kong, Wenchang Zhang, Chao Yang, and Chunfang Liu.
\newblock A survey on deep transfer learning.
\newblock In \emph{International Conference on Artificial Neural Networks},
  pp.\  270--279. Springer, 2018.

\bibitem[Tao et~al.(2022{\natexlab{a}})Tao, Wang, Zhu, Dong, Song, Huang, and
  Dai]{uni_grad}
Chenxin Tao, Honghui Wang, Xizhou Zhu, Jiahua Dong, Shiji Song, Gao Huang, and
  Jifeng Dai.
\newblock Exploring the equivalence of siamese self-supervised learning via a
  unified gradient framework.
\newblock In \emph{Proceedings of the IEEE/CVF Conference on Computer Vision
  and Pattern Recognition}, pp.\  14431--14440, 2022{\natexlab{a}}.

\bibitem[Tao et~al.(2022{\natexlab{b}})Tao, Zhu, Huang, Qiao, Wang, and
  Dai]{siameseim}
Chenxin Tao, Xizhou Zhu, Gao Huang, Yu~Qiao, Xiaogang Wang, and Jifeng Dai.
\newblock Siamese image modeling for self-supervised vision representation
  learning.
\newblock \emph{arXiv preprint arXiv:2206.01204}, 2022{\natexlab{b}}.

\bibitem[Tarvainen \& Valpola(2017)Tarvainen and Valpola]{mean_teacher}
Antti Tarvainen and Harri Valpola.
\newblock Mean teachers are better role models: Weight-averaged consistency
  targets improve semi-supervised deep learning results.
\newblock \emph{Advances in Neural Information Processing Systems}, 30, 2017.

\bibitem[Tian et~al.(2023)Tian, Jiang, Diao, Lin, Wang, and Yuan]{spark}
Keyu Tian, Yi~Jiang, Qishuai Diao, Chen Lin, Liwei Wang, and Zehuan Yuan.
\newblock Designing bert for convolutional networks: Sparse and hierarchical
  masked modeling.
\newblock \emph{arXiv preprint arXiv:2301.03580}, 2023.

\bibitem[Tian et~al.(2020{\natexlab{a}})Tian, Krishnan, and Isola]{cmc}
Yonglong Tian, Dilip Krishnan, and Phillip Isola.
\newblock Contrastive multiview coding.
\newblock In \emph{European Conference on Computer Vision}, pp.\  776--794.
  Springer, 2020{\natexlab{a}}.

\bibitem[Tian et~al.(2020{\natexlab{b}})Tian, Sun, Poole, Krishnan, Schmid, and
  Isola]{info_min}
Yonglong Tian, Chen Sun, Ben Poole, Dilip Krishnan, Cordelia Schmid, and
  Phillip Isola.
\newblock What makes for good views for contrastive learning?
\newblock \emph{Advances in Neural Information Processing Systems},
  33:\penalty0 6827--6839, 2020{\natexlab{b}}.

\bibitem[Tian et~al.(2021{\natexlab{a}})Tian, Henaff, and van~den Oord]{dnc}
Yonglong Tian, Olivier~J Henaff, and A{\"a}ron van~den Oord.
\newblock Divide and contrast: Self-supervised learning from uncurated data.
\newblock In \emph{Proceedings of the IEEE/CVF International Conference on
  Computer Vision}, pp.\  10063--10074, 2021{\natexlab{a}}.

\bibitem[Tian(2023)]{exp3}
Yuandong Tian.
\newblock Understanding the role of nonlinearity in training dynamics of
  contrastive learning.
\newblock In \emph{The Eleventh International Conference on Learning
  Representations}, 2023.
\newblock URL \url{https://openreview.net/forum?id=s130rTE3U_X}.

\bibitem[Tian et~al.(2020{\natexlab{c}})Tian, Yu, Chen, and
  Ganguli]{infonce_understand1}
Yuandong Tian, Lantao Yu, Xinlei Chen, and Surya Ganguli.
\newblock Understanding self-supervised learning with dual deep networks.
\newblock \emph{arXiv preprint arXiv:2010.00578}, 2020{\natexlab{c}}.

\bibitem[Tian et~al.(2021{\natexlab{b}})Tian, Chen, and Ganguli]{directPred}
Yuandong Tian, Xinlei Chen, and Surya Ganguli.
\newblock Understanding self-supervised learning dynamics without contrastive
  pairs.
\newblock In \emph{International Conference on Machine Learning}, pp.\
  10268--10278. PMLR, 2021{\natexlab{b}}.

\bibitem[Tomasev et~al.(2022)Tomasev, Bica, McWilliams, Buesing, Pascanu,
  Blundell, and Mitrovic]{relic_2}
Nenad Tomasev, Ioana Bica, Brian McWilliams, Lars Buesing, Razvan Pascanu,
  Charles Blundell, and Jovana Mitrovic.
\newblock Pushing the limits of self-supervised {R}es{N}ets: Can we outperform
  supervised learning without labels on {I}mage{N}et?
\newblock \emph{arXiv preprint arXiv:2201.05119}, 2022.

\bibitem[Trinh et~al.(2019)Trinh, Luong, and Le]{puzzle_embedding}
Trieu~H Trinh, Minh-Thang Luong, and Quoc~V Le.
\newblock Selfie: Self-supervised pretraining for image embedding.
\newblock \emph{arXiv preprint arXiv:1906.02940}, 2019.

\bibitem[Van~Gansbeke et~al.(2020)Van~Gansbeke, Vandenhende, Georgoulis,
  Proesmans, and Van~Gool]{scan}
Wouter Van~Gansbeke, Simon Vandenhende, Stamatios Georgoulis, Marc Proesmans,
  and Luc Van~Gool.
\newblock Scan: Learning to classify images without labels.
\newblock In \emph{European Conference on Computer Vision}, pp.\  268--285.
  Springer, 2020.

\bibitem[Vincent et~al.(2008)Vincent, Larochelle, Bengio, and
  Manzagol]{autoencoders}
Pascal Vincent, Hugo Larochelle, Yoshua Bengio, and Pierre-Antoine Manzagol.
\newblock Extracting and composing robust features with denoising autoencoders.
\newblock In \emph{Proceedings of the 25th International Conference on Machine
  Learning}, pp.\  1096--1103, 2008.

\bibitem[Wang et~al.(2021{\natexlab{a}})Wang, Kong, Zhang, Liu, and Li]{twist}
Feng Wang, Tao Kong, Rufeng Zhang, Huaping Liu, and Hang Li.
\newblock Self-supervised learning by estimating twin class distributions.
\newblock \emph{arXiv preprint arXiv:2110.07402}, 2021{\natexlab{a}}.

\bibitem[Wang et~al.(2021{\natexlab{b}})Wang, Wang, Wang, Torr, and
  Lin]{truncated_triplet}
Guangrun Wang, Keze Wang, Guangcong Wang, Philip~HS Torr, and Liang Lin.
\newblock Solving inefficiency of self-supervised representation learning.
\newblock In \emph{Proceedings of the IEEE/CVF International Conference on
  Computer Vision}, pp.\  9505--9515, 2021{\natexlab{b}}.

\bibitem[Wang et~al.(2022{\natexlab{a}})Wang, Bao, Dong, Bjorck, Peng, Liu,
  Aggarwal, Mohammed, Singhal, Som, et~al.]{beit-v3}
Wenhui Wang, Hangbo Bao, Li~Dong, Johan Bjorck, Zhiliang Peng, Qiang Liu, Kriti
  Aggarwal, Owais~Khan Mohammed, Saksham Singhal, Subhojit Som, et~al.
\newblock Image as a foreign language: {BEiT} pretraining for all vision and
  vision-language tasks.
\newblock \emph{arXiv preprint arXiv:2208.10442}, 2022{\natexlab{a}}.

\bibitem[Wang \& Qi(2022)Wang and Qi]{csla}
Xiao Wang and Guo-Jun Qi.
\newblock Contrastive learning with stronger augmentations.
\newblock \emph{IEEE Transactions on Pattern Analysis and Machine
  Intelligence}, 2022.

\bibitem[Wang et~al.(2022{\natexlab{b}})Wang, Huang, Zeng, and Qi]{caco}
Xiao Wang, Yuhang Huang, Dan Zeng, and Guo-Jun Qi.
\newblock Caco: Both positive and negative samples are directly learnable via
  cooperative-adversarial contrastive learning.
\newblock \emph{arXiv preprint arXiv:2203.14370}, 2022{\natexlab{b}}.

\bibitem[Wang et~al.(2021{\natexlab{c}})Wang, Zhang, Shen, Kong, and
  Li]{dense_cl}
Xinlong Wang, Rufeng Zhang, Chunhua Shen, Tao Kong, and Lei Li.
\newblock Dense contrastive learning for self-supervised visual pre-training.
\newblock In \emph{Proceedings of the IEEE/CVF Conference on Computer Vision
  and Pattern Recognition}, pp.\  3024--3033, 2021{\natexlab{c}}.

\bibitem[Wang et~al.(2022{\natexlab{c}})Wang, Chen, Fan, Sun, Tao, Hou, Wang,
  Yang, Zhou, Guo, Qi, Wu, Li, Nakamura, Ye, Savvides, Raj, Shinozaki, Schiele,
  Wang, Xie, and Zhang]{usb_microsoft}
Yidong Wang, Hao Chen, Yue Fan, Wang Sun, Ran Tao, Wenxin Hou, Renjie Wang,
  Linyi Yang, Zhi Zhou, Lan-Zhe Guo, Heli Qi, Zhen Wu, Yu-Feng Li, Satoshi
  Nakamura, Wei Ye, Marios Savvides, Bhiksha Raj, Takahiro Shinozaki, Bernt
  Schiele, Jindong Wang, Xing Xie, and Yue Zhang.
\newblock {USB}: A unified semi-supervised learning benchmark for
  classification.
\newblock In \emph{Thirty-sixth Conference on Neural Information Processing
  Systems Datasets and Benchmarks Track}, 2022{\natexlab{c}}.
\newblock \doi{10.48550/arxiv.2208.07204}.
\newblock URL \url{https://arxiv.org/abs/2208.07204}.

\bibitem[Wang et~al.(2022{\natexlab{d}})Wang, Li, Zhang, Wan, Zheng, Wang,
  Gong, and Liu]{aug_future_2}
Zhaoqing Wang, Qiang Li, Guoxin Zhang, Pengfei Wan, Wen Zheng, Nannan Wang,
  Mingming Gong, and Tongliang Liu.
\newblock Exploring set similarity for dense self-supervised representation
  learning.
\newblock In \emph{Proceedings of the IEEE/CVF Conference on Computer Vision
  and Pattern Recognition}, pp.\  16590--16599, 2022{\natexlab{d}}.

\bibitem[Wang et~al.(2023)Wang, Chen, Li, Guo, Yu, Gong, and Liu]{wangmosaic}
Zhaoqing Wang, Ziyu Chen, Yaqian Li, Yandong Guo, Jun Yu, Mingming Gong, and
  Tongliang Liu.
\newblock Mosaic representation learning for self-supervised visual
  pre-training.
\newblock In \emph{The Eleventh International Conference on Learning
  Representations}, 2023.

\bibitem[Wei et~al.(2022)Wei, Fan, Xie, Wu, Yuille, and
  Feichtenhofer]{masked_feature_modeling}
Chen Wei, Haoqi Fan, Saining Xie, Chao-Yuan Wu, Alan Yuille, and Christoph
  Feichtenhofer.
\newblock Masked feature prediction for self-supervised visual pre-training.
\newblock In \emph{Proceedings of the IEEE/CVF Conference on Computer Vision
  and Pattern Recognition}, pp.\  14668--14678, 2022.

\bibitem[Wu et~al.(2020)Wu, Xu, Dai, Wan, Zhang, Yan, Tomizuka, Gonzalez,
  Keutzer, and Vajda]{image_tokenization}
Bichen Wu, Chenfeng Xu, Xiaoliang Dai, Alvin Wan, Peizhao Zhang, Zhicheng Yan,
  Masayoshi Tomizuka, Joseph Gonzalez, Kurt Keutzer, and Peter Vajda.
\newblock Visual transformers: Token-based image representation and processing
  for computer vision.
\newblock \emph{arXiv preprint arXiv:2006.03677}, 2020.

\bibitem[Wu et~al.(2018{\natexlab{a}})Wu, Efros, and Yu]{memory_bank}
Zhirong Wu, Alexei~A Efros, and Stella~X Yu.
\newblock Improving generalization via scalable neighborhood component
  analysis.
\newblock In \emph{Proceedings of the European Conference on Computer Vision
  (ECCV)}, pp.\  685--701, 2018{\natexlab{a}}.

\bibitem[Wu et~al.(2018{\natexlab{b}})Wu, Xiong, Yu, and Lin]{instdist}
Zhirong Wu, Yuanjun Xiong, Stella~X Yu, and Dahua Lin.
\newblock Unsupervised feature learning via non-parametric instance
  discrimination.
\newblock In \emph{Proceedings of the IEEE/CVF Conference on Computer Vision
  and Pattern Recognition}, pp.\  3733--3742, 2018{\natexlab{b}}.

\bibitem[Xiao et~al.(2020)Xiao, Wang, Efros, and Darrell]{looc}
Tete Xiao, Xiaolong Wang, Alexei~A Efros, and Trevor Darrell.
\newblock What should not be contrastive in contrastive learning.
\newblock \emph{arXiv preprint arXiv:2008.05659}, 2020.

\bibitem[Xiao et~al.(2021)Xiao, Reed, Wang, Keutzer, and Darrell]{resim}
Tete Xiao, Colorado~J Reed, Xiaolong Wang, Kurt Keutzer, and Trevor Darrell.
\newblock Region similarity representation learning.
\newblock In \emph{Proceedings of the IEEE/CVF International Conference on
  Computer Vision}, pp.\  10539--10548, 2021.

\bibitem[Xie et~al.(2021{\natexlab{a}})Xie, Zhan, Liu, Ong, and Loy]{orl}
Jiahao Xie, Xiaohang Zhan, Ziwei Liu, Yew~Soon Ong, and Chen~Change Loy.
\newblock Unsupervised object-level representation learning from scene images.
\newblock \emph{Advances in Neural Information Processing Systems},
  34:\penalty0 28864--28876, 2021{\natexlab{a}}.

\bibitem[Xie et~al.(2022{\natexlab{a}})Xie, Zhan, Liu, Ong, and Loy]{inter_clr}
Jiahao Xie, Xiaohang Zhan, Ziwei Liu, Yew-Soon Ong, and Chen~Change Loy.
\newblock Delving into inter-image invariance for unsupervised visual
  representations.
\newblock \emph{International Journal of Computer Vision}, pp.\  1--20,
  2022{\natexlab{a}}.

\bibitem[Xie et~al.(2021{\natexlab{b}})Xie, Lin, Yao, Zhang, Dai, Cao, and
  Hu]{moby}
Zhenda Xie, Yutong Lin, Zhuliang Yao, Zheng Zhang, Qi~Dai, Yue Cao, and Han Hu.
\newblock Self-supervised learning with swin transformers.
\newblock \emph{arXiv preprint arXiv:2105.04553}, 2021{\natexlab{b}}.

\bibitem[Xie et~al.(2021{\natexlab{c}})Xie, Lin, Zhang, Cao, Lin, and
  Hu]{pix_pro}
Zhenda Xie, Yutong Lin, Zheng Zhang, Yue Cao, Stephen Lin, and Han Hu.
\newblock Propagate yourself: Exploring pixel-level consistency for
  unsupervised visual representation learning.
\newblock In \emph{Proceedings of the IEEE/CVF Conference on Computer Vision
  and Pattern Recognition}, pp.\  16684--16693, 2021{\natexlab{c}}.

\bibitem[Xie et~al.(2022{\natexlab{b}})Xie, Zhang, Cao, Lin, Bao, Yao, Dai, and
  Hu]{simmim}
Zhenda Xie, Zheng Zhang, Yue Cao, Yutong Lin, Jianmin Bao, Zhuliang Yao,
  Qi~Dai, and Han Hu.
\newblock Simmim: A simple framework for masked image modeling.
\newblock In \emph{Proceedings of the IEEE/CVF Conference on Computer Vision
  and Pattern Recognition}, pp.\  9653--9663, 2022{\natexlab{b}}.

\bibitem[Xu et~al.(2021)Xu, Fang, Zhang, Xie, Wang, Dai, Xiong, and
  Tian]{bingo}
Haohang Xu, Jiemin Fang, Xiaopeng Zhang, Lingxi Xie, Xinggang Wang, Wenrui Dai,
  Hongkai Xiong, and Qi~Tian.
\newblock Bag of instances aggregation boosts self-supervised distillation.
\newblock In \emph{International Conference on Learning Representations}, 2021.

\bibitem[Xu et~al.(2004)Xu, Neufeld, Larson, and
  Schuurmans]{clustering_collapse_2}
Linli Xu, James Neufeld, Bryce Larson, and Dale Schuurmans.
\newblock Maximum margin clustering.
\newblock \emph{Advances in Neural Information Processing Systems}, 17, 2004.

\bibitem[Yang et~al.(2017)Yang, Lu, Lin, Shechtman, Wang, and Li]{inpainting_1}
Chao Yang, Xin Lu, Zhe Lin, Eli Shechtman, Oliver Wang, and Hao Li.
\newblock High-resolution image inpainting using multi-scale neural patch
  synthesis.
\newblock In \emph{Proceedings of the IEEE/CVF Conference on Computer Vision
  and Pattern Recognition}, pp.\  6721--6729, 2017.

\bibitem[Yang et~al.(2016)Yang, Parikh, and Batra]{clustering_early_2}
Jianwei Yang, Devi Parikh, and Dhruv Batra.
\newblock Joint unsupervised learning of deep representations and image
  clusters.
\newblock In \emph{Proceedings of the IEEE/CVF Conference on Computer Vision
  and Pattern Recognition}, pp.\  5147--5156, 2016.

\bibitem[Yi et~al.(2022)Yi, Ge, Li, Yang, Li, Wu, Shan, and Qie]{conmim}
Kun Yi, Yixiao Ge, Xiaotong Li, Shusheng Yang, Dian Li, Jianping Wu, Ying Shan,
  and Xiaohu Qie.
\newblock Masked image modeling with denoising contrast.
\newblock \emph{arXiv preprint arXiv:2205.09616}, 2022.

\bibitem[Yu et~al.(2019)Yu, Lin, Yang, Shen, Lu, and Huang]{inpainting_2}
Jiahui Yu, Zhe Lin, Jimei Yang, Xiaohui Shen, Xin Lu, and Thomas~S Huang.
\newblock Free-form image inpainting with gated convolution.
\newblock In \emph{Proceedings of the IEEE/CVF international conference on
  computer vision}, pp.\  4471--4480, 2019.

\bibitem[Yu et~al.(2021)Yu, Li, Koh, Zhang, Pang, Qin, Ku, Xu, Baldridge, and
  Wu]{vit_vqgan}
Jiahui Yu, Xin Li, Jing~Yu Koh, Han Zhang, Ruoming Pang, James Qin, Alexander
  Ku, Yuanzhong Xu, Jason Baldridge, and Yonghui Wu.
\newblock Vector-quantized image modeling with improved {VQGAN}.
\newblock \emph{arXiv preprint arXiv:2110.04627}, 2021.

\bibitem[Yun et~al.(2019)Yun, Han, Oh, Chun, Choe, and Yoo]{cutmix}
Sangdoo Yun, Dongyoon Han, Seong~Joon Oh, Sanghyuk Chun, Junsuk Choe, and
  Youngjoon Yoo.
\newblock {Cutmix: Regularization strategy to train strong classifiers with
  localizable features}.
\newblock In \emph{Proceedings of the IEEE/CVF international conference on
  computer vision}, pp.\  6023--6032, 2019.

\bibitem[Zbontar et~al.(2021)Zbontar, Jing, Misra, LeCun, and
  Deny]{barlow_twins}
Jure Zbontar, Li~Jing, Ishan Misra, Yann LeCun, and St{\'e}phane Deny.
\newblock Barlow twins: Self-supervised learning via redundancy reduction.
\newblock In \emph{International Conference on Machine Learning}, pp.\
  12310--12320. PMLR, 2021.

\bibitem[Zhan et~al.(2020)Zhan, Xie, Liu, Ong, and Loy]{online_deep_clustering}
Xiaohang Zhan, Jiahao Xie, Ziwei Liu, Yew-Soon Ong, and Chen~Change Loy.
\newblock Online deep clustering for unsupervised representation learning.
\newblock In \emph{Proceedings of the IEEE/CVF Conference on Computer Vision
  and Pattern Recognition}, pp.\  6688--6697, 2020.

\bibitem[Zhang et~al.(2022{\natexlab{a}})Zhang, Zhang, Pham, Niu, Qiao, Yoo,
  and Kweon]{negative_samples_2}
Chaoning Zhang, Kang Zhang, Trung~X Pham, Axi Niu, Zhinan Qiao, Chang~D Yoo,
  and In~So Kweon.
\newblock Dual temperature helps contrastive learning without many negative
  samples: Towards understanding and simplifying {MoCo}.
\newblock In \emph{Proceedings of the IEEE/CVF Conference on Computer Vision
  and Pattern Recognition}, pp.\  14441--14450, 2022{\natexlab{a}}.

\bibitem[Zhang et~al.(2019)Zhang, Qi, Wang, and Luo]{ssl_geometry_4}
Liheng Zhang, Guo-Jun Qi, Liqiang Wang, and Jiebo Luo.
\newblock {AET} vs. {AED}: Unsupervised representation learning by
  auto-encoding transformations rather than data.
\newblock In \emph{Proceedings of the IEEE/CVF Conference on Computer Vision
  and Pattern Recognition}, pp.\  2547--2555, 2019.

\bibitem[Zhang et~al.(2016)Zhang, Isola, and Efros]{colorization_ssl_3}
Richard Zhang, Phillip Isola, and Alexei~A Efros.
\newblock Colorful image colorization.
\newblock In \emph{European Conference on Computer Vision}, pp.\  649--666.
  Springer, 2016.

\bibitem[Zhang et~al.(2017)Zhang, Isola, and Efros]{ssl_split_brain}
Richard Zhang, Phillip Isola, and Alexei~A Efros.
\newblock Split-brain autoencoders: Unsupervised learning by cross-channel
  prediction.
\newblock In \emph{Proceedings of the IEEE/CVF Conference on Computer Vision
  and Pattern Recognition}, pp.\  1058--1067, 2017.

\bibitem[Zhang et~al.(2022{\natexlab{b}})Zhang, Qiu, Zhu, Yan, Zhang, Zhao, Li,
  and Yang]{arb}
Shaofeng Zhang, Lyn Qiu, Feng Zhu, Junchi Yan, Hengrui Zhang, Rui Zhao,
  Hongyang Li, and Xiaokang Yang.
\newblock Align representations with base: A new approach to self-supervised
  learning.
\newblock In \emph{Proceedings of the IEEE/CVF Conference on Computer Vision
  and Pattern Recognition}, pp.\  16600--16609, 2022{\natexlab{b}}.

\bibitem[Zhang et~al.(2022{\natexlab{c}})Zhang, Qiu, Ke, S{\"u}sstrunk, and
  Salzmann]{aug_future_1}
Tong Zhang, Congpei Qiu, Wei Ke, Sabine S{\"u}sstrunk, and Mathieu Salzmann.
\newblock Leverage your local and global representations: A new self-supervised
  learning strategy.
\newblock In \emph{Proceedings of the IEEE/CVF Conference on Computer Vision
  and Pattern Recognition}, pp.\  16580--16589, 2022{\natexlab{c}}.

\bibitem[Zhang et~al.(2022{\natexlab{d}})Zhang, Chen, Yuan, Chen, Wang, Wang,
  Han, Chen, Pi, Yao, et~al.]{cae-v2}
Xinyu Zhang, Jiahui Chen, Junkun Yuan, Qiang Chen, Jian Wang, Xiaodi Wang,
  Shumin Han, Xiaokang Chen, Jimin Pi, Kun Yao, et~al.
\newblock {CAE} v2: Context autoencoder with {CLIP} target.
\newblock \emph{arXiv preprint arXiv:2211.09799}, 2022{\natexlab{d}}.

\bibitem[Zhao et~al.(2023)Zhao, Du, Wang, Yao, and Huang]{enh_arcl}
Xuyang Zhao, Tianqi Du, Yisen Wang, Jun Yao, and Weiran Huang.
\newblock {ArCL}: Enhancing contrastive learning with augmentation-robust
  representations.
\newblock \emph{arXiv preprint arXiv:2303.01092}, 2023.

\bibitem[Zheng et~al.(2021)Zheng, You, Wang, Qian, Zhang, Wang, and Xu]{ressl}
Mingkai Zheng, Shan You, Fei Wang, Chen Qian, Changshui Zhang, Xiaogang Wang,
  and Chang Xu.
\newblock {ReSSL}: Relational self-supervised learning with weak augmentation.
\newblock \emph{Advances in Neural Information Processing Systems},
  34:\penalty0 2543--2555, 2021.

\bibitem[Zhong et~al.(2022)Zhong, Tang, Chen, Peng, and Wang]{ssl_robust}
Yuanyi Zhong, Haoran Tang, Junkun Chen, Jian Peng, and Yu-Xiong Wang.
\newblock Is self-supervised contrastive learning more robust than supervised
  learning?
\newblock In \emph{First Workshop on Pre-training: Perspectives, Pitfalls, and
  Paths Forward at ICML 2022}, 2022.

\bibitem[Zhou et~al.(2021)Zhou, Wei, Wang, Shen, Xie, Yuille, and Kong]{ibot}
Jinghao Zhou, Chen Wei, Huiyu Wang, Wei Shen, Cihang Xie, Alan Yuille, and Tao
  Kong.
\newblock {iBOT}: Image {BERT} pre-training with online tokenizer.
\newblock \emph{arXiv preprint arXiv:2111.07832}, 2021.

\bibitem[Zhou et~al.(2022)Zhou, Zhou, Si, Yu, Ng, and Yan]{mugs}
Pan Zhou, Yichen Zhou, Chenyang Si, Weihao Yu, Teck~Khim Ng, and Shuicheng Yan.
\newblock Mugs: A multi-granular self-supervised learning framework.
\newblock \emph{arXiv preprint arXiv:2203.14415}, 2022.

\bibitem[Zhuang et~al.(2019)Zhuang, Zhai, and Yamins]{local_agg}
Chengxu Zhuang, Alex~Lin Zhai, and Daniel Yamins.
\newblock Local aggregation for unsupervised learning of visual embeddings.
\newblock In \emph{Proceedings of the IEEE/CVF International Conference on
  Computer Vision}, pp.\  6002--6012, 2019.

\bibitem[Zisserman(2018)]{zisserman_ssl}
Andrew Zisserman.
\newblock Self-supervised learning.
\newblock 2018.
\newblock URL
  \url{https://project.inria.fr/paiss/files/2018/07/zisserman-self-supervised.pdf}.

\end{thebibliography}
\bibliographystyle{tmlr}

\end{document}